\documentclass{article}

\usepackage[preprint]{arxiv_style}

\usepackage[utf8]{inputenc}
\usepackage[T1]{fontenc}
\usepackage{microtype}

\usepackage{amsmath,amsfonts,bm}

\def\eqref#1{equation~\ref{#1}}

\def\1{\bm{1}}

\DeclareMathAlphabet{\mathsfit}{\encodingdefault}{\sfdefault}{m}{sl}
\SetMathAlphabet{\mathsfit}{bold}{\encodingdefault}{\sfdefault}{bx}{n}

\newcommand{\E}{\mathbb{E}}

\newcommand{\R}{\mathbb{R}}

\newcommand{\Var}{\mathrm{Var}}

\usepackage{amsmath,amssymb,amsthm,mathtools}
\usepackage{graphicx}
\usepackage{wrapfig}
\usepackage{booktabs}
\usepackage{adjustbox}
\usepackage[table]{xcolor}   
\usepackage{pifont}
\usepackage{enumitem}
\usepackage{url}
\usepackage{hyperref}        

\theoremstyle{definition}
\newtheorem{definition}{Definition}
\newtheorem{assumption}{Assumption}
\newtheorem{proposition}{Proposition}
\theoremstyle{plain}
\newtheorem{theorem}{Theorem}
\newtheorem{lemma}{Lemma}

\RequirePackage{tikz}
\RequirePackage{xcolor}
\RequirePackage{calc}

\makeatletter

\definecolor{secmsghue}{HTML}{2E7EBB}
\definecolor{secmsgink}{HTML}{12558E}
\definecolor{secmsgtext}{HTML}{1A1A1A}

\newcommand{\secmsgfill}{secmsghue!11.5!white}
\newcommand{\secmsgedge}{secmsghue!45!white}

\newlength{\secmsgpadx}    
\newlength{\secmsgpady}    
\newlength{\secmsgrule}    
\newlength{\secmsgradius}  
\newlength{\secmsgabove}   
\newlength{\secmsgbelow}   
\newlength{\secmsgparskip} 

\newcommand{\secmsglabelfont}{\bfseries\color{secmsgink}}
\newcommand{\secmsgbodyfont}{\normalfont\color{secmsgtext}}
\newcommand{\secmsgtagsep}{:}
\newcommand{\secmsgdefaultlabel}{Message of Section~\thesection}

\newsavebox{\secmsg@body}
\newlength{\secmsg@txtw}

\newcommand{\secmsg@openlist}{%
  \list{}{%
    \setlength{\labelwidth}{0pt}%
    \setlength{\labelsep}{0pt}%
    \setlength{\leftmargin}{0pt}%
    \setlength{\rightmargin}{0pt}%
    \setlength{\itemindent}{0pt}%
    \setlength{\listparindent}{0pt}%
    \setlength{\topsep}{0pt}%
    \setlength{\partopsep}{0pt}%
    \setlength{\itemsep}{0pt}%
    \setlength{\parsep}{\secmsgparskip}}%
  \item[]%
}

\newenvironment{sectionmessage}[1][\secmsgdefaultlabel]{%
  \par\addvspace{\secmsgabove}%
  \setlength{\secmsg@txtw}{\linewidth-\secmsgrule-2\secmsgpadx}%
  \begin{lrbox}{\secmsg@body}%
  \begin{minipage}{\secmsg@txtw}%
  \secmsgbodyfont
  \setlength{\parskip}{0pt}%
  \secmsg@openlist
  {\secmsglabelfont #1\secmsgtagsep}\hskip 0.4em\relax
  \ignorespaces
}{%
  \unskip
  \endlist
  \end{minipage}%
  \end{lrbox}%
  \noindent
  \begin{tikzpicture}
    \node[anchor=south west, draw=\secmsgedge, fill=\secmsgfill,
          line width=\secmsgrule, rounded corners=\secmsgradius,
          inner xsep=\secmsgpadx, inner ysep=\secmsgpady]
      at (0,0) {\usebox{\secmsg@body}};
  \end{tikzpicture}%
  \par\addvspace{\secmsgbelow}%
}

\makeatother

\newcommand{\Sph}{\mathbb{S}}
\newcommand{\M}{\mathcal{M}}
\newcommand{\U}{\mathcal{U}}
\newcommand{\V}{\mathcal{V}}
\newcommand{\X}{\mathcal{X}}
\newcommand{\Y}{\mathcal{Y}}
\newcommand{\EX}{\mathcal{E}_X}
\newcommand{\EY}{\mathcal{E}_Y}

\newcommand{\HH}{\mathcal{H}}
\newcommand{\Jd}{\mathsf{J}}

\newcommand{\CKA}{\mathrm{CKA}}
\newcommand{\aug}{\mathrm{aug}}
\newcommand{\mknn}{m$k$NN}
\newcommand{\mmknn}{\mathrm{m}k\mathrm{NN}}
\newcommand{\hsic}{\mathrm{HSIC}}
\newcommand{\inner}[2]{\langle #1, #2 \rangle}
\newcommand{\innert}{\langle x, x' \rangle}
\newcommand{\innerp}{\langle x, x^+ \rangle}

\newcommand{\oomega}{\frac{1}{\sigma^2}}

\newcommand{\AX}{\hat{A}_X}
\newcommand{\SX}{\hat{S}_X}

\title{Unimodal Augmentation-based SSL Objectives Induce Cross-Modal Alignment}

\author{%
  Zohar Mashiah$^{*}$ \\
  Department of Computer Science and AI\\
  Bar-Ilan University \\
  Ramat-Gan, Israel \\
  \texttt{mashiahzohar4@gmail.com} \\
  \And
  Nir Ben-Ari$^{*}$ \\
  Department of Computer Science and AI\\
  Bar-Ilan University \\
  Ramat-Gan, Israel \\
  \texttt{nirnirba@gmail.com} \\
  \AND
  Ronen Talmon \\
  Electrical and Computer Engineering \\
  Technion \\
  Haifa, Israel \\
  \texttt{ronen@ee.technion.ac.il} \\
  \And
  Uri Shaham \\
  Department of Computer Science and AI\\
  Bar-Ilan University \\
  Ramat-Gan, Israel \\
  \texttt{uri.shaham@biu.ac.il} \\
}

\begin{document}

\maketitle

\renewcommand{\thefootnote}{\fnsymbol{footnote}}
\footnotetext[1]{Equal contribution, random order.}

\begin{abstract}
    Recent works have observed implicit cross-modal alignment between independently trained unimodal models. However, the mechanisms underlying the emergence of this phenomenon remain unresolved. In this paper, we demonstrate that unimodal augmentation-based self-supervised objectives implicitly induce cross-modal alignment. We show that low unimodal losses force high cross-modal alignment, whereas a high unimodal loss prevents such alignment. We theoretically connect the widely used InfoNCE loss to local CKA, the popular measure of cross-modal alignment. We further provide extensive empirical evidence supporting this claim, both in controlled settings and across 53 unconstrained frozen text, vision, and audio unimodal encoders. In both cases, unimodal augmentation-based objectives control cross-modal alignment. Overall, our results provide a theoretical and empirical foundation for understanding the emergence of implicit cross-modal alignment by establishing a direct connection between unimodal training objectives and cross-modal structure.\footnote[2]{Code: \url{https://github.com/shaham-lab/Aug-SSL-Implicit-Alignment}}
\end{abstract}

\section{Introduction}

Recent discoveries in representation learning have provided empirical evidence that independently trained unimodal vision and language models converge toward a unified representation, particularly in terms of their local neighborhood structure \citep{huh2024platonic, groger2026revisiting}. This phenomenon is known as the Platonic Representation Hypothesis (PRH), and has been further supported by extensive subsequent work \citep{li2024vision, acevedo2025quantitative, he2025seeing}, including extensions to additional modalities \citep{duraphe2025platonic, li2025causes, edamadaka2025universally, ngo2024language} and practical applications \citep{yacobi2026learning, yoon2025escaping, zhang2025assessing, groger2026limited, jha2026harnessing, gupta2026better, schnaus2025sa, li2026platonic}.

However, the understanding of the causes underlying this emergent alignment remains limited. Most works either investigate the effect of scaling-related attributes of model size and capacity or the amount of training data on the strength of this phenomenon \citep{umar2026signal, lobashev2025information}, or empirically address this question with explicit cross-modal factors. For example, shared input statistics and architectural biases \citep{kapoor2026bridging, tjandrasuwita2025understanding}, or overlap in datasets and downstream tasks \citep{li2025causes}.

Meanwhile, Augmentation-based Self-Supervised Learning (Aug-SSL) has emerged as a leading training paradigm for training large models \citep{gui2024survey}. Popular models such as DINO for vision \citep{oquab2023dinov2}, Sentence-BERT for text \citep{reimers2019sentence}, and BYOL-A for audio \citep{niizumi2021byol}, have all been trained via Aug-SSL. While its power in unimodal settings is well-studied \citep{wang2020understanding, balestriero2022contrastive}, to the best of our knowledge, its effect on implicit cross-modal alignment has remained unexplored. 

\begin{wrapfigure}{r}{0.46\textwidth}
  \centering
  \vspace{-20pt}
  \includegraphics[width=\linewidth]{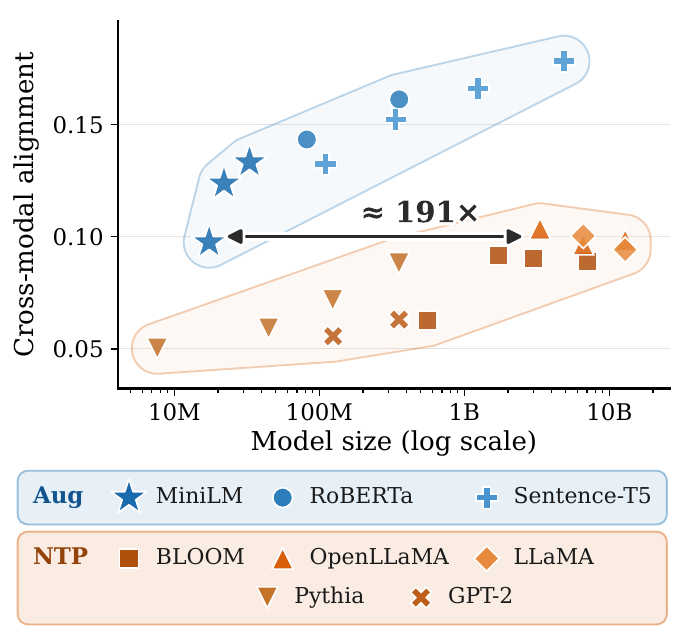}
  \vspace{-20pt}
  \caption{\textbf{Augmentation-based objectives induce cross-modal alignment.} Cross-modal text-vision alignment (mutual $k$-nearest neighbors) as a function of model size (number of parameters) across different text models to DINOv2. Notably, augmentation-based models achieve higher cross-modal alignment for a given model size, and yield superior alignment even with significantly fewer parameters.}
  \label{fig:opening}
  \vspace{-20pt}
\end{wrapfigure}

In this paper, we show that emergent cross-modal alignment (CMA) is a direct consequence of unimodal Aug-SSL objectives. That is, the unimodal objectives implicitly induce cross-modal alignment, even in the absence of explicit cross-modal supervision. The augmentations act as view-specific information removers, filtering out view-specific variation. The encoders then steer each embedding space toward the shared semantic structure underlying the modalities, thereby promoting cross-modal similarity. 
To the best of our knowledge, \emph{this is the first work to explain implicit cross-modal alignment via unimodal properties, connecting them both theoretically and in practice.}

We begin with a theoretical investigation, directly connecting the widely used InfoNCE objective \citep{oord2018representation} to local CKA, a popular measure of CMA \citep{cristianini2001kernel}. We prove that low unimodal Aug-SSL losses force high CMA (Thm.~\ref{thm:align_main}), whereas even a single high unimodal loss restricts the alignment (Thm.~\ref{thm:misalign_main}). These results hold independently of data size, model architecture, and other factors previously proposed to explain the emergence of cross-modal alignment.

We further provide extensive empirical evidence for these results in real-world settings, both in controlled experiments and across large-scale pretrained models and modalities. For example, Fig. \ref{fig:opening} shows that pre-trained text augmentation-based models consistently achieve higher CMA with a vision model than models of comparable size trained with next-token prediction (NTP). Moreover, augmentation-based models reach high CMA with substantially fewer parameters.

\paragraph{Contributions.} Our contributions are summarized as follows:
(1) We show that unimodal self-supervised objectives implicitly drive cross-modal alignment;
(2) We theoretically establish a connection between the unimodal InfoNCE loss and cross-modal alignment as measured by local CKA; and
(3) We empirically demonstrate that the unimodal Aug-SSL loss predicts cross-modal alignment across controlled experiments and large-scale pretrained models.

\section{Related Work}

\paragraph{Implicit Cross-Modal Alignment.} Numerous studies have documented the emergence of implicit CMA in vision-language models \citep{huh2024platonic, li2024vision, acevedo2025quantitative, miller2026testing, fan2025scaling, he2025seeing}. Subsequent work has shown that similar alignment phenomena arise across a broader range of modalities, including video-text \citep{ovsjanikov2026dynamic}, astronomical data \citep{duraphe2025platonic}, and scientific data \citep{li2026platonic, edamadaka2025universally}, among others. Beyond documenting its existence, several studies have examined how CMA depends on properties of the data and training setup, attributing its emergence to factors such as dataset and task overlap \citep{li2025causes} or shared data characteristics \citep{tjandrasuwita2025understanding}. However, these explanations remain empirical and do not establish a theoretical account of why CMA emerges.

\paragraph{Theory of Implicit CMA.} Several recent works have developed theoretical perspectives on implicit cross-modal alignment. \citet{lu2026indra} formalize the alignment hypothesis using category theory. \citet{insulla2025towards} establish a connection between representation alignment and downstream task performance. However, these do not characterize the mechanisms underlying its emergence. Other studies investigate potential mechanisms that may give rise to this phenomenon \citep{bangachev2026representation, ziyin2026neural, ziyin2025proof}; however, their analyses rely on strong linearity assumptions about the data or learned representations, limiting their applicability to the highly nonlinear models used in practice. \citet{lobashev2025information} further assumes that the two modalities share input statistics, imposing an additional restriction on the data-generating process. In contrast, this paper provides a theoretical account of how implicit cross-modal alignment emerges directly from the unimodal training objective in general settings.

\paragraph{Augmentation-based SSL.} Self-supervised learning (SSL) has emerged as a dominant training paradigm across a wide range of domains, including vision \citep{chen2020simple, oquab2023dinov2, he2022masked}, language \citep{devlin2019bert, gao2021simcse, liu2019roberta}, audio \citep{baevski2020wav2vec, hsu2021hubert, saeed2021contrastive, niizumi2021byol}, video \citep{tong2022videomae, bardes2024revisiting}, and scientific data \citep{wang2022molecular}. In particular, augmentation-based SSL provides a simple and effective mechanism for learning from large numbers of self-generated training examples through different transformations of the same underlying samples. Prior works have investigated the properties induced by SSL training within individual modalities \citep{wang2020understanding, balestriero2022contrastive}. Conversely, other works have demonstrated the effectiveness of explicit cross-modal SSL objectives for learning aligned representations \citep{radford2021learning, wu2022wav2clip}. In contrast, this paper considers the implications of SSL on \emph{implicit cross-modal} alignment, without any cross-modal objective.

\section{Theoretical Analysis: From Unimodal InfoNCE to CMA} \label{sec:theory}

This section presents our theoretical results connecting unimodal InfoNCE and cross-modal alignment (CMA). We first formally define the setup, CMA, and unimodal augmentation-based self-supervised learning (Aug-SSL) (Sec. \ref{sec:theory:base}). We then establish their connection through two complementary results (Sec. \ref{sec:theory:results}). In Thm.~\ref{thm:align_main}, we show that low unimodal losses force high CMA. In Thm.~\ref{thm:misalign_main}, we show that a high unimodal loss for even one model restricts CMA, independently of other factors such as model architecture and training data size. Together, these results establish that unimodal augmentation-based training is a fundamental driver of implicit CMA. For brevity, we defer the complete derivations and proofs to App.~\ref{app:proofs}.

\begin{figure}
    \centering
    \vspace{-8pt}
    \includegraphics[width=1\linewidth]{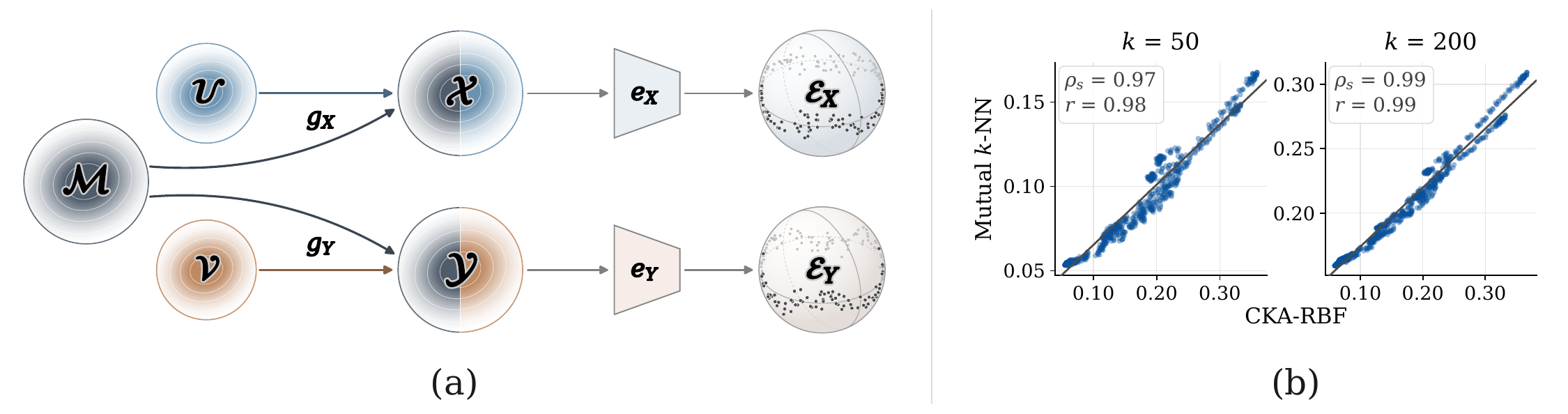}
    \vspace{-16pt}
    \caption{\textbf{(a) Theoretical setup visualization.} Three manifolds $\M$, $\U$, and $\V$ generate two observation spaces $\X$ and $\Y$ with shared semantics $\M$. Two trainable encoders then map the observations to the corresponding embedding spaces $\EX$ and $\EY$. \textbf{(b) Correlation between CKA-RBF and m$\bm{k}$NN.} \mknn\ is highly correlated with CKA-RBF across different choices of $k$ and the corresponding $\sigma$. Both Spearman ($\rho_s$) and Pearson ($r$) correlations are close to 1.}
    \vspace{-16pt}
    \label{fig:theory}
\end{figure}

\subsection{Setup} \label{sec:theory:base}

\paragraph{Notations \textmd{(visualized in Fig. \ref{fig:theory}a)}.} Let $\M,\U,\V$ be manifolds representing shared semantics and view-specific information, respectively. Observations are generated by
$g_X:\M\times\U\to\X$ and $g_Y:\M\times\V\to\Y$,
where $\X\subseteq\R^{d_x}, \Y\subseteq\R^{d_y}$. Two trainable unimodal encoders, $e_X:\X\to\EX, e_Y:\Y\to\EY$, map the observations to normalized embedding spaces $\EX\subseteq\Sph^{d_X-1}, \EY\subseteq\Sph^{d_Y-1}$, respectively.
Throughout, $0 < \tau \leq 1$ denotes the fixed temperature of the training objective.

The phenomenon of implicit CMA dictates that independently trained unimodal encoders naturally develop highly aligned representations. To measure the alignment between the embedding spaces $\EX$ and $\EY$, we use the well-established Centered Kernel Alignment (CKA) metric \citep{kornblith2019similarity}, following recent works on implicit cross-modal alignment \citep{huh2024platonic, groger2026revisiting}. We denote the observations $
x=g_X(m,u), x'=g_X(m',u'),
y=g_Y(m,v), y'=g_Y(m',v')
$. That is, $x$ and $y$ share the same underlying semantics $m$, and likewise $x'$ and $y'$ share $m'$. In practice, this shared semantic structure is obtained through paired data. The CKA is then often defined using the Hilbert-Schmidt Independence Criterion (HSIC), which is 
\begin{equation*}
    \hsic(\EX, \EY) = \E\left[k^c_X\left(e_X(x),e_X(x')\right)\: k^c_Y\left(e_Y(y), e_Y(y') \right)\right]
\end{equation*}
where $k^c$ is the doubly centered kernel $k^c= k -\E[k|x] -\E[k|x'] +\E[k]$.
CKA then measures the correlation between the centered kernels, giving:
\begin{equation}
\CKA(\EX,\EY)=
\frac{\hsic(\EX,\EY)}{\sqrt{\hsic(\EX,\EX)\ \hsic(\EY,\EY)}}
\in[0,1],
\end{equation}

Throughout the paper, we use the CKA-RBF version of CKA \citep{kornblith2019similarity}, with the Gaussian kernel $k(z,z') = e^{-\|z-z'\|^2/2\sigma^2}$, and denote the kernel variances $\sigma_X^2= \hsic(\EX,\EX), \sigma_Y^2=\hsic(\EY,\EY)$. In particular, we assume $\sigma_X, \sigma_Y > 0$ so the CKA is defined.

The Gaussian kernel emphasizes local structure, making this measure particularly suitable for studying the local nature of implicit cross-modal alignment observed in recent work \citep{groger2026revisiting}. Moreover, as shown in Fig. \ref{fig:theory}b, CKA-RBF is strongly correlated with the mutual $k$-nearest-neighbors (\mknn) metric used in recent works, with the kernel bandwidth $\sigma$ playing a role analogous to the neighborhood parameter $k$ (see App. \ref{app:guasscka_mknn}).

\paragraph{Augmentation-based SSL.} We define an \emph{augmentation} as a resampling of both the semantics and the view, while preserving the underlying semantic structure. Specifically:
\begin{equation}
    x^+ := \aug(x)=g_X(m^+,u^+),\qquad m^+\sim P(m,\cdot),\quad u^+\sim\nu_\U
\ \text{independent}
\end{equation}
where $P(m,\cdot):=\Jd(m,\cdot)\,\mu$ is a Markov kernel on $\M$ with density $\Jd$. This definition aligns with prior theoretical studies of augmentation-based SSL \citep{von2021self}, while not restricting the augmentation to exactly preserve the semantics. We exclude the degenerate case $\Jd\equiv1$, in which the augmentation resamples the semantics independently and carries no signal.

An Aug-SSL model is trained to maximize the similarity between augmented samples, while preventing collapse to a constant representation, for instance by contrasting against randomly sampled observations. A popular objective for this purpose is the InfoNCE loss \citep{oord2018representation}, which has been widely adopted in vision \citep{chen2020simple, he2020momentum}, text \citep{reimers2019sentence, gao2021simcse}, audio \citep{niizumi2021byol}, and other domains \citep{qian2021spatiotemporal, wang2022molecular}. Let $x,x'\in\X$ denote observations and let $x^+=\aug(x)$ be an augmentation of $x$. The $K$-negative InfoNCE loss is
\begin{equation} \label{eq:infonce-k}
    \ell^{(K)}_X=\E\left[-\log\frac{\exp\big(\langle e_X(x),e_X(x^+)\rangle/\tau\big)}{\exp\big(\langle e_X(x),e_X(x^+)\rangle/\tau\big)+\sum_{k=1}^{K}\exp\big(\langle e_X(x),e_X(x_k)\rangle/\tau\big)}\right],
\end{equation}
where $K$ is a hyperparameter controlling the number of negative samples. Following \citet{wang2020understanding} (Thm. 1), we use its population form:
\[
N_X=\lim_{K\to\infty}\big(\ell^{(K)}_X-\log K\big)
=-\,\frac{1}{\tau}\,\E\big[\langle e_X(x),e_X(x^+)\rangle\big]+\E_x\Big[\log \E_{x'}\big[e^{\langle e_X(x),e_X(x')\rangle/\tau}\big]\Big].
\]

For convenience, we normalize the objective as follows, with both ends attainable (Lemma \ref{lem:range}):
\[
L_X=\frac{N_X+1/\tau}{R_\tau} \in [0, 1],
\]
where $R_\tau=\frac{2}{\tau}+\log\cosh\frac1\tau$. $L_Y$ is analogously defined for $\Y$.

\subsection{Connecting InfoNCE and local CKA} \label{sec:theory:results}

We now present the theoretical results that formalize the connection between the unimodal Aug-SSL losses (InfoNCE) and CMA under the setup above. The two results establish complementary directions of this relationship. Specifically, Thm.~\ref{thm:align_main} shows that low losses in both modalities guarantee high CMA, while Thm.~\ref{thm:misalign_main} shows that high loss in either modality prevents alignment.

\paragraph{Implicit alignment.} The bound in Thm. \ref{thm:align_main} utilizes the concentration strength of the augmentations, $\lVert J-1\rVert_{L^2}=:\chi$, where $J(m,m^+)$ is the likelihood ratio of $m^+$ being an augmentation of $m$ versus an independent sample. Thus, $\chi$ quantifies the semantic information retained by augmentations: $\chi=0$ when $J\equiv1$ and augmentations ignore the original semantics, and $\chi=\infty$ when they preserve them exactly. We consider the practical regime $0<\chi<\infty$, where augmentations alter but retain semantics. Together with the InfoNCE temperature $\tau$ and CKA-RBF scale $\sigma$, this allows us to lower-bound CKA in terms of the unimodal losses.

\begin{theorem}[Implicit alignment]\label{thm:align_main}
\[
\CKA(\EX,\EY)\ \ge\ f(L_X, L_Y \mid \tau, \sigma, \chi),
\]
with $f$ non-increasing in both $L_X$ and $L_Y$, and converging to $1$ as both losses approach $0$.
\end{theorem}

This theorem establishes that when both \textbf{\emph{unimodal}} Aug-SSL losses ($L_{X}$ and $L_{Y}$) are effectively minimized, a high degree of cross-modal alignment is guaranteed. Importantly, the bound depends only on the attained loss values, not on the training procedure.
The monotonic convergence of $f$ to $1$ strictly forces the CKA toward its maximum value when the losses are sufficiently small. Specifically,
$$f(L_X, L_Y) = \cos(\theta_X + \theta_Y), \quad \theta(L) = \operatorname{clamp}\big( \arccos\big( \tfrac{a_{\sigma}-b_{\sigma,\tau}L}{\chi\sqrt{D_{\sigma,\tau} L}} \big),\ 0,\ \tfrac{\pi}{2} \big),$$
where $\operatorname{clamp}$ restricts $\theta(L)$ to $[0, \tfrac{\pi}{2}]$, and $\theta_X = \theta(L_X), \theta_Y = \theta(L_Y)$.
In particular, $f$ depends on the augmentations concentration strength $\chi$, and on the InfoNCE temperature $\tau$ and the CKA-RBF scale $\sigma$ through the hyperparameter-dependent constants: $a_{\sigma}, b_{\sigma,\tau}$ and $D_{\sigma,\tau}$. For brevity, their exact definitions are deferred to App. \ref{app:proofs:constants}.

Intuitively, augmentations remove view-specific information, filtering out variations that are not shared across modalities. Low loss then encourages augmented samples to cluster together, guiding each embedding space toward the shared semantic structure underlying the modalities. Thereby, promoting cross-modal similarity, i.e., high CKA.

\paragraph{Implicit misalignment.} While the above highlights the geometric convergence driven by successful unimodal optimization, evaluating the implications of high unimodal loss requires the complementary perspective addressed in Thm. \ref{thm:misalign_main}. For this, we assume the unimodal deep SSL encoders $e_X, e_Y$ exhibit Lipschitz continuity. That is, $\lVert e_X(x)-e_X(x')\rVert\le\ell_X\,d_\X(x,x')$ for all $x,x'\in\X$, with an analogous bound $\ell_Y$ for $e_Y$. This is consistent with recent literature, proving both empirically and theoretically that this happens in practice \citep{virmaux2018lipschitz, lee2021compressive}. This theorem utilizes an augmentation-dependent constant, the semantic displacement:
$$\varepsilon_X = \E_{\pi,u}\Big[d_\X\big(g_X(m,u),g_X(m^+,u)\big)^2\Big],$$
where $\pi(m,m^+)=J(m,m^+)\mu(m)\mu(m^+)$ is the joint law of the semantic pair.
Intuitively, $\varepsilon_X$ measures the extent to which the semantic variations caused by the augmentation process are manifested in the observation space.

\begin{theorem}[Implicit misalignment]\label{thm:misalign_main}
If the encoder $e_X$ is Lipschitz with a constant $\ell_X$, and $\tau\le\sigma^2$, then
\[
\CKA(\EX,\EY)\ \le\ g(L_X \mid \tau, \sigma, \ell_X, \varepsilon_X),
\]
where $g$ is strictly decreasing in $L_X$. This holds symmetrically for $L_Y$ with $\ell_Y, \varepsilon_Y$.
\end{theorem}

Thm. \ref{thm:misalign_main} shows that a sufficiently high \textbf{\emph{unimodal}} loss strictly upper-bounds CMA, preventing convergence. In particular, for practical hyperparameters, the bound remains bounded away from $1$ as $L_X$ approaches its maximum (App. \ref{app:theory_bound}). Consequently, a poorly optimized encoder strictly precludes the emergence of shared geometric structure, regardless of the optimization quality in the counterpart modality. The condition $\tau\le\sigma^{2}$ is consistent with standard InfoNCE practice, where $\tau$ is a small fraction and $\sigma^{2}$ is typically larger for normalized embeddings.

The bound is given by
$$g(L_X) = \tfrac{d_{\sigma, \tau} + \ell_X^2\varepsilon_X}{2\sigma^2\sigma_X} + \tfrac{h_{\sigma, \tau}}{\sigma_X}(1-L_X).$$
In particular, $g$ depends on the Lipschitz constant $\ell_X$, the semantic displacement $\varepsilon_X$, and on the InfoNCE temperature $\tau$ and the CKA-RBF scale $\sigma$ via the hyperparameter-dependent constants: $d_{\sigma, \tau}$ and $h_{\sigma, \tau}$ (App. \ref{app:proofs:constants}).

Intuitively, only semantic variation can align across modalities, so CKA is bounded by the fraction of embedding variation that is semantic. As the unimodal loss increases, less variation reflects semantics, limiting the achievable CKA.
Both bounds are visualized in App. \ref{app:theory_bound}.

Together, these theoretical results provide a rigorous mathematical foundation, showing that implicit cross-modal alignment is a geometric consequence of the unimodal InfoNCE objective. In the next section, we empirically demonstrate these predictions, showing that Aug-SSL models exhibit higher CMA than their counterparts.

\begin{sectionmessage}
    Low unimodal augmentation-based SSL losses induce cross-modal convergence, while a high loss in either modality forces misalignment.
\end{sectionmessage}

\section{Aug-SSL is Key and Predictor of CMA in Practice} \label{sec:results}

In this section, we empirically validate the core premise of this paper. We consider both controlled experiments (Sec. \ref{sec:results:controlled}) and frozen pretrained models (Sec. \ref{sec:results:pretrained}), providing evidence that unimodal augmentation-based objectives play a central role in the emergence of implicit CMA. Further details concerning the experimental setup and implementation are provided in App. \ref{app:implementation_details}.

\begin{figure}
    \centering
    \vspace{-19pt}
    \includegraphics[width=\linewidth]{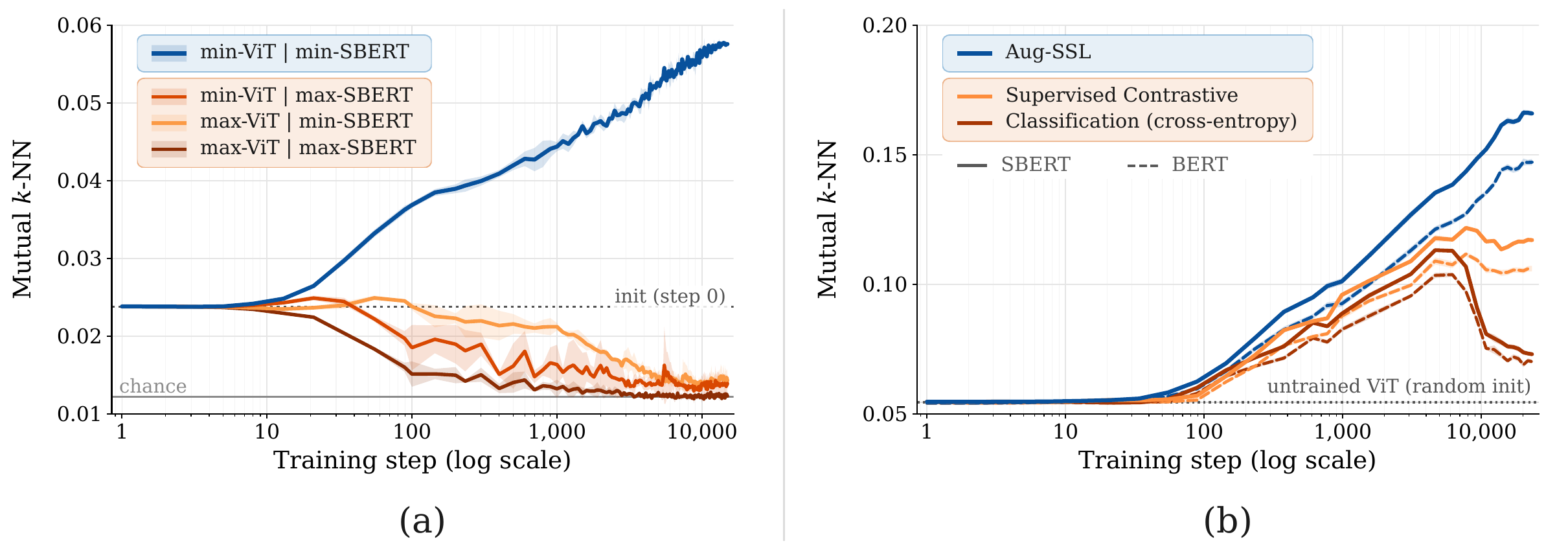}
    \vspace{-21pt}
    \caption{\textbf{(a) Empirical validation of Thm. \ref{thm:align_main} and \ref{thm:misalign_main}.} Vision (ViT) and text (SBERT) models trained to minimize/maximize the Aug-SSL loss (Eq. \ref{eq:infonce-k}). Minimizing both unimodal losses improves CMA (Thm. \ref{thm:align_main}), whereas maximizing either loss prevents alignment, with CMA converging to the chance level (Thm. \ref{thm:misalign_main}). \textbf{(b) Augmentations vs. Supervision.} CMA throughout training of a vision model (ViT) against frozen text models: an augmentation-based model (SBERT) and a non-augmentation-based model (BERT). Training ViT with augmentations achieves substantially higher CMA, including compared to the BERT model.}
    \vspace{-16pt}
    \label{fig:controlled_results}
\end{figure}

\paragraph{Datasets.} We utilize four benchmark datasets. Following \citet{huh2024platonic} and \citet{groger2026revisiting}, we use the Wikipedia image-text (WIT) dataset \citep{srinivasan2021wit}. We additionally use the MS-COCO image-text benchmark \citep{lin2014microsoft}, which provides semantic labels that allow us to more closely examine the effect of augmentations on cross-modal alignment. To further validate the phenomenon across modalities, we use the audio-text CLOTHO \citep{drossos2020clotho} and AudioCaps \citep{kim2019audiocaps} benchmarks.

\paragraph{Evaluating Cross-Modal Alignment.} Following \citet{groger2026revisiting}, we evaluate CMA using the mutual $k$-nearest neighbors (\mknn) metric. Given a set of paired observations $\mathcal{S}={(x_1,y_1),\ldots,(x_n,y_n)}\in\EX\times\EY$, let $N_X^k(x)$ and $N_Y^k(y)$ denote the sets of the $k$ nearest neighbors of $x$ and $y$ within their respective embedding spaces. We define
\begin{equation}
\mmknn(\mathcal{S}) =
\frac{1}{n}\sum_{i=1}^n
\frac{\left|N_X^k(x_i)\cap N_Y^k(y_i)\right|}{k}.
\end{equation}

As shown in Fig. \ref{fig:theory}b, \mknn\ can be viewed as a discrete approximation of the CKA-RBF considered in Sec. \ref{sec:theory}, making it a natural choice for evaluating CMA in our experiments. Unless stated otherwise, we use $k=50$. Additional results using different values of $k$, and CKNNA are provided in App. \ref{app:more_cma_metrics}, and exhibit the same qualitative trends.

\subsection{Direct Validation of the Theory} 
\label{sec:results:controlled}
In this section, we isolate the effect of unimodal Aug-SSL training by controlling the model architectures and training setup. We train vision and text encoders from scratch using a ViT \citep{dosovitskiy2020image} for vision and an SBERT architecture \citep{reimers2019sentence} for text, initialized from a pretrained BERT model \citep{devlin2019bert}.

\paragraph{Isolating the InfoNCE effect.} To empirically validate Thm. \ref{thm:align_main} and \ref{thm:misalign_main}, we train two vision and two text encoders on COCO as described above, with one model per modality minimizing and the other maximizing the unimodal Aug-SSL loss. We use the InfoNCE objective from Sec. \ref{sec:theory}. For vision, we employ standard image augmentations (e.g., random crops, horizontal flips). For text, we use an LLM (Qwen2.5 \citep{qwen2025qwen25technicalreport}) to generate five rephrased versions of each text.

Fig. \ref{fig:controlled_results}a presents the resulting CMA for all pairings of the four encoders. The pairing of the two loss-minimized encoders corresponds to Thm.~\ref{thm:align_main}: minimizing the unimodal Aug-SSL objectives leads to increasing CMA. In contrast, every pairing involving a loss-maximized encoder corresponds to Thm.~\ref{thm:misalign_main}, demonstrating that a high unimodal loss in either modality prevents cross-modal alignment, with CMA approaching the chance level. Training on WIT shows the same trends (App. \ref{app:other_llm}).

\paragraph{Aug-SSL Outperforms Explicit Supervision.} To isolate the contribution of unimodal augmentations to CMA, we compare augmentation-based SSL with supervised learning on the COCO dataset. Specifically, we consider two supervised objectives: Supervised Contrastive Learning (SupCon) \citep{khosla2020supervised} and cross-entropy classification, trained using the labels in COCO. In these experiments, we freeze the pretrained text encoder (BERT/SBERT) and train only the vision encoder.

\begin{figure}
    \centering
    \vspace{-10pt}
    \includegraphics[width=0.98\linewidth]{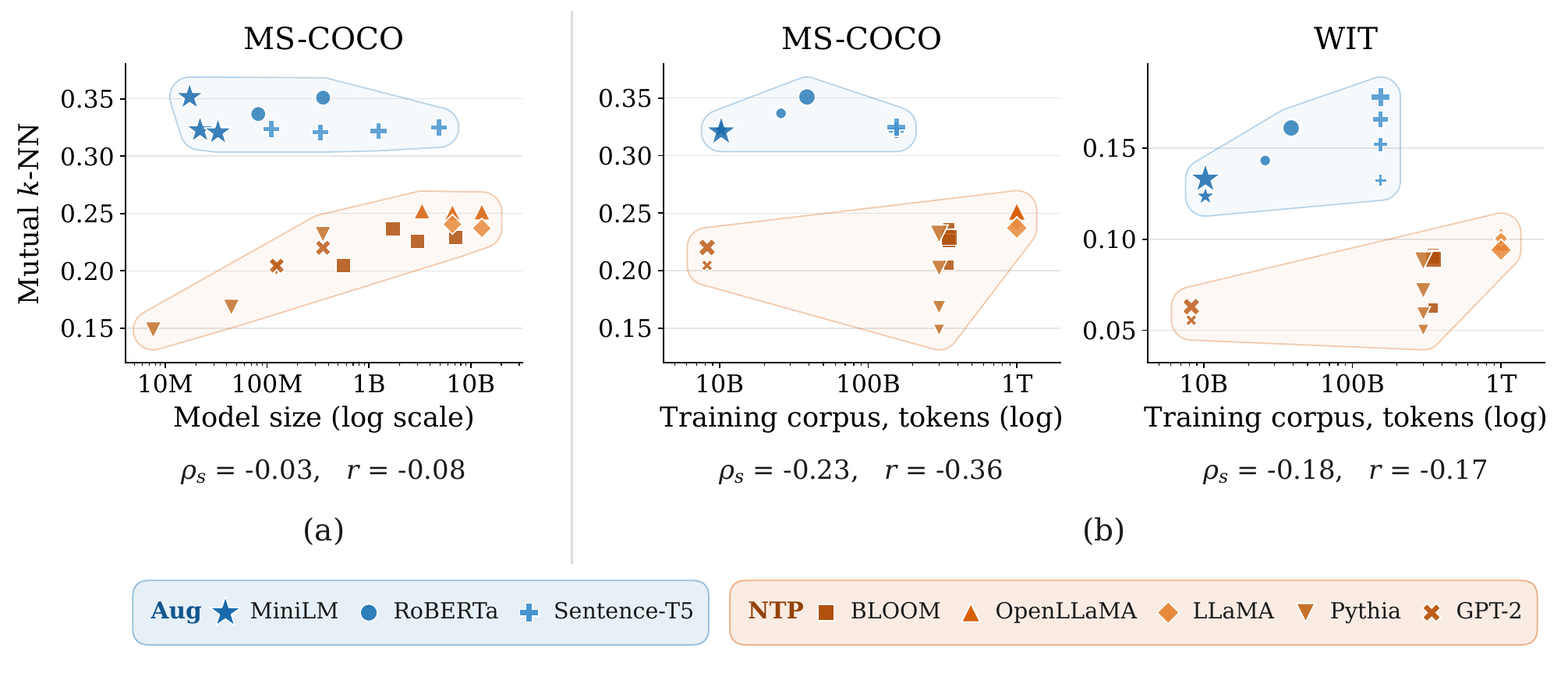}
    \vspace{-6pt}
    \caption{\textbf{Alignment by model/corpus size.} CMA between text models and the DINOv2 vision encoder, using augmentation-based (Aug) and next-token prediction-based (NTP) models. Shown as a function of \textbf{(a) model size} (number of parameters), and \textbf{(b) corpus size} (amount of tokens the model was trained on). Notably, Aug models consistently achieve higher CMA, even at substantially smaller model and corpus sizes. Also, considering all models jointly, model/corpus sizes are uncorrelated with CMA, indicated by both weak Spearman ($\rho_s$) and Pearson ($r$) correlations.}
    \label{fig:pretrained_uncorr}
    \vspace{-10pt}
\end{figure}

Fig. \ref{fig:controlled_results}b presents the resulting CMA. Most notably, the Aug-SSL-trained vision encoder achieves the highest CMA with both text encoders. Its alignment with SBERT, which is an augmentation-based model, is consistently stronger than with BERT, further supporting that augmentation-based training promotes alignment. Importantly, however, the Aug-SSL vision encoder also achieves stronger CMA with BERT than the supervised baselines. Thus, the effect cannot be attributed simply to matching the training paradigm across modalities: augmentation-based training promotes CMA even with a text encoder trained using a different objective. These results further support the role of augmentations as a driver of implicit CMA.

\subsection{Universal Emergence in Foundation Models} \label{sec:results:pretrained}
We extend our analysis to unconstrained, frozen pre-trained models from diverse model kinds and modalities. We show that augmentations implicitly induce CMA across architectures, training datasets, and model scales.

\paragraph{Models.} We consider 53 pretrained unimodal encoders across text, vision, and audio. Specifically, we use eight families of 24 text encoders from two training paradigms: augmentation-based models, including MiniLM \citep{wang2020minilm}, RoBERTa \citep{reimers2019sentence}, and Sentence-T5 \citep{ni2022sentence}; and next-token prediction (NTP)-based models, including BLOOM \citep{le2022bloom}, OpenLLaMA \citep{openlm2023openllama}, LLaMA \citep{touvron2023llamaopenefficientfoundation}, Pythia \citep{biderman2023pythia}, and GPT-2 \citep{radford2019language}. Also, we consider four families of 14 vision encoders: DINOv2 \citep{oquab2023dinov2}, MAE \citep{he2022masked}, ImageNet21K \citep{russakovsky2015imagenet}, and CLIP \citep{radford2021learning}. For each family, we evaluate multiple models of different sizes. We compare the text models with the DINOv2 vision encoder and the BYOL-A v2 audio encoder \citep{niizumi2022byol}, with the other vision encoders and a total of 15 audio encoders shown in App. \ref{app:more_vision_encoders} and App. \ref{app:more_audio_encoders}. Further details regarding the models are provided in App. \ref{app:implementation_details:models}.

\paragraph{Alignment vs. model/corpus size.} Fig. \ref{fig:pretrained_uncorr}a (and \ref{fig:opening}) shows the CMA of text pre-trained models to DINOv2 as a function of model size. Similarly, Fig. \ref{fig:pretrained_uncorr}b shows it as a function of the corpus size the model was trained on. Interestingly, augmentation-based models consistently achieve higher CMA than NTP-based models of comparable size, and often outperform even substantially larger NTP models, trained with significantly more data. This provides evidence that unimodal augmentations play a key role in inducing CMA. Indicated by both Pearson and Spearman correlations, when considering different model families jointly, model or corpus sizes are uncorrelated with CMA.

\begin{figure}
    \centering
    \includegraphics[width=\linewidth]{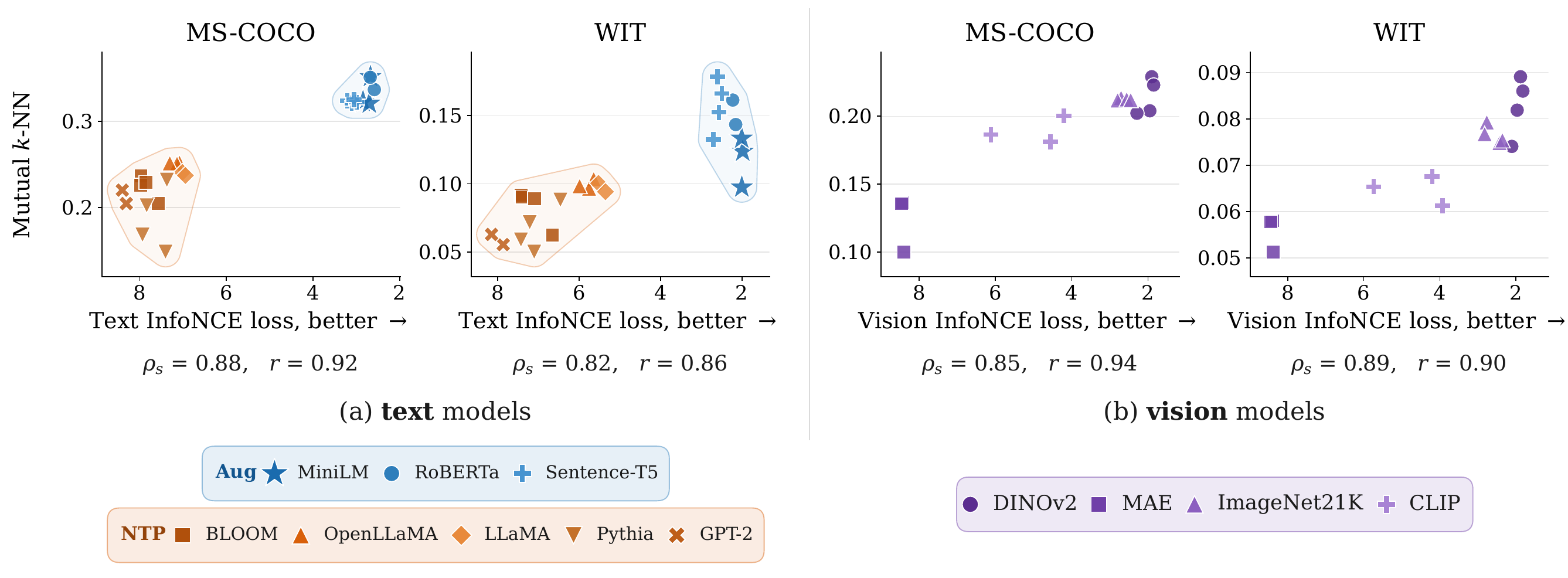}
    \vspace{-15pt}
    \caption{\textbf{Alignment by unimodal loss.} Image-text CMA between text models and DINOv2, and vision models and BLOOM as a function of unimodal \textbf{(a) text InfoNCE loss}, and \textbf{(b) image InfoNCE loss}. In contrast to model or corpus size, the unimodal Aug-SSL loss serves as a robust predictor of CMA. It cleanly separates low- and high-alignment models, exhibiting strong Spearman ($\rho_s$) and Pearson ($r$) correlations.}
    \label{fig:pretrained_loss}
    \vspace{-10pt}
\end{figure}

\paragraph{Unimodal InfoNCE predicts CMA.} Fig. \ref{fig:pretrained_loss} and Tab. \ref{tab:vision_corr} show the CMA as a function of the unimodal InfoNCE loss. We apply the same standard image and LLM-based text augmentations described in Sec. \ref{sec:results:controlled} post-hoc to the frozen pre-trained models and evaluate their unimodal InfoNCE loss. Interestingly, lower \emph{unimodal InfoNCE} loss is highly correlated with higher CMA, indicated by both high Spearman and Pearson correlations. This further validates the theory presented in Sec. \ref{sec:theory}, showing that InfoNCE can serve as a unimodal predictor of cross-modal alignment.

\definecolor{pythaugtint}{HTML}{2E7EBB}
\begin{table}[t]
  \centering
  \caption{\textbf{Image-text CMA is highly correlated with unimodal InfoNCE loss.} Spearman ($\rho_s$) and Pearson ($r$) correlations between \mknn\ and unimodal properties over 24/14 text/vision models. Results are mean $\pm$ std over all second modality models. Notably, CMA is consistently highly correlated with the negative text and vision unimodal InfoNCE loss.}
  \vspace{2pt}
  \label{tab:vision_corr}
  \begin{adjustbox}{max width=\linewidth}
    \begin{tabular}{lcccc}
      \toprule
      & \multicolumn{2}{c}{MS-COCO} & \multicolumn{2}{c}{WIT} \\
      \cmidrule(lr){2-3} \cmidrule(lr){4-5}
      ${\rho_s} \,/\, {r}$ & Text models & Vision models & Text models & Vision models \\
      \midrule
      Model size & ${-0.05_{\pm 0.02}} \,/\, {-0.14_{\pm 0.06}}$ & ${-0.23_{\pm 0.06}} \,/\, {-0.24_{\pm 0.01}}$ & ${\phantom{-}0.09_{\pm 0.02}} \,/\, {\phantom{-}0.08_{\pm 0.04}}$ & ${-0.25_{\pm 0.12}} \,/\, {-0.28_{\pm 0.10}}$ \\
      Training-corpus size & ${-0.32_{\pm 0.04}} \,/\, {-0.43_{\pm 0.04}}$ & ${\phantom{-}0.20_{\pm 0.05}} \,/\, {\phantom{-}0.58_{\pm 0.03}}$ & ${-0.18_{\pm 0.01}} \,/\, {-0.21_{\pm 0.03}}$ & ${\phantom{-}0.34_{\pm 0.06}} \,/\, {\phantom{-}0.45_{\pm 0.16}}$ \\
      \rowcolor{pythaugtint!11.5}InfoNCE loss ($-\mathcal{L}$) & ${\phantom{\mathbf{-}}\mathbf{0.85}_{\pm 0.03}} \,/\, {\phantom{\mathbf{-}}\mathbf{0.94}_{\pm 0.03}}$ & ${\phantom{\mathbf{-}}\mathbf{0.70}_{\pm 0.14}} \,/\, {\phantom{\mathbf{-}}\mathbf{0.92}_{\pm 0.02}}$ & ${\phantom{\mathbf{-}}\mathbf{0.84}_{\pm 0.02}} \,/\, {\phantom{\mathbf{-}}\mathbf{0.88}_{\pm 0.02}}$ & ${\phantom{\mathbf{-}}\mathbf{0.86}_{\pm 0.09}} \,/\, {\phantom{\mathbf{-}}\mathbf{0.90}_{\pm 0.03}}$ \\
      \bottomrule
    \end{tabular}
  \end{adjustbox}
\end{table}

\begin{wraptable}{r}{0.54\textwidth}
  \centering
  \vspace{-8pt}
  \caption{\textbf{Audio-text CMA is highly correlated with unimodal InfoNCE loss.} Spearman ($\rho_s$) and Pearson ($r$) correlations between \mknn\ and text unimodal properties over 24 text models, averaged $\pm$ std over 15 audio models. Notably, CMA is consistently highly correlated with the negative unimodal InfoNCE loss.}
  \vspace{2pt}
  \label{tab:audio_corr}
  \begin{adjustbox}{max width=\linewidth}
    \begin{tabular}{lcc}
      \toprule
      ${\rho_s} \,/\, {r}$ & Clotho & AudioCaps \\
      \midrule
      Model size & ${\phantom{-}0.05_{\pm 0.02}} \,/\, {\phantom{-}0.18_{\pm 0.06}}$ & ${-0.08_{\pm 0.05}} \,/\, {-0.21_{\pm 0.03}}$ \\
      Training-corpus size & ${-0.19_{\pm 0.01}} \,/\, {-0.10_{\pm 0.04}}$ & ${-0.22_{\pm 0.04}} \,/\, {-0.30_{\pm 0.04}}$ \\
      \rowcolor{pythaugtint!11.5}InfoNCE loss ($-\mathcal{L}$) & ${\phantom{\mathbf{-}}\mathbf{0.92}_{\pm 0.01}} \,/\, {\phantom{\mathbf{-}}\mathbf{0.85}_{\pm 0.03}}$ & ${\phantom{\mathbf{-}}\mathbf{0.80}_{\pm 0.03}} \,/\, {\phantom{\mathbf{-}}\mathbf{0.97}_{\pm 0.01}}$ \\
      \bottomrule
    \end{tabular}
  \end{adjustbox}
  \vspace{-15pt}
\end{wraptable}

\begin{figure}
    \centering
    \vspace{-21pt}
    \includegraphics[width=0.98\linewidth]{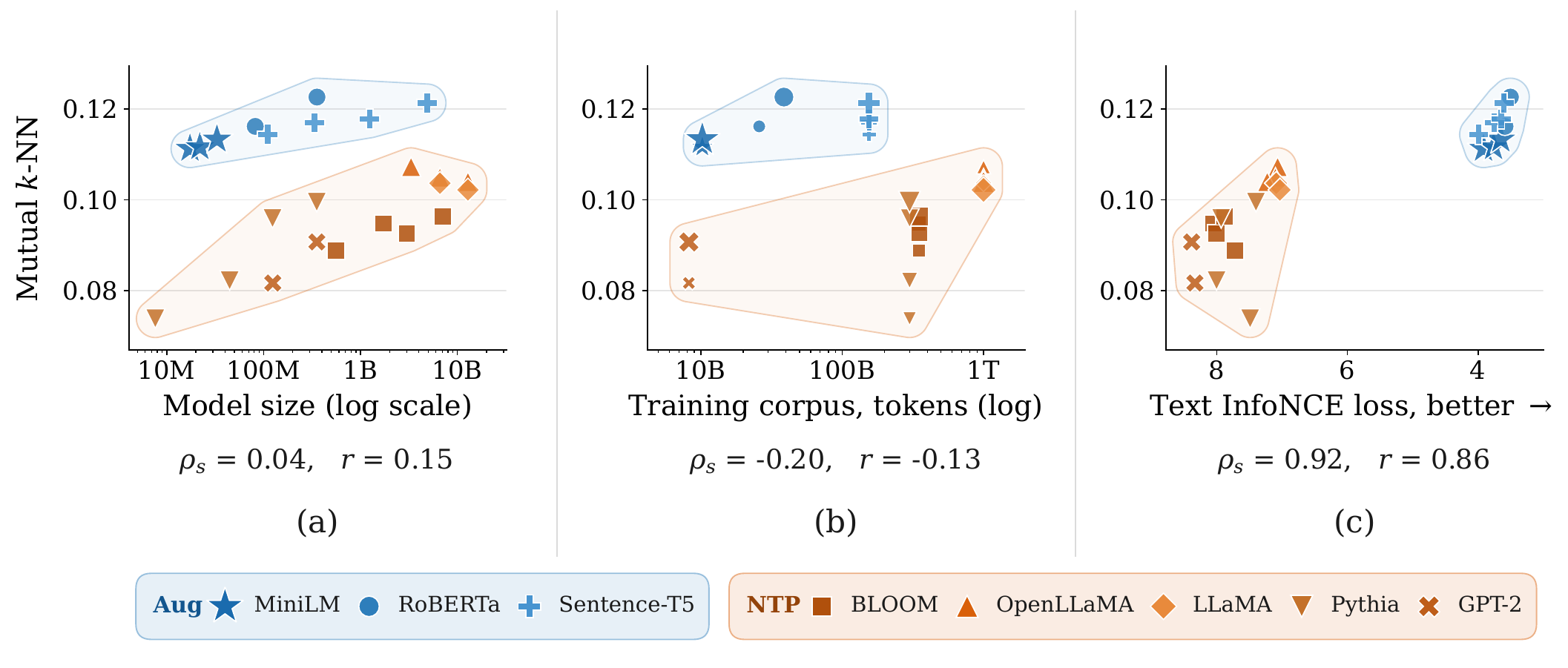}
    \vspace{-10pt}
    \caption{\textbf{Audio-text results.} CMA between text models and the BYOL-A audio encoder as a function of \textbf{(a) model size} (number of parameters), \textbf{(b) corpus size} (amount of tokens the model was trained on), and \textbf{(c) unimodal text InfoNCE loss}. Notably, Aug models consistently achieve higher CMA, even at substantially smaller model and corpus sizes. Also, considering all models jointly, model/corpus sizes are uncorrelated with CMA, while the unimodal loss is highly correlated, indicated by both Spearman ($\rho_s$) and Pearson ($r$) correlations.}
    \vspace{-10pt}
    \label{fig:pretrained_audio}
\end{figure}

\paragraph{Beyond image-text.} Fig. \ref{fig:pretrained_audio} and Tab. \ref{tab:audio_corr} show that the same trends observed in vision-text hold robustly to audio-text. Specifically, model/corpus sizes are uncorrelated with CMA, showing augmentation-based models experience higher alignment nonetheless. Also, the unimodal InfoNCE loss is a strong predictor of the cross-modal alignment. This provides further evidence for the universality of these findings across modalities.

\paragraph{Ablations.} Due to lack of space, we defer ablations to the appendix. Specifically, we show the results are consistent across different kinds of text augmentations (App. \ref{app:other_llm}), other vision encoders (App. \ref{app:more_vision_encoders}), other audio encoders (App. \ref{app:more_audio_encoders}), other CMA metrics (App. \ref{app:more_cma_metrics}), different test set sizes (App. \ref{app:test_size}), and different choices of temperature $\tau$ (App. \ref{app:more_tau}).

\begin{sectionmessage}
    Across controlled training, frozen pre-trained encoders, and modalities, unimodal augmentation-based SSL loss predicts cross-modal alignment.
\end{sectionmessage}

\section{Conclusions and Summary}
In this paper, we presented an explanation for the emergence of implicit cross-modal alignment through unimodal augmentation-based objectives. We established theoretical connections between the widely used InfoNCE loss and cross-modal alignment, as measured by local CKA. In particular, we showed that minimizing Aug-SSL unimodal objectives promotes high cross-modal alignment (Thm. \ref{thm:align_main}), while sufficiently high unimodal loss restricts cross-modal alignment (Thm. \ref{thm:misalign_main}). We then empirically validated these results in both controlled experiments (Sec. \ref{sec:results:controlled}) and unconstrained pre-trained models (Sec. \ref{sec:results:pretrained}), demonstrating that unimodal augmentations are a strong driving force behind cross-modal alignment.

We believe this work provides a foundation for understanding the emergence of CMA from both theoretical and empirical perspectives. While Aug-SSL is commonly motivated by its ability to leverage large-scale data, our results reveal an additional, unexplored benefit: augmentation-based objectives can induce cross-modal alignment implicitly, and much more efficiently, without explicitly incorporating any cross-modal information into the training objective.

\bibliography{iclr2027_conference}
\bibliographystyle{iclr2027_conference}

\appendix
\section{Proofs} \label{app:proofs}

\subsection{Rationale}\label{sec:rationale}

This section gives an informal overview of the argument: the objects it involves, the role played by each lemma, and how they chain into the two theorems. The formal settings are given in Sec. \ref{sec:app:setting}, and the formal statements and proofs follow in Sec. \ref{sec:unimodal} and
Sec. \ref{sec:mis}.
 
\paragraph{Overview.} The centered kernel $k_X^c$ (Eq.~\ref{eq:centred}) lies in $L^2$ with norm $\sigma_X$, and Eq.~\ref{eq:cka} expresses CKA as the cosine of the angle between $k_X^c$ and $k_Y^c$. Since the encoders never interact directly, alignment arises through their shared semantic structure: each encoder is trained against its own augmentations, which act as view-specific information removers by filtering out view-specific variation. Minimizing the unimodal losses therefore steers each embedding space toward the shared semantics underlying the modalities, promoting cross-modal similarity. To formalize this connection, we introduce the augmentation gap $\Delta_X$, which measures the difference between a sample's similarity to its augmentations and its similarity to random samples. The proofs of both theorems follow chains of inequalities, bounding the loss with the gap, and the gap with the CKA.
 
\paragraph{Theorem \ref{thm:align}: low losses force alignment.} To prove Thm. \ref{thm:align}, we bound the angle between each kernel and the shared semantics separately, quantifying the fraction of the encoder structure explained by the shared semantics, and then combine the two bounds via the triangle inequality. In details, Lemma \ref{lem:small} involves the basic components underlying both loss and the CKA: the inner products between a sample and its augmentations, and between a sample and independently sampled points, both of which are bounded in terms of the loss. Lemma \ref{lem:coef} then lower-bounds the augmentation gap by the loss, while Lemma~\ref{lem:anchor} relates the augmentation gap to the fraction of the encoder structure explained by the semantics (induced by the augmentations). Lemma \ref{lem:var} shows that the kernel variance, which appears in the denominator of CKA, is upper-bounded in terms of the loss. Finally,
Lemma \ref{lem:cshape} and Prop. \ref{prop:bound_shape} establish that the bound in Thm. \ref{thm:align} is non-increasing in both losses and attains $1$ when both reach their minima.
 
\paragraph{Theorem \ref{thm:mis}: a high loss prevents alignment.} To prove Thm. \ref{thm:mis}, we bound the fraction of each encoder's structure that is able to carry cross-modal alignment at all, and then bound that fraction by the loss. In details, Lemma \ref{lem:pyth} shows that conditioned on the shared semantics the two embedding spaces are independent, so only the semantic part of each kernel contributes to the CKA. It therefore upper-bounds the CKA by the semantic variance $\beta_X$, measuring how much the representation varies with the semantics, normalized by the kernel variance. Lemma \ref{lem:lip} then relates the semantic variance to the augmentation gap, up to a term measuring how much the augmentations displace the observations, which is where the Lipschitz assumption (Assumption \ref{ass:lip}) is used. Lemma \ref{lem:large} completes the chain by upper-bounding the augmentation gap in terms of the loss, complementing Lemma \ref{lem:coef}. Finally, Prop. \ref{prop:mis_bound_shape} establishes that the bound in Thm. \ref{thm:mis} is strictly decreasing in the loss, and since it involves a single modality, a high loss in either one alone caps the CKA.

\subsection{Setting}\label{sec:app:setting}

Every statement made for $X$ holds verbatim for $Y$.

\paragraph{Data.} We fix a probability measure $\mu$ on the semantic manifold $\M$, and
nuisance laws $\nu_\U$ on $\U$ and $\nu_\V$ on $\V$. Observations are $g_X(m,u)\in\X$
and $g_Y(m,v)\in\Y$, and the trained encoders send them to the unit spheres of the
embedding spaces $\EX,\EY$,
\[
x=e_X\big(g_X(m,u)\big)\in\EX \subseteq \Sph^{d_X-1},
\qquad
y=e_Y\big(g_Y(m,v)\big)\in\EY \subseteq \Sph^{d_Y-1}.
\]
Note that for readability, the appendix refers to $x \in \EX$ and $y \in \EY$ as embedding points rather than observations. We write $\rho_X$ for the law of $x$ and $\bar x=\E[x]$. Throughout, $(m,u,v)$ and
$(m',u',v')$ are two independent draws, and all four of $x,x',y,y'$ are built from
them, so that $x,y$ share the semantics $m$ and $x',y'$ share $m'$.

Every object below is a square-integrable
function of the pair $\big((m,u,v),(m',u',v')\big)$. We write $L^2$ for the space of
such functions, with $\inner{\cdot}{\cdot}$ and $\lVert\cdot\rVert$ its inner product
and norm.

\paragraph{Kernel.} Fix $\omega>0$. On the sphere $\lVert x-x'\rVert^2=2-2\innert$, so
the Gaussian kernel of bandwidth $\omega^{-1/2}$ is exactly
\begin{equation}\label{eq:kernel}
k(x,x')=e^{\omega(\innert-1)} .
\end{equation}
We abbreviate $k_X=k(x,x')$ and $k_Y=k(y,y')$, and use $\omega$ rather than a
bandwidth symbol so that nothing collides with the kernel scales $\sigma_X,\sigma_Y$.
The centred kernel and the alignment are
\begin{equation}\label{eq:centred}
k^c_X=k_X-\E[k_X\mid x]-\E[k_X\mid x']+\E[k_X],
\qquad
\sigma_X^2=\E\big[(k^c_X)^2\big],
\end{equation}
\begin{equation}\label{eq:cka}
\CKA(\EX,\EY)=\frac{\hsic(\EX,\EY)}{\sqrt{\hsic(\EX,\EX)\hsic(\EY,\EY)}}
=\frac{\E[k^c_Xk^c_Y]}{\sigma_X\sigma_Y}
=\inner{\tfrac{k^c_X}{\sigma_X}}{\tfrac{k^c_Y}{\sigma_Y}},
\end{equation}
the latter defined whenever $\sigma_X\sigma_Y>0$. Equation \ref{eq:cka} exhibits the
CKA as the cosine of an angle in $L^2$ between two unit vectors; both directions
exploit this.

\paragraph{Three scalars.}
\begin{equation}\label{eq:scalars}
\AX:=\E\big[k(x,x^+)\big],
\qquad
\SX:=\E\big[k(x,x')\big],
\qquad
\Delta_X:=\AX-\SX .
\end{equation}
$\AX$ is how similar a sample looks to its own augmentation, $\SX$ how similar two
unrelated samples look, and the \emph{augmentation gap} $\Delta_X$ the excess of the
former over the latter.

\begin{definition}[Augmentation]\label{def:aug}
An \emph{augmentation} redraws both the semantics and the nuisance, while keeping the
semantics close. That is,
\begin{equation}\label{eq:aug}
x^+=e_X\big(g_X(m^+,u^+)\big),\qquad m^+\sim P(m,\cdot),\quad u^+\sim\nu_\U
\ \text{independent},
\end{equation}
where $P(m,\cdot)=\Jd(m,\cdot)\,\mu$ is a Markov kernel on $\M$ with density $\Jd$.

Write $\pi(m,m^+)=\Jd(m,m^+)\mu(m)\mu(m^+)$ for the joint law of the semantic pair.
\end{definition}

The \emph{concentration strength} of the augmentation is
\begin{equation}\label{eq:chi}
\chi:=\lVert\Jd-1\rVert_{L^2(\mu\otimes\mu)} .
\end{equation}
We assume $\Jd\in L^2(\mu\otimes\mu)$ and $\chi>0$; the degenerate case $\chi=0$ means
$\Jd\equiv1$, i.e.\ the augmentation resamples the semantics independently and carries
no signal. Both marginals of $\pi$ are $\mu$, so $x$ and $x^+$ each have marginal law
$\rho_X$, a fact used repeatedly below.

\paragraph{The objective.} With $0<\tau\le1$ the temperature, the population InfoNCE
loss and its normalisation are
\begin{equation}\label{eq:loss}
N_X=-\frac{A_X}{\tau}+U_X,
\qquad
A_X:=\E\big[\innerp\big],
\qquad
U_X:=\E_x\Big[\log\E_{x'}\big[e^{\innert/\tau}\big]\Big],
\end{equation}
\begin{equation}\label{eq:normloss}
L_X=\frac{N_X+1/\tau}{R},
\qquad
R=\tfrac2\tau+\log\cosh\tfrac1\tau .
\end{equation}
The derivations often use
\begin{equation}\label{eq:kappa}
\tau R=2+\tau\log\cosh\tfrac1\tau\ >\ 2 .
\end{equation}
The constants of Section~\ref{sec:unimodal} are
\begin{equation}\label{eq:constants}
a:=e^{-\omega}\sinh\omega\in\big(0,\tfrac12\big),
\quad
b:=\big(\omega+e^{-\omega}\sinh\omega\big)\tau R,
\quad
D:=\big(1-e^{-\omega}\big)^2\Big(1+\tfrac{4}{\log2}\Big)\tau R .
\end{equation}

\subsection{Implicit Alignment}\label{sec:unimodal}

This section proves the implicit alignment theorem. That is, small unimodal losses restrict the CKA to be large (Thm. \ref{thm:align}). The key is the augmentations, which are accessed via $J-1$, which depends only on the shared semantics and is therefore common to both modalities. Since the CKA is the cosine of the angle between the centered kernels $k^c_X$ and $k^c_Y$ (Eq. \ref{eq:cka}), it suffices to show that each kernel is close in angle to $J-1$. This angle is determined by two quantities: the inner product $\langle k^c_X, J-1\rangle$, measuring how much of $k^c_X$ points in the direction of $J-1$, and the length of $k^c_X$, namely the kernel scale $\sigma_X$. Lemma \ref{lem:anchor} shows that the inner product equals the augmentation gap $\Delta_X$, i.e., how much more similar a sample is to its augmentation than to a random sample. Lemma \ref{lem:coef} shows that a low loss forces this gap to be large, while Lemma \ref{lem:var} shows that it forces $\sigma_X$ to be small, both using the moment bounds of Lemma \ref{lem:small}. Together, a low loss forces $k^c_X$ to align with $J-1$, and likewise for $k^c_Y$. Thm. \ref{thm:align} then combines the two angles via the triangle inequality, forcing the two kernels to align with each other.

\begin{lemma}[Moments]\label{lem:small}
\[
\lVert\bar x\rVert^2\le\tau R\,L_X,
\qquad
\E\big[\innert^2\big]\le\Big(1+\tfrac{4}{\log2}\Big)\tau R\,L_X,
\qquad
A_X\ \ge\ 1-\tau R\,L_X .
\]
\end{lemma}

\begin{proof}
Write $r(x)=\E[\innert\mid x]=\inner{x}{\bar x}$ and $V(x)=\Var(\innert\mid x)$.
Conditionally on $x$, the variable $\delta=\innert-r(x)$ has mean zero and lies in an
interval of length $2$, so Lemma~\ref{lem:mgf} with $\lambda=1/\tau\ge1$ gives
$\log\E[e^{\delta/\tau}\mid x]\ge\tfrac{\log2}{4\tau}V(x)$. Averaging and using
$\E[r(x)]=\lVert\bar x\rVert^2$,
\begin{equation}\label{eq:unifloor}
U_X = \frac{\E[r(x)]}{\tau} + \E\left[\log\E[e^{\delta/\tau}\mid x]\right] \ge\ \frac{\lVert\bar x\rVert^2}{\tau}+\frac{\log2}{4\tau}\,\E[V]\ \ge\ 0 .
\end{equation}
Since $A_X\le1$, Eq. \ref{eq:loss} gives $U_X=N_X+A_X/\tau\le N_X+1/\tau=R\,L_X$. Both
terms on the last inequality of Eq. \ref{eq:unifloor} are nonnegative, so each is at most $RL_X$:
this is the first bound, and $\E[V]\le\tfrac{4}{\log2}\tau R L_X$. The second bound
follows from $\E[\innert^2]=\E[V]+\E[r(x)^2]$ and
$r(x)^2\le\lVert\bar x\rVert^2\le\tau R L_X$. For the third, $U_X\ge0$ in
Eq. \ref{eq:loss} gives $A_X/\tau=U_X-N_X\ge-N_X=\tfrac1\tau-R\,L_X$.
\end{proof}

\begin{lemma}[The loss controls the gap from below]\label{lem:coef}
$\displaystyle \Delta_X\ \ge\ a-b\,L_X$, with $a,b$ as in Eq. \ref{eq:constants}.
\end{lemma}

\begin{proof}
For $\AX$,
\[
\AX=\E\big[e^{\omega(\innerp-1)}\big]\ \ge\ e^{\omega(A_X-1)}
\ \ge\ e^{-\omega\tau R L_X}\ \ge\ 1-\omega\tau R\,L_X ,
\]
where the first inequality is Jensen, then the third bound of
Lemma~\ref{lem:small}, then $e^{-t}\ge1-t$.

For $\SX=e^{-\omega}\E[e^{\omega\innert}]$,
Lemma~\ref{lem:chord}(i) with $\alpha=\omega$, $c = \E[\innert]=\lVert\bar x\rVert^2$
$$\SX \le e^{-\omega}\left( \cosh\omega + \lVert\bar x\rVert^2 \sinh\omega \right) .$$
Applying the first bound of Lemma~\ref{lem:small} gives
$$\SX\ \le\ e^{-\omega}\cosh\omega+e^{-\omega}\tau R L_X\sinh\omega .$$
Subtracting and using
$1-e^{-\omega}\cosh\omega=\tfrac12(1-e^{-2\omega})=e^{-\omega}\sinh\omega$ gives
$\Delta_X\ge a-b\,L_X$.
\end{proof}

\begin{lemma}[Centered kernel and the augmentation gap]\label{lem:anchor}
$\inner{k^c_X}{\Jd-1}=\Delta_X$, and $\lVert\Jd-1\rVert=\chi$.
\end{lemma}

\begin{proof}
$\Jd-1$ is a function of $(m,m')$ alone, so for any $f\in L^2$, writing
$\bar f(m,m')=\E[f\mid m,m']$ and using
$\E_{\mu\otimes\mu}[\bar f\,\Jd]=\E_\pi[\bar f]$,
\begin{equation}\label{eq:proj}
\inner{f}{\Jd-1}=\E_{\mu\otimes\mu}\big[\bar f\,(\Jd-1)\big]=\E_\pi[f]-\E[f].
\end{equation}

Apply Eq. \ref{eq:proj} to $f=k^c_X$. By centering Eq. \ref{eq:centred}, 
$\E[k^c_X]= 0$. For the $\pi$-expectation,
$$\E_\pi[k^c_X]= \E_\pi [k_X]-\E_\pi\E[k_X\mid x]-\E_\pi\E[k_X\mid x^+]+\E_\pi\E[k_X].$$
Because the marginals of $\pi$ are $\mu$, $\E_\pi\E[k_X\mid x]=\E_\pi\E[k_X\mid x^+] = \E[k_X]$, hence
$$\E_\pi[k^c_X] =\AX-\SX-\SX+\SX=\Delta_X,$$ and Eq. \ref{eq:proj} gives the claim.

The norm is Eq. \ref{eq:chi}: the marginal of $L^2$ on $(m,m')$ is $\mu\otimes\mu$, so
embedding a function of $(m,m')$ into $L^2$ preserves its norm.
\end{proof}

\begin{lemma}[Kernel scale]\label{lem:var}
$\displaystyle \sigma_X^2\ \le\ D\,L_X$, with $D$ as in Eq. \ref{eq:constants}.
\end{lemma}

\begin{proof}
Let $Cg=g-\E[g\mid x]-\E[g\mid x']+\E[g]$, so that $k^c_X=Ck_X$. Since $x\perp x'$ we
may factor $C=(\mathrm{Id}-P_1)(\mathrm{Id}-P_2)$ with $P_1g=\E[g\mid x']$ and
$P_2g=\E[g\mid x]$; each $P_i$ is an orthogonal projection, so each $\mathrm{Id}-P_i$
is a contraction and $\lVert Cg\rVert\le\lVert g\rVert$. As $Cc = 0$ for every constant $c$,
$\sigma_X=\lVert k_X^c \rVert = \lVert C(k_X-c)\rVert\le\lVert k_X-c\rVert$ for every constant $c$.

Take $c=e^{-\omega}$. By Eq. \ref{eq:kernel},
$k_X-e^{-\omega}=e^{-\omega}\big(e^{\omega\innert}-1\big)$, and
Lemma~\ref{lem:chord}(ii) with $\alpha=\omega$ gives
$\lvert k_X-e^{-\omega}\rvert = e^{-\omega} \lvert \big(e^{\omega\innert}-1\big) \rvert \le(1-e^{-\omega})\lvert\innert\rvert$. Squaring and
applying the second bound of Lemma~\ref{lem:small} gives $\sigma_X^2\le D\,L_X$.
\end{proof}

\setcounter{theorem}{0}
\begin{theorem}[Implicit alignment]\label{thm:align}
Suppose $\sigma_X\sigma_Y>0$ and $\chi>0$. Define, with $a,b,D$ as in
Eq. \ref{eq:constants},
\begin{equation}\label{eq:c}
c(L)\ :=\ \frac{\big[\,a-bL\,\big]_+}{\chi\,\sqrt{D\,L}},
\qquad c_X=c(L_X),\quad c_Y=c(L_Y).
\end{equation}
Then $c_X,c_Y\in[0,1]$ and
\[
\boxed{\ \ \CKA(\EX,\EY)\ \ge\ c_Xc_Y-\sqrt{(1-c_X^2)(1-c_Y^2)}\ =:\ f(L_X,L_Y)\ \ }
\]
\end{theorem}

\begin{proof}
Lemma~\ref{lem:var} gives
$L_X\ge\sigma_X^2/D>0$ because $\sigma_X>0$, so $c(L_X)$ is well defined; likewise $c(L_Y)$.

If $c_X=0$ or $c_Y=0$ the claimed right-hand side is $\le0$, and the bound follows
from $\CKA\ge0$ (see Eq. \ref{eq:cka}). Assume therefore $a-bL_X>0$ and $a-bL_Y>0$, and
work in $L^2$ with the three unit vectors
\[
\hat a=\frac{k^c_X}{\sigma_X},\qquad \hat b=\frac{k^c_Y}{\sigma_Y},\qquad
\hat\jmath=\frac{\Jd-1}{\chi},
\]
and let $\theta_X=\angle(\hat a,\hat\jmath)$, $\theta_Y=\angle(\hat b,\hat\jmath)$.
Lemma~\ref{lem:anchor} evaluates the cosine, Lemma~\ref{lem:coef} bounds its numerator
below by a positive quantity, and Lemma~\ref{lem:var} bounds its denominator above:
\begin{equation}\label{eq:cos}
\cos\theta_X=\frac{\inner{k^c_X}{\Jd-1}}{\chi\,\sigma_X}=\frac{\Delta_X}{\chi\,\sigma_X}
\ \ge\ \frac{a-bL_X}{\chi\sqrt{D\,L_X}}=c_X\ >\ 0 ,
\end{equation}
and likewise $\cos\theta_Y\ge c_Y>0$. In particular $c_X\le\cos\theta_X\le1$, so $c_X\in(0,1]$ and $\theta_X\le\arccos c_X\le\pi/2$; the same for
$Y$.

The angle $\angle(\cdot,\cdot)=\arccos\inner{\cdot}{\cdot}$ is the geodesic distance on
the unit sphere of $L^2$, hence satisfies the triangle inequality:
\[
\angle(\hat a,\hat b)\ \le\ \angle(\hat a,\hat\jmath)+\angle(\hat b,\hat\jmath)
=\theta_X+\theta_Y\ \le\ \arccos c_X+\arccos c_Y\ \le\ \pi .
\]
Apply $\cos$, decreasing on $[0,\pi]$, and expand
$\cos(\arccos c_X+\arccos c_Y)=c_Xc_Y-\sqrt{(1-c_X^2)(1-c_Y^2)}$. Noting
$\cos\angle(\hat a,\hat b)=\CKA(\EX,\EY)$ gives
$$\CKA(\EX,\EY)\ \ge\ c_Xc_Y-\sqrt{(1-c_X^2)(1-c_Y^2)}$$
\end{proof}

\subsection{Implicit misalignment}\label{sec:mis}

This section gives the complement direction and proves the implicit misalignment theorem (Thm. \ref{thm:mis}). That is, a large unimodal loss restricts the CKA to be small. The key observation is that only semantic variation can align across modalities. Lemma \ref{lem:pyth} formalizes this: given the shared semantics, the two embeddings are independent, so only the semantic part of each kernel contributes to the CKA. Hence, the CKA is upper-bounded by the semantic variance $\beta_X$, i.e., how much the embedding varies with the semantics, relative to its overall variation $\sigma_X$. It then remains to show that a large loss forces $\beta_X$ to be small. Lemma \ref{lem:lip} bounds the semantic variance by the augmentation gap $\Delta_X$, up to how much the augmentations displace the observations, which is where the Lipschitz assumption (Assumption \ref{ass:lip}) is used. Lemma \ref{lem:large} shows that a large loss forces this gap to be small. Together, a large loss leaves much of the variation view-specific, and Thm. \ref{thm:mis} chains these inequalities into an upper bound on the CKA.

\subsubsection{CKA and the semantic variance}

\paragraph{Notations.} Let $\phi(x)=k(x,\cdot)$ be the canonical feature map of the Gaussian kernel (Eq. \ref{eq:kernel}) into its
RKHS $\HH$ \citep{Muandet_2017}, so that
$\inner{\phi(x)}{\phi(x')}_\HH=k(x,x')$ and $\lVert\phi(x)\rVert_\HH=1$ on the sphere.
Put
\begin{equation}\label{eq:mean-emb}
\mu_m:=\E[\phi(x)\mid m],\qquad
\bar\mu:=\E[\phi(x)],\qquad
\delta_m:=\mu_m-\bar\mu ,
\end{equation}
and define the \emph{semantic kernel} and the \emph{semantic variance} induced by $X$
\begin{equation}\label{eq:semantic}
h_X(m,m'):=\E\big[k^c_X\mid m,m'\big],
\qquad
\beta_X:=\E_{m\sim\mu}\big[h_X(m,m)\big] .
\end{equation}

\begin{lemma}[]\label{lem:pyth}
$h_X(m,m')=\inner{\delta_m}{\delta_{m'}}_\HH$ and
$\beta_X=\E_m\lVert\delta_m\rVert^2_\HH\ge0$. Moreover, writing
$s_X:=\lVert h_X\rVert$,
\[
\CKA(\EX,\EY)\ \le\ \frac{s_X}{\sigma_X}\ \le\ \frac{\beta_X}{\sigma_X}.
\]
\end{lemma}

\begin{proof}
Expanding Eq. \ref{eq:centred} with $\E[k_X\mid x]=\inner{\phi(x)}{\bar\mu}_\HH$ gives
the pointwise identity
$k^c_X=\inner{\phi(x)-\bar\mu}{\phi(x')-\bar\mu}_\HH$. Given $(m,m')$ the nuisances
$u,u'$ are independent, so taking conditional expectations inside the inner product
yields $h_X(m,m')=\inner{\delta_m}{\delta_{m'}}_\HH$; in particular
$h_X(m,m)=\lVert\delta_m\rVert^2_\HH$ and $\beta_X=\E_m\lVert\delta_m\rVert^2_\HH$.

Because $(u,u')$ and $(v,v')$ are independent of each other and of $(m,m')$, the pairs
$(x,x')$ and $(y,y')$ are conditionally independent given $(m,m')$, so
$\E[k^c_Xk^c_Y\mid m,m']=h_Xh_Y$ and $\hsic(\EX,\EY)=\inner{h_X}{h_Y}$. Now $h_X$ is
the orthogonal projection of $k^c_X$ onto the functions of $(m,m')$, hence
$s_X\le\sigma_X$ and likewise $s_Y\le\sigma_Y$. Cauchy--Schwarz in $L^2$ gives
$\hsic(\EX,\EY)\le s_Xs_Y\le s_X\sigma_Y$; divide by $\sigma_X\sigma_Y$.

Finally, Cauchy--Schwarz in $\HH$ and independence of $m,m'$ give
\[
s_X^2=\E_{m,m'}\inner{\delta_m}{\delta_{m'}}^2
\le\E_m\lVert\delta_m\rVert^2\,\E_{m'}\lVert\delta_{m'}\rVert^2=\beta_X^2 . \qedhere
\]
\end{proof}

\subsubsection{Semantic displacement and the augmentation gap}

\begin{assumption}[Lipschitz encoders]\label{ass:lip}
$\lVert e_X(p)-e_X(q)\rVert\le\ell_X\,d_\X(p,q)$ for all $p,q\in\X$, and likewise for
$e_Y$ with constant $\ell_Y$.
\end{assumption}

\begin{definition}[Semantic displacement]\label{def:eps}
\begin{equation}\label{eq:eps}
\varepsilon_X:=\E_{\pi,\,u}\Big[d_\X\big(g_X(m,u),g_X(m^+,u)\big)^2\Big],
\end{equation}
the mean-square distance the augmentation moves an observation when the nuisance
is held fixed, with $d_\X$ a metric on $\X$.
\end{definition}

\begin{lemma}[Semantic variance versus augmentation gap]\label{lem:lip}
Under Assumption~\ref{ass:lip},
$$\beta_X \leq \Delta_X + \tfrac12\,\omega\,\ell_X^2\,\varepsilon_X .$$
\end{lemma}

\begin{proof}
By Lemma~\ref{lem:anchor} and Eq. \ref{eq:proj},
$\Delta_X=\E_\pi[k^c_X]=\E_\pi[h_X]=\E_\pi\inner{\delta_m}{\delta_{m^+}}$. Both
marginals of $\pi$ are $\mu$, so
$\E_\pi\lVert\delta_m\rVert^2=\E_\pi\lVert\delta_{m^+}\rVert^2=\beta_X$ and
\[
\E_\pi\big\lVert\mu_m-\mu_{m^+}\big\rVert^2
=\E_\pi\big\lVert\delta_m-\delta_{m^+}\big\rVert^2=2\beta_X-2\Delta_X.
\]

For the bound, couple the two mean embeddings through the same nuisance: with
$\psi=\phi\circ e_X$, $\mu_m-\mu_{m^+}=\E_u[\psi(g_X(m,u))-\psi(g_X(m^+,u))]$, so by
Jensen
$$\lVert\mu_m-\mu_{m^+}\rVert\le\E_u\lVert\psi(g_X(m,u))-\psi(g_X(m^+,u))\rVert .$$
On the sphere, $1-e^{-t}\le t$ gives
\[
\lVert\phi(p)-\phi(q)\rVert_\HH^2=2\big(1-e^{-\omega\lVert p-q\rVert^2/2}\big)
\le\omega\lVert p-q\rVert^2 ,
\]
so $\phi$ is $\sqrt\omega$-Lipschitz and $\psi$ is $\sqrt\omega\,\ell_X$-Lipschitz.
Hence
$$\lVert\mu_m-\mu_{m^+}\rVert\le\sqrt\omega\,\ell_X\,\E_u\,d_\X(g_X(m,u),g_X(m^+,u)),$$
and a second application of Jensen gives
$\E_\pi\lVert\mu_m-\mu_{m^+}\rVert^2\le\omega\,\ell_X^2\,\E_{\pi,u}[d_\X^2]
=\omega\,\ell_X^2\varepsilon_X$.
\end{proof}

Write
\begin{equation}\label{eq:Lambda}
\Lambda(\tau):=\coth\tfrac1\tau-\tau+\tau\log\big(\tau\tanh\tfrac1\tau\big) - 1\ \in\ (-1,0).
\end{equation}

\begin{lemma}[The loss controls the gap from above]\label{lem:large}
Suppose $\tau\,\omega\le1$. Then,
$$\Delta_X\ \le\ a\,\big(\Lambda(\tau)+\tau R ( 1-L_X ) \big) .$$
\end{lemma}

\begin{proof}
The objective controls two quantities: the alignment $A_X$ from above and the
uniformity $T:=\E\big[e^{\innert/\tau}\big]>0$ from below. Chord bounds convert $\AX$
and $\SX$ into exactly those two, after which the constraint linking them makes $T$
cancel.

First, Lemma~\ref{lem:chord}(i) with $\alpha=\omega$ gives
$e^{\omega}\AX\le\cosh\omega+A_X\sinh\omega$. Second, Lemma~\ref{lem:chord}(iii) with
$b=\omega\le\alpha=1/\tau$, gives,
$$e^{\omega \innert}\ge e^{-\omega}+\frac{\sinh \omega}{\sinh\frac1\tau}\big(e^{\frac1\tau \innert}-e^{-\frac1\tau}\big).$$
After averaging,
$$e^{\omega}\SX=\E[e^{\omega\innert}]\ge
e^{-\omega}+\frac{\sinh\omega}{\sinh\frac1\tau}\big(T-e^{-1/\tau}\big).$$ Subtracting $\AX -\SX$,
and using $\cosh\omega-e^{-\omega}=\sinh\omega$,
\begin{equation}\label{eq:pre}
\frac{e^{\omega}\Delta_X}{\sinh\omega}\ \le\ 1+A_X-\frac{T-e^{-1/\tau}}{\sinh\frac1\tau}.
\end{equation}
Now eliminate $A_X$. By Eq. \ref{eq:loss} and Jensen ($U_X\le\log T$),
$A_X=\tau(U_X-N_X)\le\tau(\log T-N_X)$. Bound the logarithm by its tangent at
$c=\tau\sinh\frac1\tau$, namely $\log T\le T/c-1+\log c$, valid for every $T>0$:
\[
A_X\ \le\ \frac{T}{\sinh\frac1\tau}-\tau+\tau\log\big(\tau\sinh\tfrac1\tau\big)-\tau N_X .
\]
Using $1+e^{-1/\tau}/\sinh\frac1\tau=\coth\frac1\tau$,
\[
\frac{e^{\omega}\Delta_X}{\sinh\omega}\ \le\
\coth\tfrac1\tau-\tau+\tau\log\big(\tau\sinh\tfrac1\tau\big)-\tau N_X .
\]
Finally $N_X=R\,L_X-\tfrac1\tau$ and $\tau\log\cosh\tfrac1\tau=\tau R-2$ by
Eq. \ref{eq:normloss} and Eq. \ref{eq:kappa}, so
$\tau\log(\tau\sinh\tfrac1\tau)-\tau N_X=\tau\log(\tau\tanh\tfrac1\tau)-1+\tau R (1-L_X)$,
giving the claim.
\end{proof}

\subsubsection{Connecting a high loss with low CKA}

\begin{theorem}[Implicit misalignment]\label{thm:mis}
Suppose $\sigma_X>0$, $\tau\,\omega\le1$, and Assumption~\ref{ass:lip} holds. Then

\[
\boxed{\ \ \CKA(\EX,\EY)\ \le\
\frac{d + \ell_X^2\varepsilon_X}
{2\sigma^2\sigma_X} + \frac{\tau R a}{\sigma_X}(1-L_X) =:\ g(L_X)\ \ }
\]
with $d = 2\sigma^2 a\left(\coth\tfrac1\tau-\tau+\tau\log\big(\tau\tanh\tfrac1\tau\big) - 1 \right)$, and the same statement holds with $Y$ in place of $X$.
\end{theorem}

\begin{proof}
Chain Lemma~\ref{lem:pyth}, Lemma~\ref{lem:lip} and Lemma~\ref{lem:large}:
$$\CKA(\EX,\EY)\ \le\ \frac{\beta_X}{\sigma_X}
\ \le\ \frac{\Delta_X+\tfrac12\,\omega\,\ell_X^2\,\varepsilon_X}{\sigma_X}
\ \le\ \frac{a\big(\Lambda(\tau)+\tau R\,(1-L_X)\big)+\tfrac12\,\omega\,\ell_X^2\,\varepsilon_X}
{\sigma_X}.$$

Reorganization gives the bound.
\end{proof}

\subsection{The Loss Range}

\begin{lemma}[Range of the normalized loss]\label{lem:range}
$L_X \in [0,1]$. Moreover:
\begin{enumerate}[label=(\roman*),leftmargin=*]
  \item $L_X = 0$ if and only if $\langle x, x^+\rangle = 1$ and $\langle x, x'\rangle = 0$
        almost surely;
    \item $L_X = 1$ if and only if there is a unit vector $e$ such that
        $x \in \{e,-e\}$ almost surely with $\Pr(x = e) = \tfrac12$, and $x^+=-x$
        almost surely;
  \item If additionally         Assumption~\ref{ass:lip} is satisfied and $\ell_X^2\varepsilon_X<4$, then
        $$L_X\ \le\ 1-\frac{1-\tfrac14\ell_X^2\varepsilon_X}{\tau R}\ <\ 1,$$
        with equality if and only if there is a unit vector $e$ such that $x\in\{e,-e\}$ almost
        surely with $\Pr(x=e)=\tfrac12$, and the augmentation reverses $q(m)=\E[\inner{x}{e}\mid m]$ ($q(m^+)=-q(m)$) $\pi$-almost surely.
\end{enumerate}
\end{lemma}
 
\begin{proof}
\emph{Lower bound.} $\E[\innerp]\le1$ and the uniformity term
is nonnegative (Eq. \ref{eq:unifloor}), so $N_X\ge-1/\tau$, i.e.\ $L_X\ge0$.

\emph{Case (i).} Equality
forces $\innerp=1$ a.s.\ and, by Lemma~\ref{lem:small}, $\lVert\bar x\rVert^2=0$ and
$\E[V]=0$; then $r(x)^2\le\lVert\bar x\rVert^2=0$ and $V(x)=0$ a.s., so $\innert=0$
a.s. The converse is a direct substitution.

\emph{Upper bound.} Since both marginals of $\pi$ are $\mu$, the pair $(x,x^+)$ is a
coupling of $\rho_X$ with itself. Writing $z= x - \bar x$ and
$z^+=x^+ - \bar x$, the cross terms vanish and $A_X=\lVert \bar x \rVert^2+\E\inner{z}{z^+}$; as $\lVert x\rVert=1$
we have $\E\lVert z\rVert^2=\E\lVert z^+\rVert^2=1-\lVert \bar x \rVert^2$, so Cauchy--Schwarz gives
\begin{equation}\label{eq:range-align}
A_X\ \ge\ \lVert \bar x \rVert^2 - \sqrt{\E\lVert z \rVert^2} \sqrt{\E \lVert z^+ \rVert^2} = \lVert \bar x \rVert^2 - (1 - \lVert \bar x \rVert^2)\ =\ 2\lVert \bar x \rVert^2-1 .
\end{equation}
By Jensen and Lemma~\ref{lem:chord}(i) with $\alpha=1/\tau$,
$c=\innert$,
\begin{equation} \label{eq:range-upper}
    U_X = \E_x\Big[\log\E_{x'}\big[e^{\innert/\tau}\big]\Big]\le\log\E\big[e^{\innert/\tau}\big]
    \le\log\Big(\cosh\tfrac1\tau+\lVert\bar x\rVert^2\sinh\tfrac1\tau\Big).
\end{equation}
 
Denote $t = \lVert \bar x \rVert^2$. With Eq. \ref{eq:range-align} this gives $N_X\le f(t)$ for
$f(t)=(1-2t)/\tau+\log(\cosh\tfrac1\tau+t\sinh\tfrac1\tau)$, and
$f'(t)\le-\tfrac2\tau+\tanh\tfrac1\tau<0$ for $t\ge0$, so
$N_X\le f(0)=\tfrac1\tau+\log\cosh\tfrac1\tau=R-\tfrac1\tau$. This gives $L_X \leq 1$.
 
\emph{Case (ii).}
$L_X = 1$, meaning $N_X = R - \tfrac1\tau$, forces  equality throughout. That is, the chord bound in Eq. \ref{eq:range-upper} is tight;
Jensen in Eq. \ref{eq:range-upper} is tight; Cauchy--Schwarz in Eq. \ref{eq:range-align} is tight;
and $\|\bar x\|^2 = 0$.
 
Since $c \mapsto e^{c/\tau}$ is strictly convex, it meets its chord on $[-1,1]$ only at
the endpoints, so tightness of the chord bound forces $\langle x, x'\rangle \in \{-1,1\}$ a.e., i.e.\ $x' \in \{x,-x\}$. Hence, there is a single
$e$ in the support with $\rho_X(\{e,-e\}) = 1$. Writing $p = \Pr(x = e)$ we get
$\bar x = (2p-1)e$, so $\|\bar x\|^2 = 0$ gives $p = \tfrac12$. Then
$\mathbb{E}_{x'}[e^{\langle x,x'\rangle/\tau}] = \cosh\tfrac{1}{\tau}$ for both values of $x$. With $\bar x=0$ we have $z=x$, and equality in
Eq. \ref{eq:range-align} forces $x^+=-x$ a.s. Conversely these conditions give $A_X=-1$ and
$U_X=\log\cosh\tfrac1\tau$, hence $N_X = \tfrac1\tau+\log\cosh\tfrac{1}{\tau}
= R - \tfrac{1}{\tau}$ by direct substitution.

\emph{Case (iii).} Let $c_m:=\E[x\mid m]$ be the semantic centroid, $d_m:=c_m-\bar x$,
$\gamma:=\E_m\lVert d_m\rVert^2$ and $\Gamma:=\E_\pi\lVert d_m-d_{m^+}\rVert^2$.
Since $u$ and $u^+$ are drawn independently, $A_X=\E_\pi\inner{c_m}{c_{m^+}}$, and since both
marginals of $\pi$ are $\mu$, decomposing $\E_\pi\inner{d_m}{d_{m^+}}$ gives,
\begin{equation}\label{eq:range-split}
A_X\ =\ \lVert\bar x\rVert^2+\gamma-\tfrac12\Gamma .
\end{equation}
Now we bound $\Gamma$ in two ways. Coupling the two centroids through the same nuisance,
$c_m-c_{m^+}=\E_u[e_X(g_X(m,u))-e_X(g_X(m^+,u))]$, so Jensen and Assumption~\ref{ass:lip}
give
$$\Gamma = \E_\pi \lVert c_m - c_{m^+} \rVert^2 \leq \E_\pi\E_u[ \lVert e_X(g_X(m,u))-e_X(g_X(m^+,u)) \rVert]^2 \leq \ell_X^2\varepsilon_X.$$
Also, $\lVert d_m-d_{m^+}\rVert^2
\le2\lVert d_m\rVert^2+2\lVert d_{m^+}\rVert^2$ gives $\Gamma\le4\gamma$. Feeding the
smaller of the two into Eq. \ref{eq:range-split},
$$A_X-\lVert\bar x\rVert^2\ \ge\ \gamma-\tfrac12\min\{\ell_X^2\varepsilon_X,4\gamma\}
\ =\ \max\{\gamma-\tfrac12\ell_X^2\varepsilon_X,\,-\gamma\}\ \ge\ -\tfrac14\ell_X^2\varepsilon_X,$$
the last step because the two branches meet at $\gamma=\tfrac14\ell_X^2\varepsilon_X$.

Replacing Eq. \ref{eq:range-align} by this and repeating the argument of the upper bound,
$N_X\le\psi(\lVert\bar x\rVert^2)$ for
$\psi(t)=(\tfrac14\ell_X^2\varepsilon_X-t)/\tau+\log(\cosh\tfrac1\tau+t\sinh\tfrac1\tau)$,
and $\psi'(t)\le-\tfrac1\tau+\tanh\tfrac1\tau<0$, so
$N_X\le\psi(0)=R-\tfrac1\tau-\tfrac1\tau\big(1-\tfrac14\ell_X^2\varepsilon_X\big)$,
which is the claim; it is $<1$ because $\ell_X^2\varepsilon_X<4$.

Equality forces $\lVert\bar x\rVert^2=0$, both steps of Eq. \ref{eq:range-upper} tight, and
$\gamma=\tfrac14\ell_X^2\varepsilon_X$ with $\Gamma=4\gamma$. As in case (ii) the first two
give $\rho_X(\{e,-e\})=1$ with $\Pr(x=e)=\tfrac12$, and $d_m=q(m)e$ for
$q(m)=\E[\inner{x}{e}\mid m]$; then $\Gamma=4\gamma$ holds exactly when $q(m^+)=-q(m)$
$\pi$-a.s., and $\gamma=\E_m q(m)^2$.
\end{proof}

\subsection{The shape of the bounds}

\subsubsection{The shape of the implicit alignment bound}

This section analyses the shape of the implicit alignment bound from Thm. \ref{thm:align}. Specifically, it proves that the bound $f(L_X, L_Y)$ is nonincreasing in both $L_X$ and $L_Y$, and converges to the desired limits (Prop. \ref{prop:bound_shape}).

\begin{lemma}[Shape of the alignment coefficient]\label{lem:cshape}
For $L\in(0,1]$. Then:
\begin{enumerate}[label=(\roman*),leftmargin=*]
\item $c$ is nonincreasing on $(0,1]$, and strictly decreasing on the set where
$0<c(L)<1$.
\item $c(L)=1$ for every $L\le L_1$, where
\[
L_1:=\frac{2ab+\chi^2D-\sqrt{\chi^2D\,(4ab+\chi^2D)}}{2b^2}\ \in(0,a/b),
\]
and $c(L)<1$ for $L>L_1$. In particular $\lim_{L\to0^+}c(L)=1$.
\end{enumerate}
\end{lemma}

\begin{proof}
(i) Since $a, b, D, \chi > 0$, on $(0,\infty)$ the map $L\mapsto[a-bL]_+$ is nonnegative and nonincreasing and
$L\mapsto\chi\sqrt{DL}$ is positive and increasing, so their ratio $g(L)$ is
nonincreasing. Composing with the map $t\mapsto\min\{1,t\}$ preserves
this. On $\{L<a/b\}$ we may drop the positive part and write
$g(L)=(aL^{-1/2}-bL^{1/2})/(\chi\sqrt D)$, whence
$g'(L)=-\tfrac12(aL^{-3/2}+bL^{-1/2})/(\chi\sqrt D)<0$; since $c=g$ wherever
$0<c<1$, the strict decrease follows.

(ii) For $L>0$, $c(L)=1$ iff $a-bL\ge\chi\sqrt{DL}$, which forces $L<a/b$ and is then
equivalent, after squaring, to $q(L)\ge0$ with
$q(L):=b^2L^2-(2ab+\chi^2D)L+a^2$. The parabola $q$ opens upwards, and
\[
q(0)=a^2>0,\qquad q(a/b)=a^2-\big(2ab+\chi^2D\big)\tfrac ab+a^2=-\chi^2D\,\tfrac ab<0 ,
\]
so $q$ has two positive roots $L_1<a/b<L_2$ and $\{q\ge0\}\cap(0,a/b)=(0,L_1]$. The
quadratic formula gives the stated $L_1$, using
$(2ab+\chi^2D)^2-4a^2b^2=\chi^2D(4ab+\chi^2D)$. Hence, $c\equiv1$ on $(0, L_1]$, and since $L_1>0$, $c\equiv1$ is attainable, and gives the limit.

\end{proof}

\begin{proposition}[Monotonicity of the bound]\label{prop:bound_shape}
Denote the right-hand side of Thm. \ref{thm:align} $f(L_X,L_Y)$. With $L_1$ as in Lemma~\ref{lem:cshape}:
\begin{enumerate}[label=(\roman*),leftmargin=*]
\item $f$ is nonincreasing in $L_X$ and in $L_Y$ on $(0,1]^2$, and is strictly
decreasing in $L_X$ at every point with $c(L_X)\in(0,1)$ and $c(L_Y)>0$.
\item $f\equiv1$ on $(0,L_1]^2$.
\end{enumerate}
\end{proposition}

\begin{proof}
Let $g(c(L_X), c(L_Y)) = f(L_X, L_Y)$. Write $\theta_X=\arccos c_X,\ \theta_Y=\arccos c_Y\in[0,\pi/2]$, so that
$$ g(c_X, c_Y) = c_Xc_Y-\sqrt{(1-c_X^2)(1-c_Y^2)}
=\cos\theta_X\cos\theta_Y-\sin\theta_X\sin\theta_Y=\cos(\theta_X+\theta_Y).$$
Now $c_X\mapsto\theta_X$ is decreasing on $[0,1]$,
$\theta_X+\theta_Y\in[0,\pi]$, and $\cos$ is decreasing on $[0,\pi]$; a decreasing
function of a decreasing function is nondecreasing, so $g$ is nondecreasing in $c_X$,
and by symmetry in $c_Y$. For the strict monotonicity of $g$ on $c_X\in(0,1)$ the chain rule
gives
\[
\frac{\partial g}{\partial c_X}=\frac{\sin(\theta_X+\theta_Y)}{\sin\theta_X}\ \ge\ 0 ,
\]
and $\sin(\theta_X+\theta_Y)=0$ with $\theta_X+\theta_Y\in[0,\pi]$ forces
$\theta_X+\theta_Y\in\{0,\pi\}$, i.e.\ $c_X=c_Y=1$ or $c_X=c_Y=0$; at $c_X=0$ the
one-sided derivative is $\sin(\pi/2+\theta_Y)=\cos\theta_Y=c_Y>0$ unless $c_Y=0$.

(i) From Lemma~\ref{lem:cshape}(i): $f$ is a nondecreasing function
of $c(L_X)$, which is nonincreasing in $L_X$. Strictness likewise combines the strict
decrease of $c$ on $\{0<c<1\}$ with the strict increase of $g$.

(ii) By Lemma~\ref{lem:cshape}(ii), $c\equiv1$ on $(0,L_1]$ and $g(1,1)=1$.
\end{proof}

\subsubsection{The shape of the implicit misalignment bound}

This section analyses the shape of the implicit misalignment bound from Thm. \ref{thm:mis}. Specifically, it proves that the bound $g(L)$ is strictly decreasing $L$ and converges to the desired limits (Prop. \ref{prop:mis_bound_shape}).

\begin{proposition}[Monotonicity of the misalignment bound]\label{prop:mis_bound_shape}
    The bound $g$ is strictly decreasing in $L_X$, and is minimized at the largest possible loss as in Lemma \ref{lem:range}(iii), where it equals $$g\left( 1-\frac{1-\tfrac14 \ell_X^2\varepsilon_X }{\tau R} \right) = \frac{a\left(1 + \Lambda(\tau) - \tfrac14 \ell_X^2\varepsilon_X \right)+\tfrac12\omega\ell_X^2\varepsilon_X}{\sigma_X},$$
    if $\ell_X^2\varepsilon_X < 4$, and otherwise at $g(1) = \left( a\Lambda(\tau) +\tfrac12\omega\ell_X^2\varepsilon_X \right) / {\sigma_X}$.
\end{proposition}

\begin{proof}
    Monotonicity is immediate: $g$ is affine in $L_X$ with negative slope
$-a\tau R/\sigma_X$.
\end{proof}

\subsection{General inequalities}

This section provides general mathematical inequalities used in the proofs above.

\begin{lemma}[Chord bounds]\label{lem:chord}
Let $\alpha>0$ and $c\in[-1,1]$. Then
\begin{enumerate}[label=(\roman*),leftmargin=*]
\item $e^{\alpha c}\le\cosh\alpha+c\sinh\alpha$;
\item $\lvert e^{\alpha c}-1\rvert\le(e^{\alpha}-1)\lvert c\rvert$;
\item for every $b$ with $0<b\le\alpha$, \
$\displaystyle e^{bc}\ge e^{-b}+\frac{\sinh b}{\sinh\alpha}\big(e^{\alpha c}-e^{-\alpha}\big)$.
\end{enumerate}
\end{lemma}

\begin{proof}
(i) By convexity, $c\mapsto e^{\alpha c}$ lies below its chord on $[-1,1]$, which is
$\cosh\alpha+c\sinh\alpha$.

(ii) The difference quotient $c\mapsto(e^{\alpha c}-1)/c=\alpha\int_0^1e^{\alpha cs}\,ds$
(value $\alpha$ at $c=0$) is positive and nondecreasing, hence at most its value
$e^{\alpha}-1$ at $c=1$; multiply by $\lvert c\rvert$.

(iii) Put $t=e^{\alpha c}\in[e^{-\alpha},e^{\alpha}]$. Then $e^{bc}=t^{b/\alpha}$ is
concave in $t$ because $b/\alpha\le1$, so it lies above the chord joining
$(e^{-\alpha},e^{-b})$ and $(e^{\alpha},e^{b})$, whose slope is
$\tfrac{e^{b}-e^{-b}}{e^{\alpha}-e^{-\alpha}}=\tfrac{\sinh b}{\sinh\alpha}$. The right side
being affine in $t$, and the left concave, the inequality survives taking expectations.
\end{proof}

\begin{lemma}[Variance floor for the log-MGF]\label{lem:mgf}
Let $\delta$ have mean zero and support in an interval of length at most $2$, and let
$\lambda\ge1$. Then $\log\E[e^{\lambda\delta}]\ge\tfrac{\log2}{4}\lambda\Var(\delta)$.
\end{lemma}

\begin{proof}
Write $V=\Var(\delta)=\E[\delta^2]$ and let $[\ell,u]$ contain the support with
$0\in[\ell,u]$ and $u-\ell\le2$, so $\delta\ge-2$.

Taylor's formula with integral remainder gives $e^{t}=1+t+t^2\psi(t)$ with
$\psi(t)=\int_0^1(1-s)e^{st}\,ds=(e^t-1-t)/t^2$ nondecreasing. Since
$\lambda\delta\ge-2\lambda$,
\begin{equation}\label{eq:mgffloor}
\E\big[e^{\lambda\delta}\big]\ \ge\ 1+\lambda^2\psi(-2\lambda)V\ \ge\ 1+\tfrac{\lambda}{4}V,
\end{equation}
because $\lambda^2\psi(-2\lambda)=\tfrac{e^{-2\lambda}-1+2\lambda}{4}\ge\tfrac{2\lambda-1}{4}
\ge\tfrac{\lambda}{4}$ for $\lambda\ge1$.

\emph{Case 1: $\lambda V/4\le1$.} By concavity $\log(1+s)\ge s\log2$ on $[0,1]$, so
Eq. \ref{eq:mgffloor} gives the claim.

\emph{Case 2: $\lambda V/4>1$.} Let $p=\Pr(\delta>0)$, positive because $p=0$ with
$\E\delta=0$ forces $V=0$, and let
$h=\E[\delta\mathbf 1_{\delta>0}]=\E[-\delta\mathbf 1_{\delta<0}]$. Bounding
$\delta\le u$ on $\{\delta>0\}$ and $-\delta\le-\ell$ on $\{\delta<0\}$ gives
$V\le(u-\ell)h\le2h$, so $\lambda h\ge\lambda V/2>2\ge p$. Discarding the negative part
and applying Jensen to the law of $\delta$ given $\{\delta>0\}$, gives
$$\E[e^{\lambda\delta}]\ \ge\ p\,\E[e^{\lambda\delta}\mid\delta>0]\ \ge\ p\,e^{\lambda h/p}.$$
The map $p\mapsto p\,e^{\lambda h/p}$ is nonincreasing on $(0,\lambda h]$, so at
$p\le1$ it is at least $e^{\lambda h}$, whence
$$\log\E[e^{\lambda\delta}]\ \ge\ \lambda h\ \ge\ \tfrac{\lambda V}{2}\ \ge\ \tfrac{\log2}{4}\lambda V.$$
\end{proof}

\subsection{Constants Details} \label{app:proofs:constants}

Thm. \ref{thm:align_main}, as presented in the main text, incorporates several hyperparameter-dependent constants: $$a_{\sigma} := e^{-\oomega}\sinh\oomega,$$ which depends on the CKA-RBF scale $\sigma$. Recall $R_\tau = \tfrac{2}{\tau} + \log\cosh\tfrac{1}{\tau}$
$$b_{\sigma,\tau} := \big(\oomega+e^{-\oomega}\sinh\oomega\big)\tau R_\tau,\quad D_{\sigma,\tau} :=\Big(1+\frac{4}{\log2}\Big)\big(1-e^{-\oomega}\big)^2\,\tau R_\tau,$$ which also rely on the InfoNCE temperature $\tau$.

Similarly, Thm. \ref{thm:misalign_main} incorporates other hyperparameter-dependent constants:
$$h_{\sigma, \tau} = \tau R_\tau a_{\sigma},$$
which depends on the InfoNCE temperature $\tau$, and
$$d_{\sigma, \tau} = 2\sigma^2 a_{\sigma}\left(\coth\tfrac1\tau-\tau+\tau\log\big(\tau\tanh\tfrac1\tau\big) -1 \right),$$
which also incorporates the CKA-RBF scale $\sigma$.

\section{AudioCaps Results}

In Sec. \ref{sec:results:pretrained}, the evaluation of unconstrained, pre-trained audio models focused primarily on the CLOTHO dataset. To further support our claim across audio datasets, we extend our analysis to AudioCaps.

\begin{figure}
    \centering
    \includegraphics[width=\linewidth]{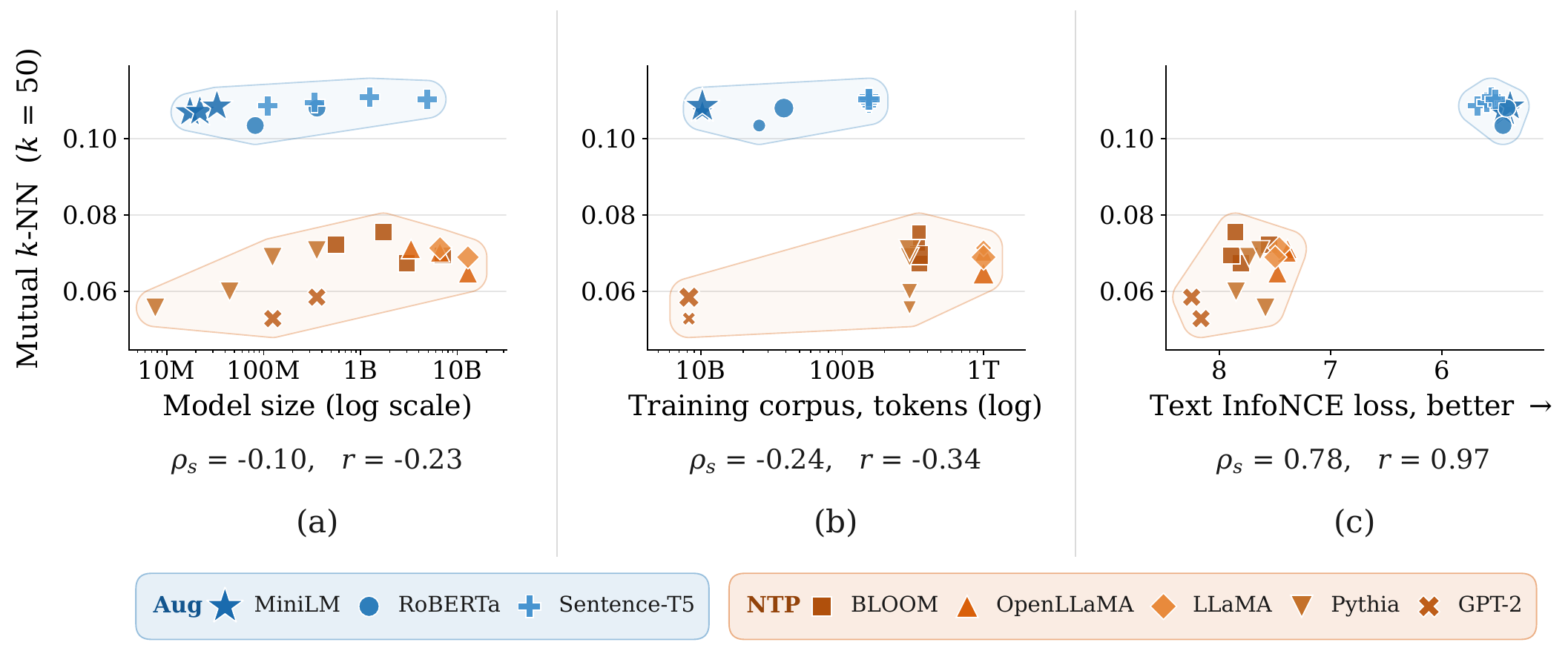}
    \caption{\textbf{Audio-text results on AudioCaps dataset.} CMA between text models and the BYOL-A audio encoder as a function of \textbf{(a) model size} (number of parameters), \textbf{(b) corpus size} (amount of tokens the model was trained on), and \textbf{(c) unimodal text InfoNCE loss}. Notably, Aug models consistently achieve higher CMA, even at substantially smaller model and corpus sizes. Also, considering all models jointly, model/corpus sizes are uncorrelated with CMA, while the unimodal loss is highly correlated, indicated by both Spearman ($\rho_s$) and Pearson ($r$) correlations.}
    \label{fig:audiocaps}
\end{figure}

Fig. \ref{fig:audiocaps}a extends Fig. \ref{fig:pretrained_audio}a from the main text to the AudioCaps dataset, while Fig. \ref{fig:audiocaps}b extends Fig. \ref{fig:pretrained_audio}b. Consistent with our primary findings, augmentation-based text models (Aug) consistently achieve higher cross-modal alignment on AudioCaps than next-token prediction (NTP) models of comparable or even substantially larger sizes. These results further support the role of unimodal augmentation objectives as a fundamental driver of implicit cross-modal alignment.

Fig. \ref{fig:audiocaps}c extends Fig. \ref{fig:pretrained_audio}c from the main text to the AudioCaps dataset. The strong correlation between lower unimodal text Aug-SSL InfoNCE loss and higher cross-modal alignment remains robust on AudioCaps. This finding further supports unimodal InfoNCE loss as a predictor of cross-modal alignment quality.

\section{Statistical Significance} \label{app:significance}

In Sec.~\ref{sec:results:pretrained}, the analysis of unconstrained, pre-trained models reports Spearman
($\rho_s$) and Pearson ($r$) correlations between CMA and unimodal model properties over the available encoders. To demonstrate that these correlations are not an artifact of the sample size,
this section establishes their statistical significance.

\definecolor{pythagsig}{HTML}{2E8B57}
\definecolor{pythagns}{HTML}{D9600A}
\providecommand{\pythagyes}{\textcolor{pythagsig}{\ding{51}}}
\providecommand{\pythagno}{\textcolor{pythagns}{\ding{55}}}
\begin{table}[h]
  \centering
  \caption{\textbf{Significance of the correlations in the three pre-trained figures \ref{fig:pretrained_uncorr}, \ref{fig:pretrained_loss}, \ref{fig:pretrained_audio}.} Size and corpus size are correlated on $\log_{10}$ and the loss as $-\mathcal{L}$. $p$ is two-sided from 200,000 permutations of the alignment scores against the property; \pythagyes{} marks $p<0.05$ after Holm correction over the 10 panels, \pythagno{} that it does not. Only the correlation with the unimodal InfoNCE loss is statistically significant.}
  \vspace{2pt}
  \label{tab:corr-signif}
  \begin{adjustbox}{max width=\linewidth}
    \begin{tabular}{llr r@{~}c r r@{~}c}
      \toprule
      & & \multicolumn{3}{c}{Spearman} & \multicolumn{3}{c}{Pearson} \\
      \cmidrule(lr){3-5} \cmidrule(lr){6-8}
      Panel & Property & $\rho_s$ & $p$ & & $r$ & $p$ & \\
      \midrule
      \multicolumn{8}{@{}l}{\emph{Fig.~\ref{fig:pretrained_uncorr}} --- text encoders against DINOv2 ViT-B/14} \\
      \quad (a)~MS-COCO & Model size ($\log_{10}$) & $-0.030$ & $0.889$ & \pythagno & $-0.081$ & $0.706$ & \pythagno \\
      \quad (b)~MS-COCO & Training-corpus size ($\log_{10}$) & $-0.233$ & $0.282$ & \pythagno & $-0.360$ & $0.092$ & \pythagno \\
      \quad (b)~WIT & Training-corpus size ($\log_{10}$) & $-0.176$ & $0.421$ & \pythagno & $-0.173$ & $0.431$ & \pythagno \\
      \addlinespace
      \multicolumn{8}{@{}l}{\emph{Fig.~\ref{fig:pretrained_loss}} --- text encoders against DINOv2 ViT-B/14 in (a); image towers against BLOOM-7b1 in (b)} \\
      \quad (a)~MS-COCO & Text InfoNCE loss ($-\mathcal{L}$) & $\phantom{-}0.880$ & $<\!5.0\!\times\!10^{-6}$ & \pythagyes & $\phantom{-}0.925$ & $<\!5.0\!\times\!10^{-6}$ & \pythagyes \\
      \quad (a)~WIT & Text InfoNCE loss ($-\mathcal{L}$) & $\phantom{-}0.820$ & $1.5\!\times\!10^{-5}$ & \pythagyes & $\phantom{-}0.857$ & $<\!5.0\!\times\!10^{-6}$ & \pythagyes \\
      \quad (b)~MS-COCO & Vision InfoNCE loss ($-\mathcal{L}$) & $\phantom{-}0.855$ & $2.6\!\times\!10^{-4}$ & \pythagyes & $\phantom{-}0.939$ & $<\!5.0\!\times\!10^{-6}$ & \pythagyes \\
      \quad (b)~WIT & Vision InfoNCE loss ($-\mathcal{L}$) & $\phantom{-}0.890$ & $5.0\!\times\!10^{-5}$ & \pythagyes & $\phantom{-}0.896$ & $<\!5.0\!\times\!10^{-6}$ & \pythagyes \\
      \addlinespace
      \multicolumn{8}{@{}l}{\emph{Fig.~\ref{fig:pretrained_audio}} --- text encoders against BYOL-A v2} \\
      \quad (a)~Clotho & Model size ($\log_{10}$) & $\phantom{-}0.043$ & $0.837$ & \pythagno & $\phantom{-}0.149$ & $0.486$ & \pythagno \\
      \quad (b)~Clotho & Training-corpus size ($\log_{10}$) & $-0.196$ & $0.368$ & \pythagno & $-0.128$ & $0.561$ & \pythagno \\
      \quad (c)~Clotho & Text InfoNCE loss ($-\mathcal{L}$) & $\phantom{-}0.920$ & $<\!5.0\!\times\!10^{-6}$ & \pythagyes & $\phantom{-}0.860$ & $<\!5.0\!\times\!10^{-6}$ & \pythagyes \\
      \bottomrule
    \end{tabular}
  \end{adjustbox}
\end{table}
 
For each panel, we test the null hypothesis that the alignment scores are exchangeable with respect
to the property under consideration. We obtain a two-sided $p$-value from $200{,}000$ random
permutations of the alignment scores against the property, where model size and training-corpus size
are taken on a $\log_{10}$ scale and the InfoNCE loss as $-\mathcal{L}$, matching the axes of the
corresponding figures. Since the three figures comprise 10 panels tested jointly, we control the
family-wise error rate using the Holm correction and consider a correlation significant when its
corrected $p$-value is below $0.05$. Values reported as $<\!5.0\!\times\!10^{-6}$ lie at the
resolution limit of the permutation test.
 
Table~\ref{tab:corr-signif} reports the results for the three pre-trained figures
(\ref{fig:pretrained_uncorr}, \ref{fig:pretrained_loss}, \ref{fig:pretrained_audio}). Consistent
with our primary findings, every correlation between the unimodal InfoNCE loss and CMA is
significant, for both Spearman and Pearson, and across vision--text and audio--text alike, with all
$p$-values below 2.6x$10^{-4}$ before correction. This confirms that the unimodal Aug-SSL loss is a
reliable predictor of cross-modal alignment, and that this relation is not driven by a small number
of individual models.
 
In contrast, the correlations with model and corpus size are inconsistent. They are
consistently insignificant across datasets and modalities. Notably, in no panel do we find evidence that larger models or larger
training corpora yield higher CMA. Together with the loss results, this further supports that
unimodal augmentation objectives, rather than scale, act as a fundamental driver of implicit
cross-modal alignment.

\section{Additional Text Augmentations} \label{app:other_llm}

In Sec. \ref{sec:results}, our evaluation of the text InfoNCE utilized LLM-augmentations generated by Qwen2.5. However, this holds universally across different text augmentations. This section extends the analysis in the main text to different LLM augmentations and back-translation augmentations. For the LLM augmentations, we used two additional LLMs (Phi-3.5-Mini \citep{haider2024phi} and Gemma 4 12b \citep{team2026gemma}) to generate five rephrased versions of each text. For the back-translation, we used bilingual models to transfer the English text to another language (e.g., French) and back to English. Additional details are provided in App. \ref{app:implementation_details}.

\begin{figure}
    \centering
    \includegraphics[width=0.9\linewidth]{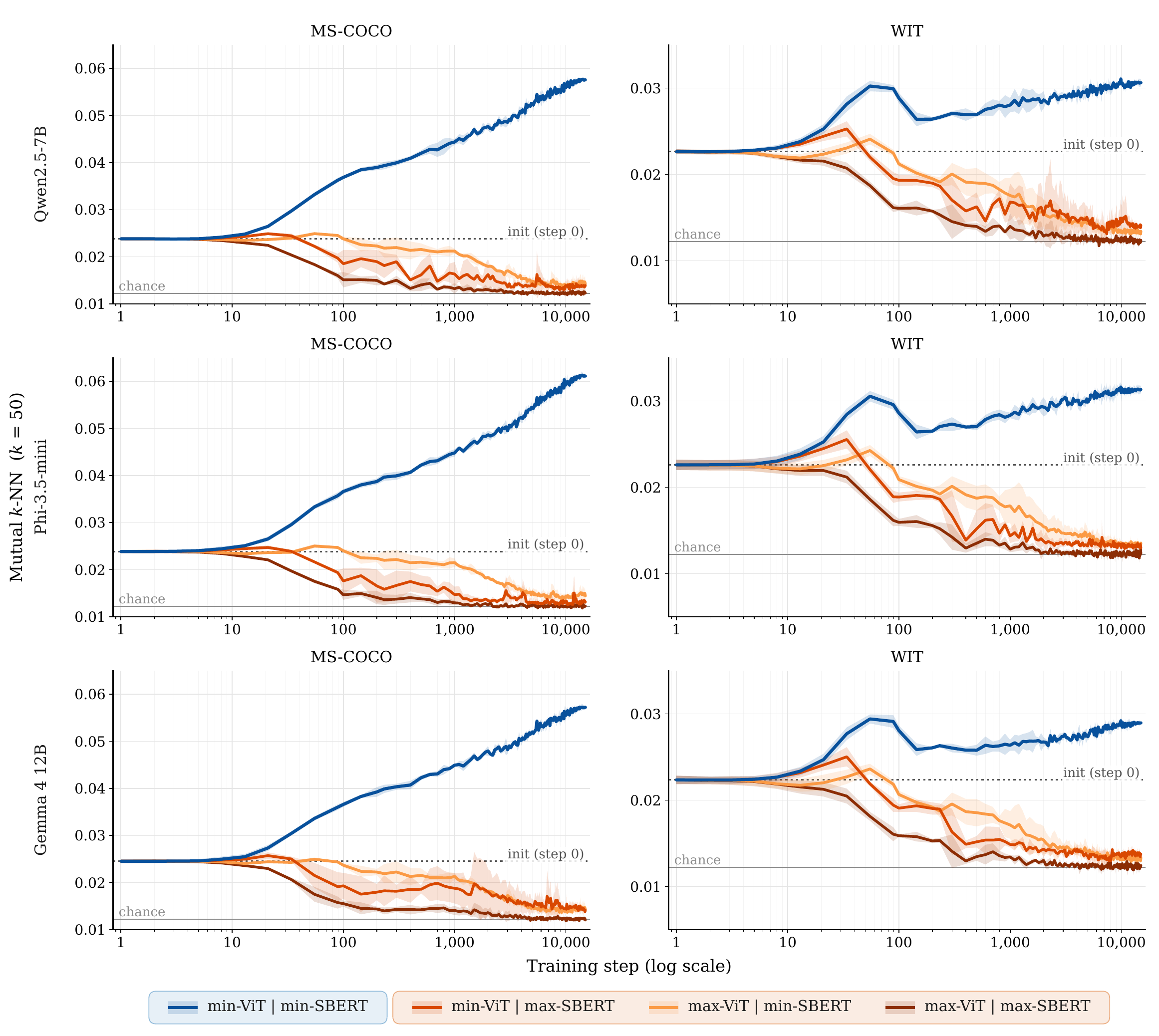}
    \caption{\textbf{ Extended Empirical validation of Thm. \ref{thm:align_main} and \ref{thm:misalign_main} with different LLMs augmentations.} Vision (ViT) and text (SBERT) models trained to minimize/maximize the Aug-SSL loss (Eq. \ref{eq:infonce-k}) using Phi-3.5-Mini and Gemma 4 12B text augmentations. Minimizing both unimodal losses improves CMA (Thm. \ref{thm:align_main}), whereas maximizing either loss prevents alignment, with CMA converging to the chance level (Thm. \ref{thm:misalign_main}).}
    \label{fig:minmax_otherllms}
\end{figure}

\begin{figure}
    \centering
    \includegraphics[width=0.9\linewidth]{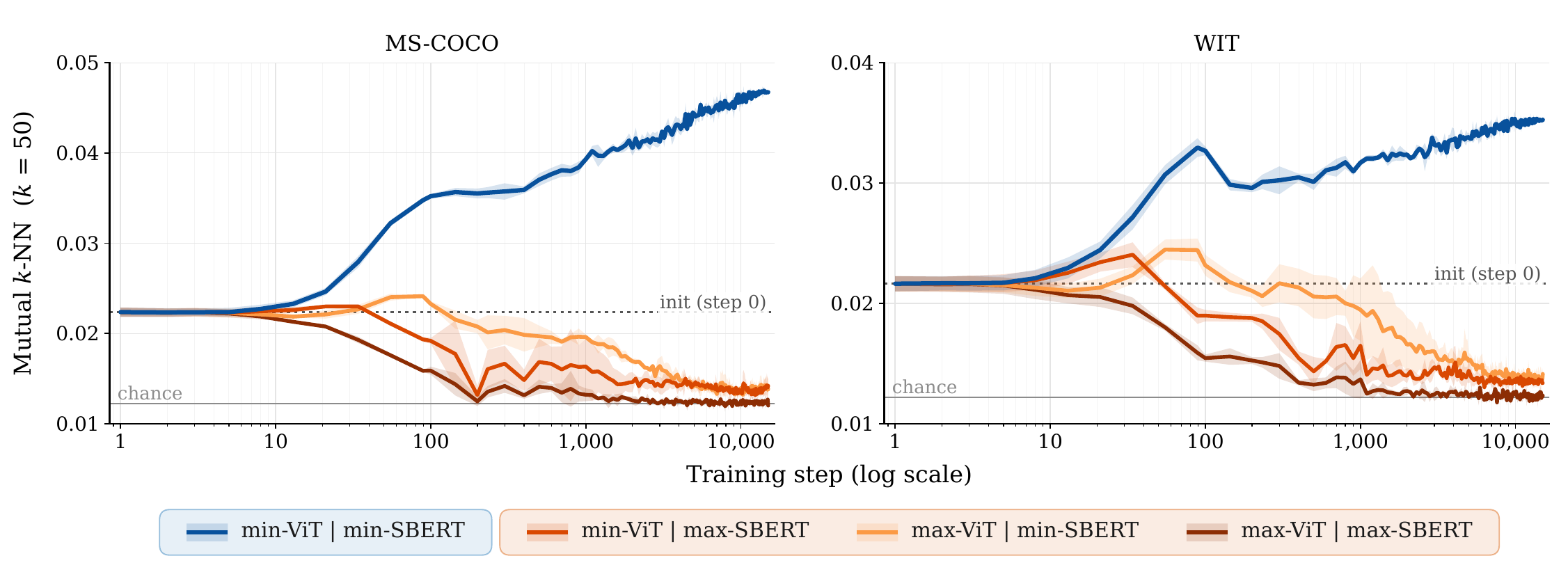}
    \caption{\textbf{Extended Empirical validation of Thm. \ref{thm:align_main} and \ref{thm:misalign_main} with back-translation augmentations.} Vision (ViT) and text (SBERT) models trained to minimize/maximize the Aug-SSL loss (Eq. \ref{eq:infonce-k}) using back-translation text augmentations. Minimizing both unimodal losses improves CMA (Thm. \ref{thm:align_main}), whereas maximizing either loss prevents alignment, with CMA converging to the chance level (Thm. \ref{thm:misalign_main}).}
    \label{fig:minmax_backtrans}
\end{figure}

Fig. \ref{fig:minmax_otherllms} and \ref{fig:minmax_backtrans} extend Fig. \ref{fig:controlled_results}a from the main text to the two additional LLMs and back-translation augmentations, respectively. Consistent with our primary findings, the results support Thm.~\ref{thm:align_main}: minimizing the unimodal Aug-SSL objectives leads to increasing CMA. In contrast, every pairing involving a loss-maximized encoder is consistent with Thm.~\ref{thm:misalign_main}, showing that a high unimodal loss in either modality prevents cross-modal alignment, with CMA approaching chance level.

\begin{figure}
    \centering
    \includegraphics[width=0.75\linewidth]{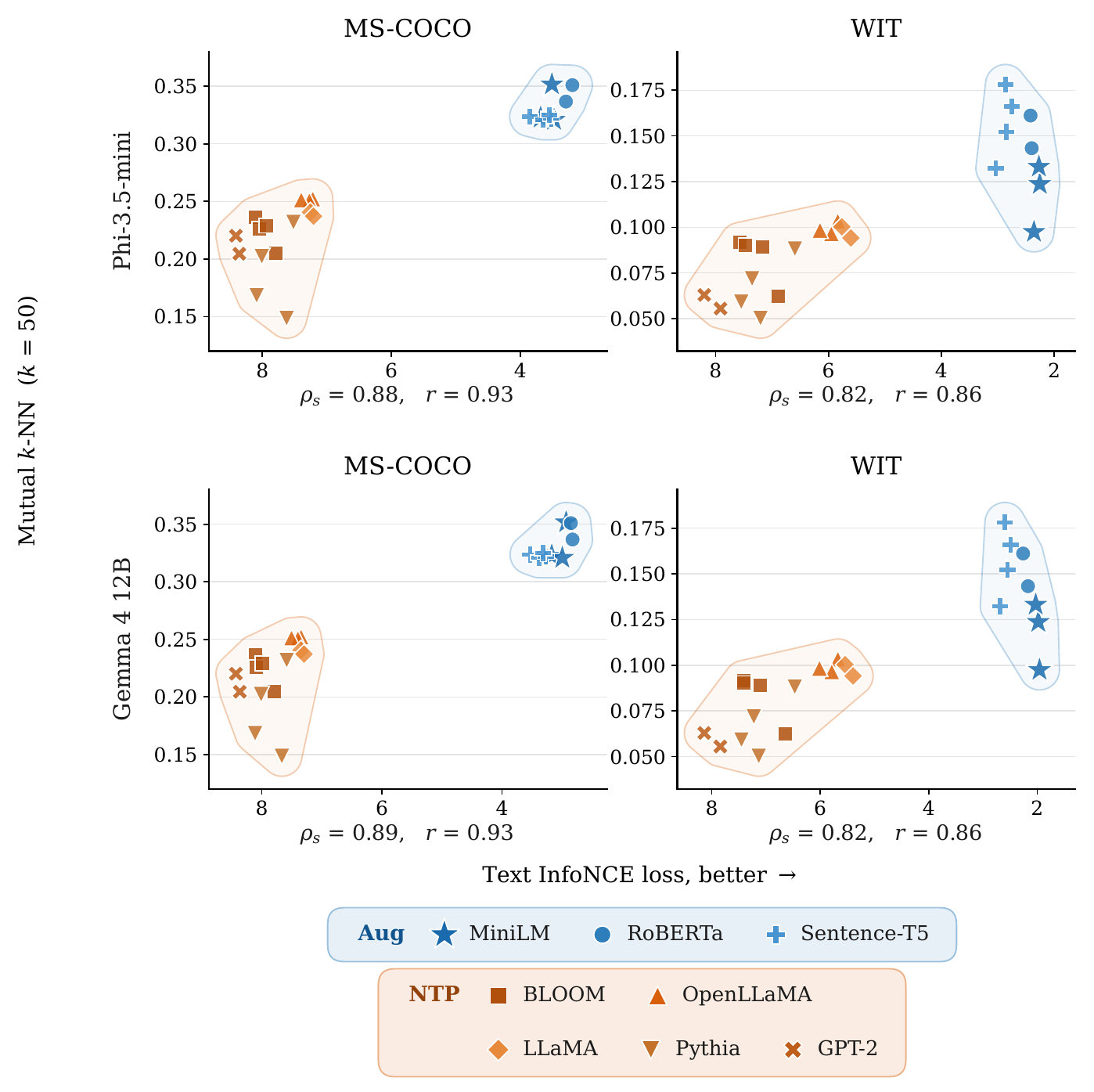}
    \caption{\textbf{CMA by InfoNCE across LLMs text augmentations on the COCO and WIT datasets.} CMA by unimodal text InfoNCE loss across different LLMs (Phi-3.5-mini, Gemma 4 12B) augmentations. Notably, the unimodal Aug-SSL loss is highly correlated with CMA, clearly separating low- and high-alignment models, and having large Spearman ($\rho_s$) and Pearson ($r$) correlations.}
    \label{fig:infonce_otherllms}
\end{figure}

\begin{figure}
    \centering
    \includegraphics[width=0.75\linewidth]{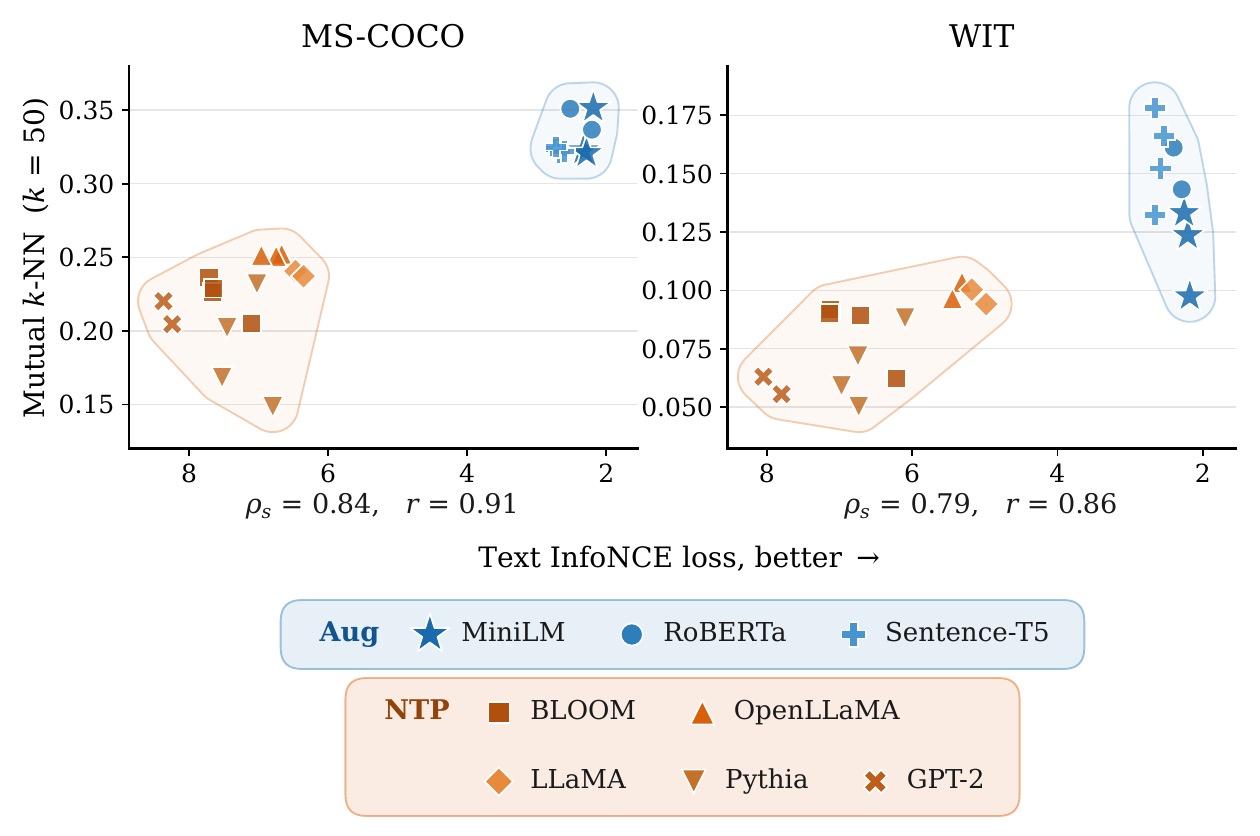}
    \caption{\textbf{CMA by InfoNCE across back-translation text augmentations on the COCO and WIT datasets.} CMA by unimodal text InfoNCE loss using back-translation augmentations. Notably, the unimodal Aug-SSL loss is highly correlated with CMA, clearly separating low- and high-alignment models, and having large Spearman ($\rho_s$) and Pearson ($r$) correlations.}
    \label{fig:infonce_backtrans}
\end{figure}

Fig. \ref{fig:infonce_otherllms} and \ref{fig:infonce_backtrans} extend Fig. \ref{fig:pretrained_loss}a from the main text to two additional LLMs and back-translation augmentations, respectively. The strong correlation between lower unimodal text Aug-SSL InfoNCE loss and higher cross-modal alignment remains robust across LLMs used to generate the augmentations.

\section{Different Vision Encoders} \label{app:more_vision_encoders}

In Sec. \ref{sec:results:pretrained}, the evaluation of CMA to vision models focused on DINOv2, and to text models on BLOOM. This section shows that the same trends hold across different vision and text architectures and training paradigms. In particular, it extends the analysis to a broader suite of 14 vision models and 24 text models, spanning the model families discussed in Sec. \ref{sec:results:pretrained}.

\begin{figure}
    \centering
    \includegraphics[width=1\linewidth]{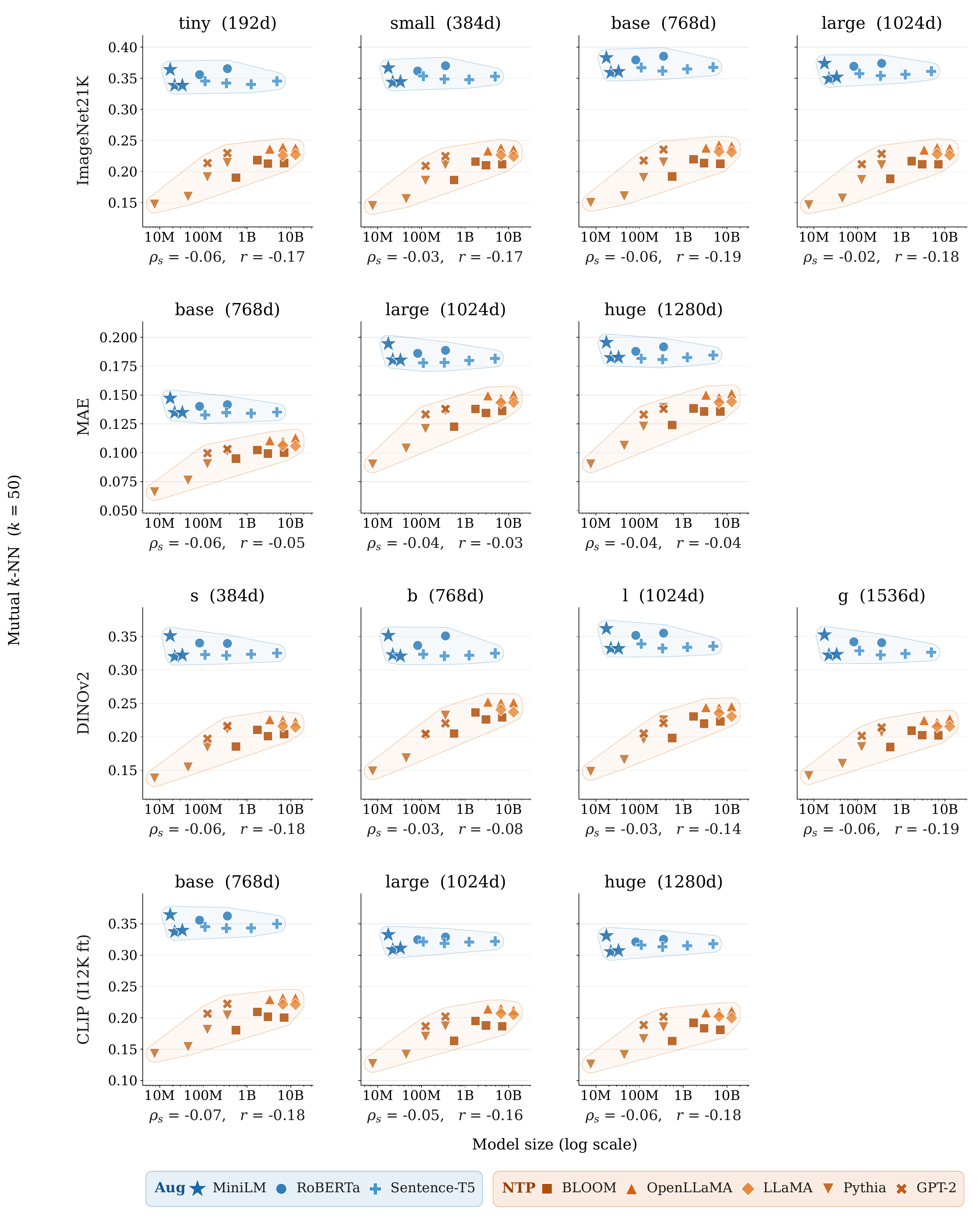}
    \caption{\textbf{CMA across vision architectures on the COCO dataset.} CMA by model size (number of parameters) across different vision encoder families (ImageNet21K, MAE, DINOv2, and CLIP) at various capacity scales. Notably, Aug models consistently achieve higher CMA, even at substantially smaller model and corpus sizes. Also, considering all models jointly, model/corpus sizes are uncorrelated with CMA, indicated by both weak Spearman ($\rho_s$) and Pearson ($r$) correlations.}
    \label{fig:parameters_zoo_coco}
\end{figure}

\begin{figure}
    \centering
    \includegraphics[width=1\linewidth]{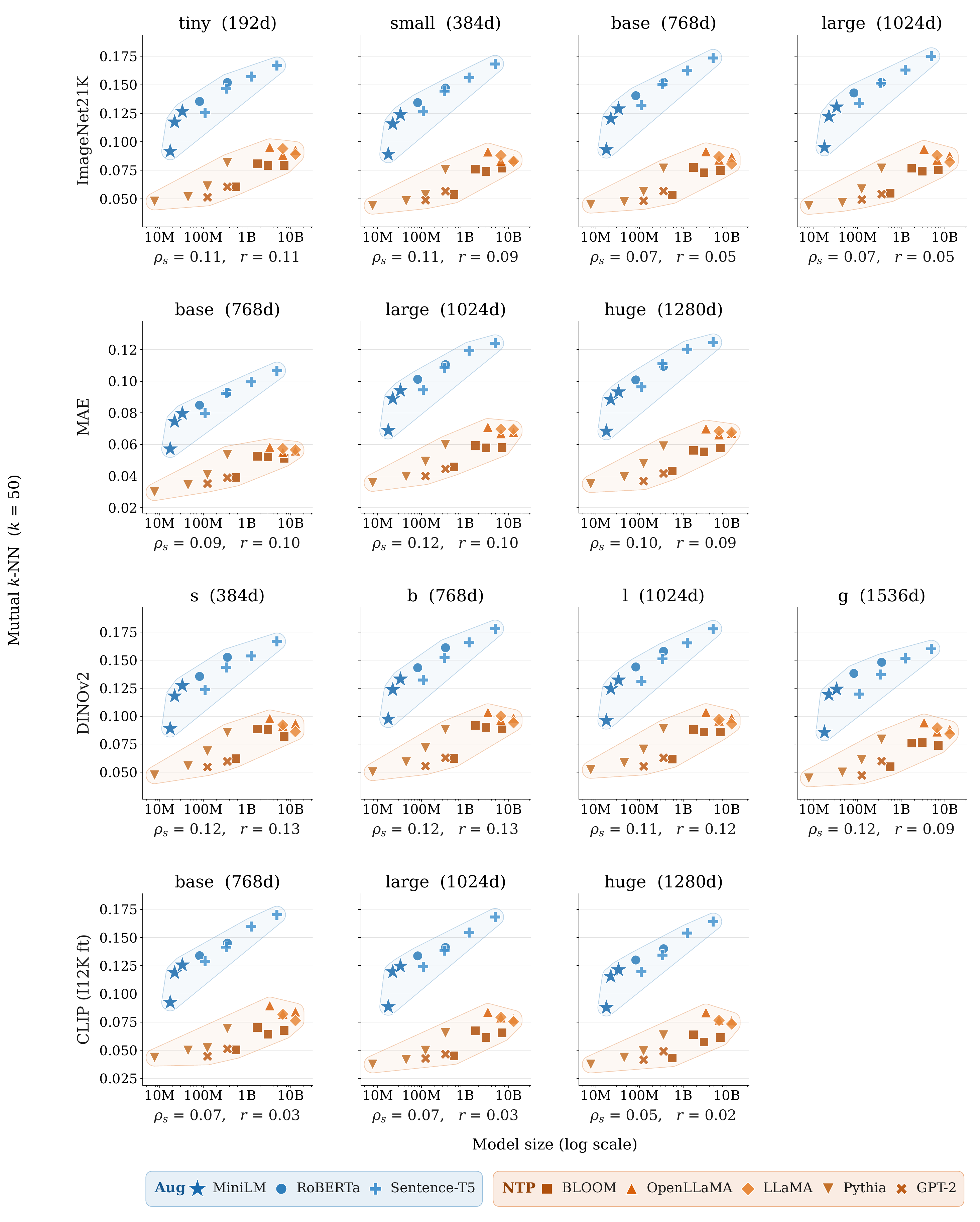}
    \caption{\textbf{CMA across vision architectures on the WIT dataset.} CMA by model size (number of parameters) across different vision encoder families (ImageNet21K, MAE, DINOv2, and CLIP) at various capacity scales. Notably, Aug models consistently achieve higher CMA, even at substantially smaller model and corpus sizes. Also, considering all models jointly, model/corpus sizes are uncorrelated with CMA, indicated by both weak Spearman ($\rho_s$) and Pearson ($r$) correlations.}
    \label{fig:parameters_zoo_wit}
\end{figure}

Fig. \ref{fig:parameters_zoo_coco} extends Fig. \ref{fig:pretrained_uncorr}a to the 14 vision models, while Fig. \ref{fig:parameters_zoo_wit} extends Fig. \ref{fig:opening}. Consistent with our primary findings, the results demonstrate that augmentation-based text models (Aug) consistently achieve higher cross-modal alignment across all tested vision encoders compared to next-token prediction (NTP) models of comparable or even substantially larger sizes.

\begin{figure}
    \centering
    \includegraphics[width=1\linewidth]{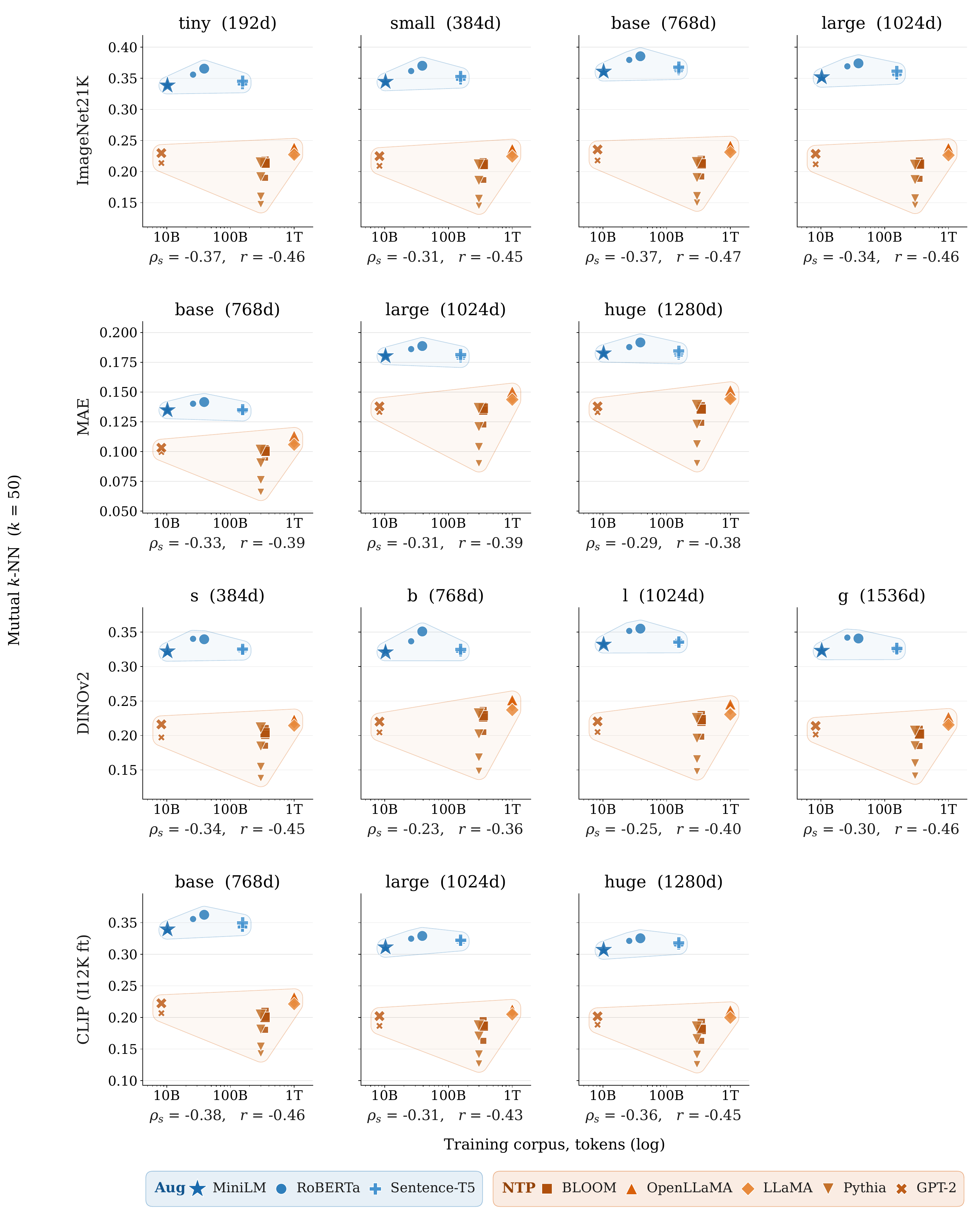}
    \caption{\textbf{CMA by corpus size across vision architectures on the COCO dataset.} CMA by corpus size (tokens trained on) across different vision encoder families (ImageNet21K, MAE, DINOv2, and CLIP) at various capacity scales. Notably, Aug models consistently achieve higher CMA, even at substantially smaller model and corpus sizes. Also, considering all models jointly, model/corpus sizes are uncorrelated with CMA, indicated by both weak Spearman ($\rho_s$) and Pearson ($r$) correlations.}
    \label{fig:data_zoo_coco}
\end{figure}

\begin{figure}
    \centering
    \includegraphics[width=1\linewidth]{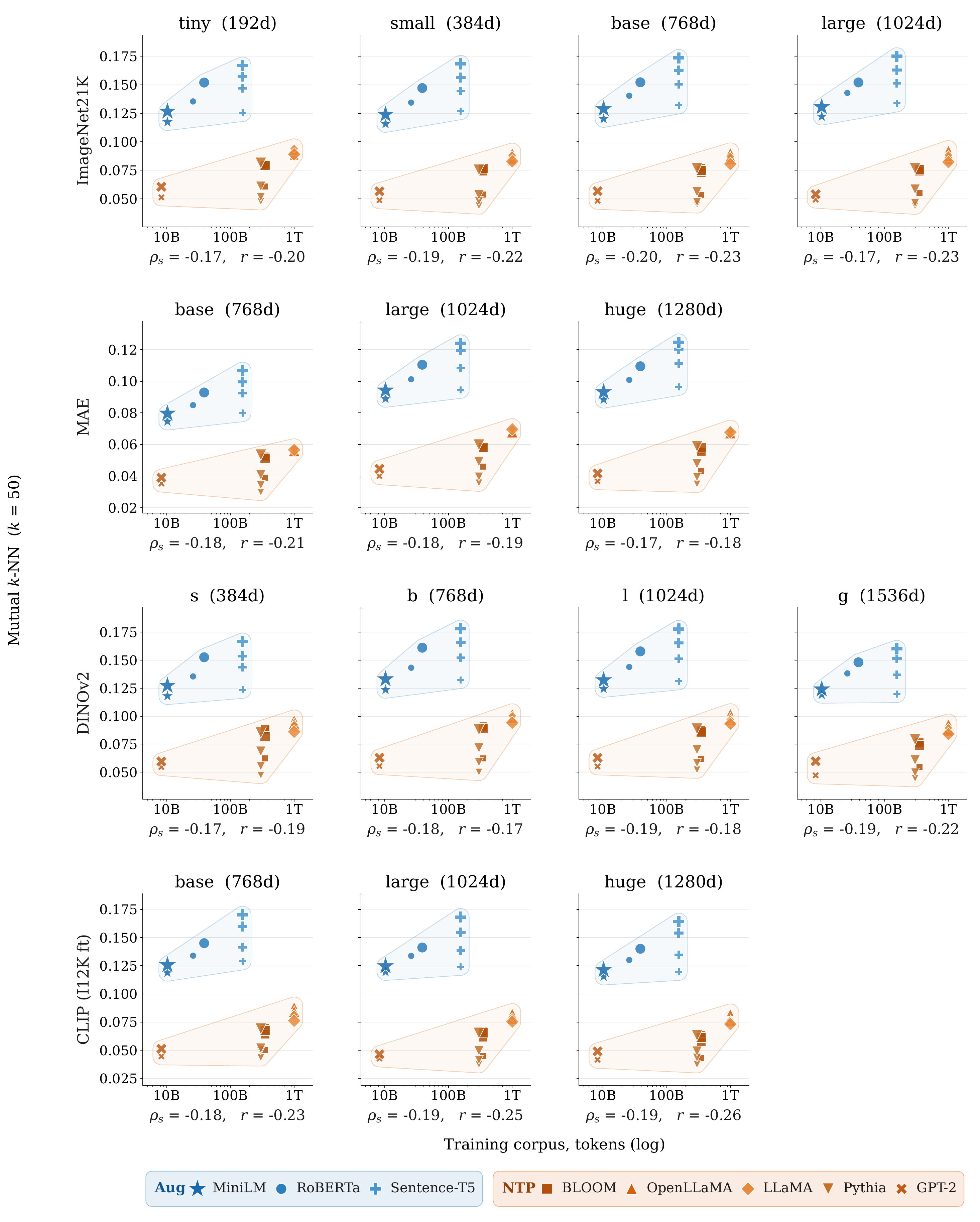}
    \caption{\textbf{CMA by corpus size across vision architectures on the WIT dataset.} CMA by corpus size (tokens trained on) across different vision encoder families (ImageNet21K, MAE, DINOv2, and CLIP) at various capacity scales. Notably, Aug models consistently achieve higher CMA, even at substantially smaller model and corpus sizes. Also, considering all models jointly, model/corpus sizes are uncorrelated with CMA, indicated by both weak Spearman ($\rho_s$) and Pearson ($r$) correlations.}
    \label{fig:data_zoo_wit}
\end{figure}

Fig. \ref{fig:data_zoo_coco} and \ref{fig:data_zoo_wit} extend Fig. \ref{fig:pretrained_uncorr}b to the 14 vision models, on COCO and WIT respectively. Consistent with our primary findings, the results demonstrate that augmentation-based text models (Aug) consistently achieve higher cross-modal alignment across all tested vision encoders compared to next-token prediction (NTP) models of comparable or even substantially smaller model and corpus sizes.

\begin{figure}
    \centering
    \includegraphics[width=1\linewidth]{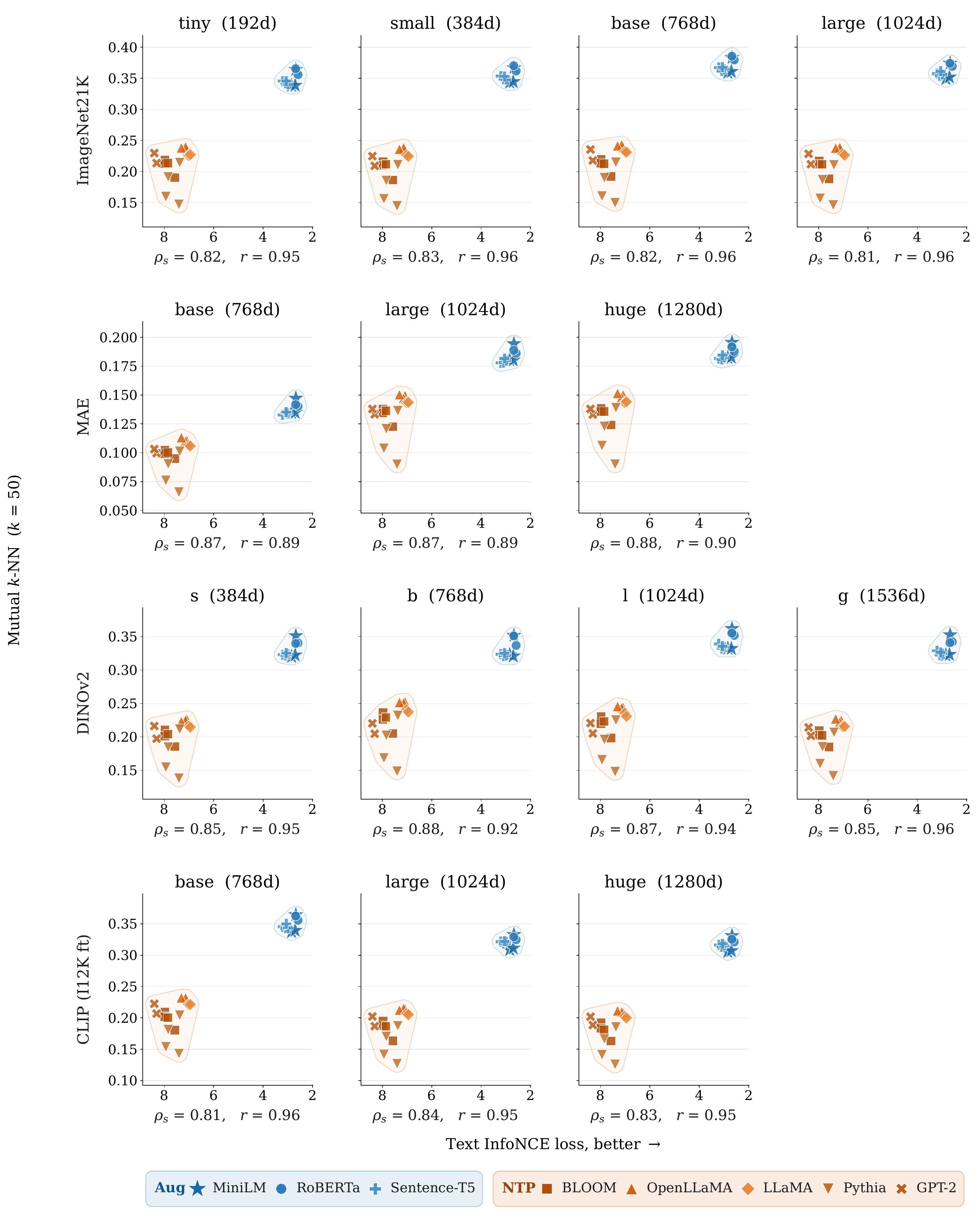}
    \caption{\textbf{CMA by InfoNCE across vision architectures on the COCO dataset.} CMA by unimodal text InfoNCE loss across different vision encoder families (ImageNet21K, MAE, DINOv2, and CLIP) at various capacity scales. Notably, the unimodal Aug-SSL loss is highly correlated with CMA, clearly separating low- and high-alignment models, and having large Spearman ($\rho_s$) and Pearson ($r$) correlations.}
    \label{fig:loss_zoo_coco}
\end{figure}

\begin{figure}
    \centering
    \includegraphics[width=1\linewidth]{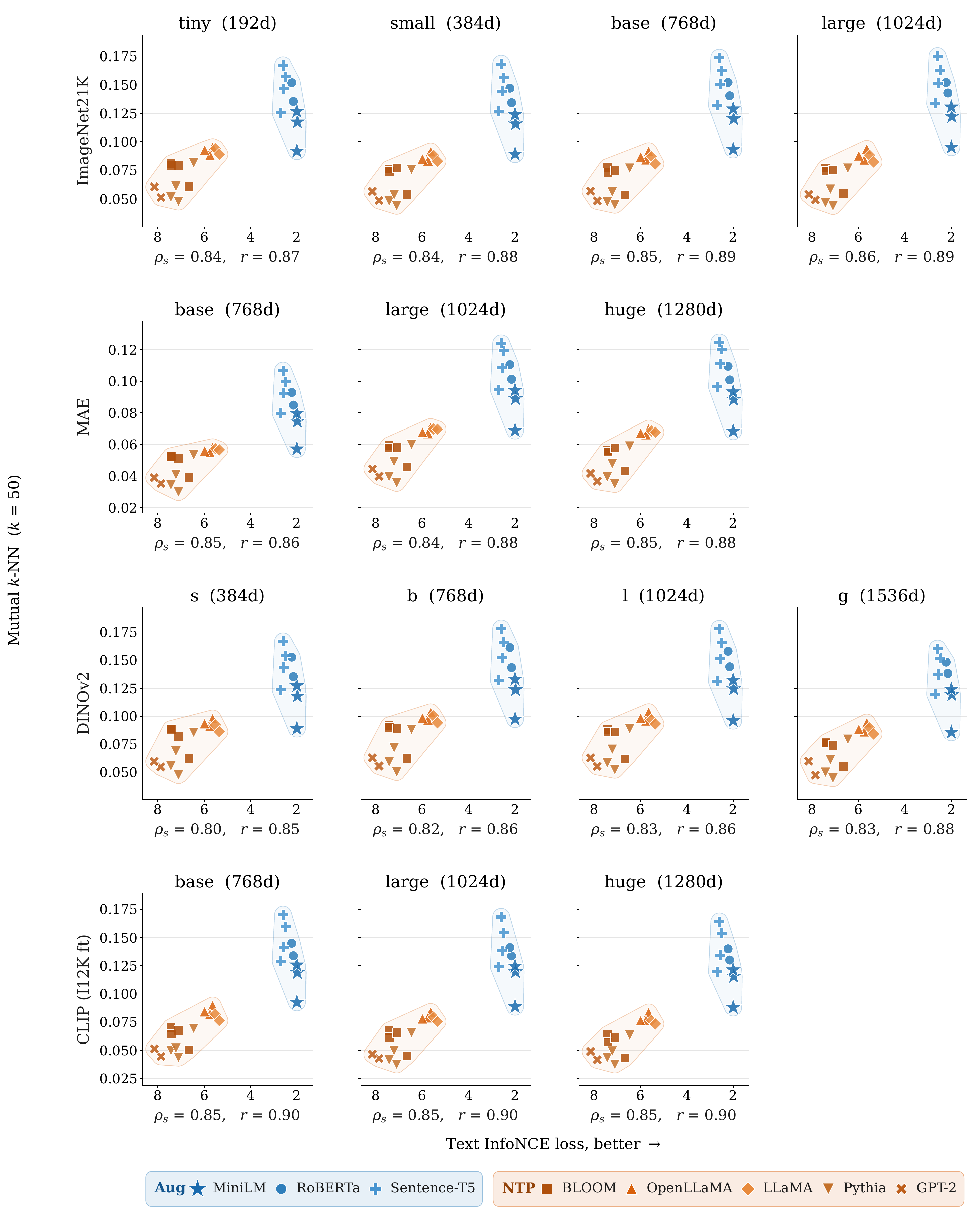}
    \caption{\textbf{CMA by InfoNCE across vision architectures on the WIT dataset.} CMA by unimodal text InfoNCE loss across different vision encoder families (ImageNet21K, MAE, DINOv2, and CLIP) at various capacity scales. Notably, the unimodal Aug-SSL loss is highly correlated with CMA, clearly separating low- and high-alignment models, and having large Spearman ($\rho_s$) and Pearson ($r$) correlations.}
    \label{fig:loss_zoo_wit}
\end{figure}

Fig. \ref{fig:loss_zoo_coco} and \ref{fig:loss_zoo_wit} extend Fig. \ref{fig:pretrained_loss}a to the 14 vision models, on COCO and on WIT, respectively. The strong correlation between lower unimodal text Aug-SSL InfoNCE loss and higher cross-modal alignment remains robust regardless of the specific vision architecture or its scale.

\begin{figure}
    \centering
    \vspace{-10pt}
    \includegraphics[width=0.8\linewidth]{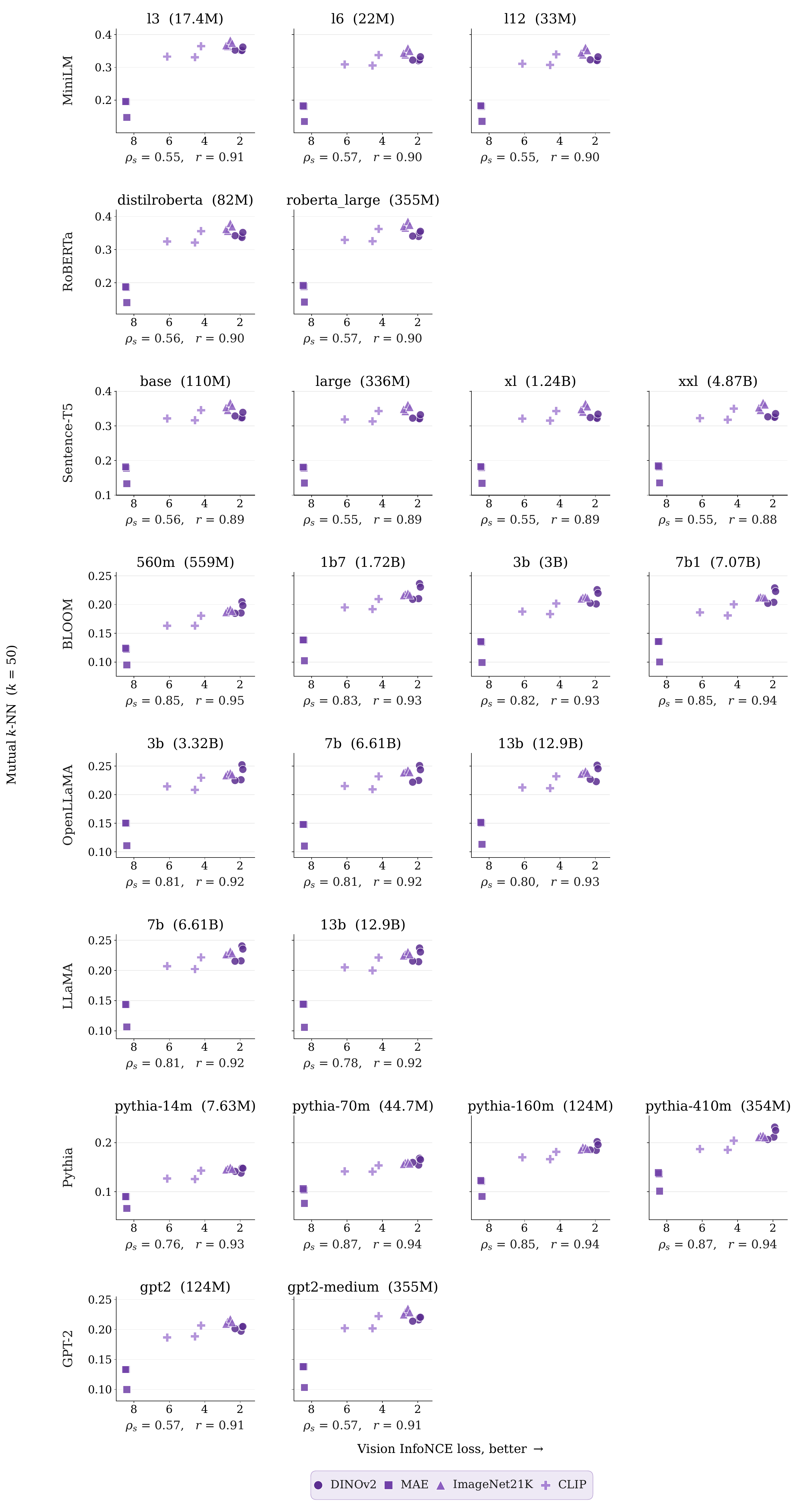}
    \caption{\textbf{CMA by Vision InfoNCE across text architectures on the COCO dataset.} CMA by unimodal vision InfoNCE loss across different text encoder families (MiniLM, RoBERTa, Sentence-T5, BLOOM, OpenLLaMA, LLaMA, Pythia, and GPT-2) at various capacity scales. Notably, the unimodal Aug-SSL loss is highly correlated with CMA, clearly separating low- and high-alignment models, and having large Spearman ($\rho_s$) and Pearson ($r$) correlations.}
    \label{fig:loss_vision_zoo_coco}
\end{figure}

\begin{figure}
    \centering
    \vspace{-10pt}
    \includegraphics[width=0.8\linewidth]{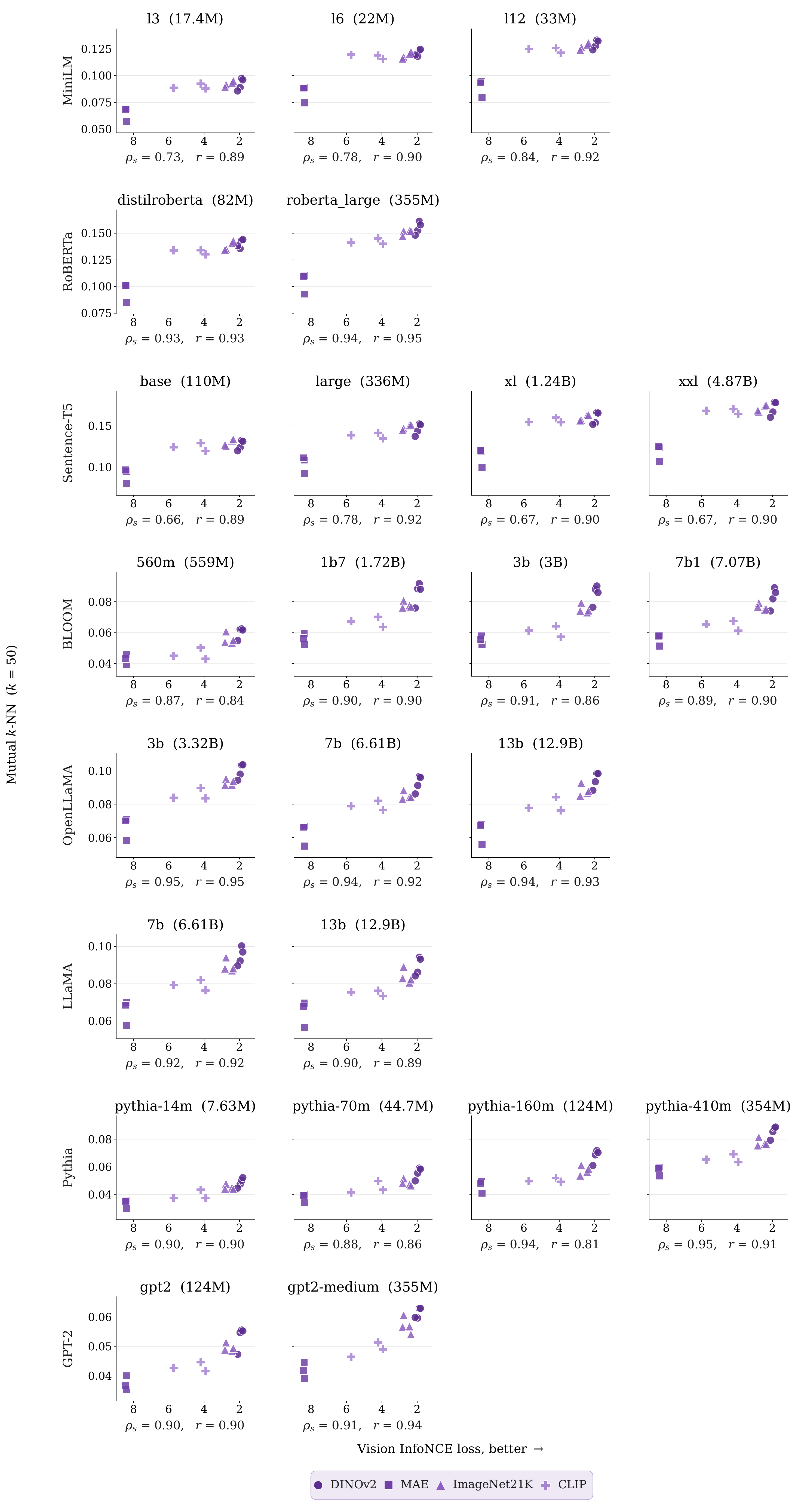}
    \caption{\textbf{CMA by Vision InfoNCE across text architectures on the WIT dataset.} CMA by unimodal vision InfoNCE loss across different text encoder families (MiniLM, RoBERTa, Sentence-T5, BLOOM, OpenLLaMA, LLaMA, Pythia,  and GPT-2) at various capacity scales. Notably, the unimodal Aug-SSL loss is highly correlated with CMA, clearly separating low- and high-alignment models, and having large Spearman ($\rho_s$) and Pearson ($r$) correlations.}
    \label{fig:loss_vision_zoo_wit}
\end{figure}

Similarly, Fig. \ref{fig:loss_vision_zoo_coco} and \ref{fig:loss_vision_zoo_wit} extend Fig. \ref{fig:pretrained_loss}b to the 24 text models, on COCO and on WIT, respectively. The strong correlation between lower unimodal vision Aug-SSL InfoNCE loss and higher cross-modal alignment remains robust regardless of the specific text architecture or its scale.

\section{Different Audio Encoders} \label{app:more_audio_encoders}

In Sec. \ref{sec:results:pretrained}, the evaluation of audio models focused on BYOL-A-v2. This section extends our analysis to a broader suite of 15 audio models, spanning the six model families: wav2vec2 \citep{baevski2020wav2vec}, HuBERT \citep{hsu2021hubert}, WavLM \citep{chen2022wavlm}, data2vec \citep{baevski2022data2vec}, AudioMAE \citep{huang2022masked}, BYOL-A \citep{niizumi2021byol}.

\begin{figure}
    \centering
    \includegraphics[width=\linewidth]{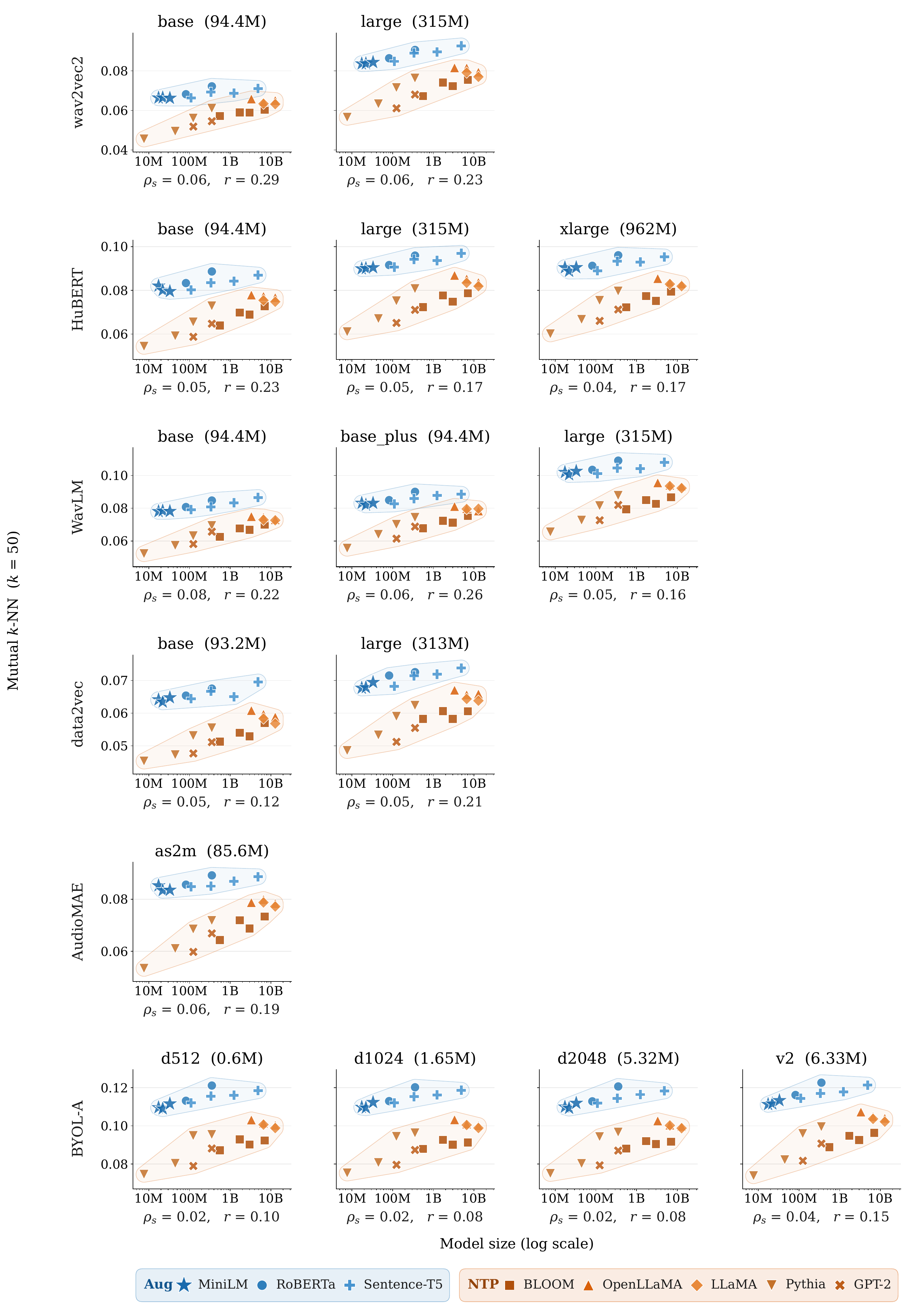}
    \caption{\textbf{CMA across audio architectures on the CLOTHO dataset.} CMA by model size (number of parameters) across different audio encoder families (wav2vec2, HuBERT, WavLM, data2vec, AudioMAE, and BYOL-A) at various capacity scales. Notably, Aug models consistently achieve higher CMA, even at substantially smaller model and corpus sizes. Also, considering all models jointly, model/corpus sizes are uncorrelated with CMA, indicated by both weak Spearman ($\rho_s$) and Pearson ($r$) correlations.}
    \label{fig:parameters_zoo_clotho}
\end{figure}

\begin{figure}
    \centering
    \includegraphics[width=\linewidth]{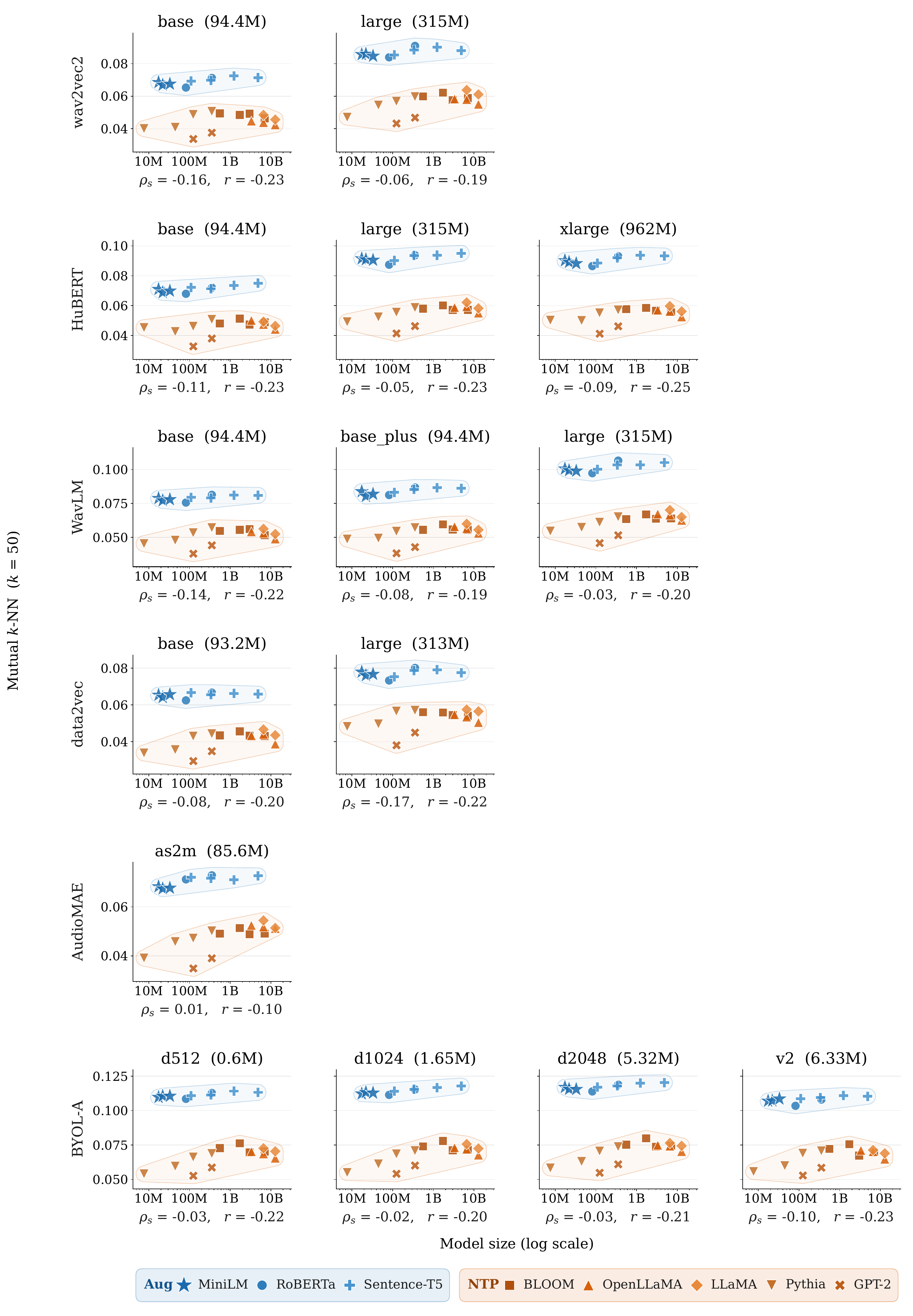}
    \caption{\textbf{CMA across audio architectures on the AudioCaps dataset.} CMA by model size (number of parameters) across different audio encoder families (wav2vec2, HuBERT, WavLM, data2vec, AudioMAE, and BYOL-A) at various capacity scales. Notably, Aug models consistently achieve higher CMA, even at substantially smaller model and corpus sizes. Also, considering all models jointly, model/corpus sizes are uncorrelated with CMA, indicated by both weak Spearman ($\rho_s$) and Pearson ($r$) correlations.}
    \label{fig:parameters_zoo_audiocaps}
\end{figure}

Fig. \ref{fig:parameters_zoo_clotho} and \ref{fig:parameters_zoo_audiocaps} extend the model size analysis (Fig. \ref{fig:pretrained_audio}a) to these additional audio models, on Clotho and AudioCaps, respectively. Consistent with our primary findings, the results demonstrate that augmentation-based text models (Aug) consistently achieve higher cross-modal alignment across all tested audio encoders compared to next-token prediction (NTP) models of comparable or even substantially larger sizes. 

\begin{figure}
    \centering
    \includegraphics[width=\linewidth]{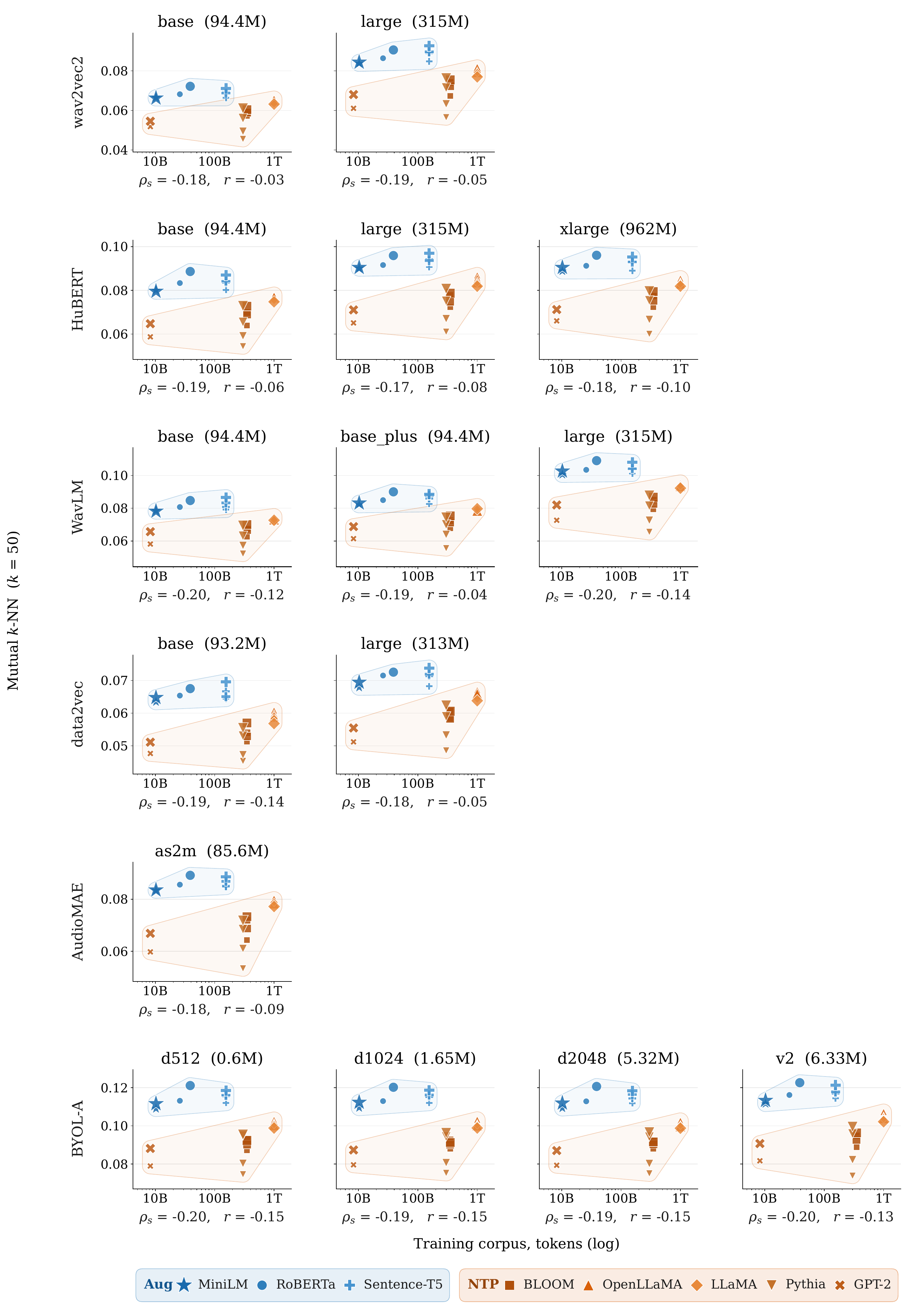}
    \caption{\textbf{CMA by corpus size across audio architectures on the CLOTHO dataset.} CMA by corpus size (tokens trained on) across different audio encoder families (wav2vec2, HuBERT, WavLM, data2vec, AudioMAE, and BYOL-A) at various capacity scales. Notably, Aug models consistently achieve higher CMA, even at substantially smaller model and corpus sizes. Also, considering all models jointly, model/corpus sizes are uncorrelated with CMA, indicated by both weak Spearman ($\rho_s$) and Pearson ($r$) correlations.}
    \label{fig:data_zoo_clotho}
\end{figure}

\begin{figure}
    \centering
    \includegraphics[width=\linewidth]{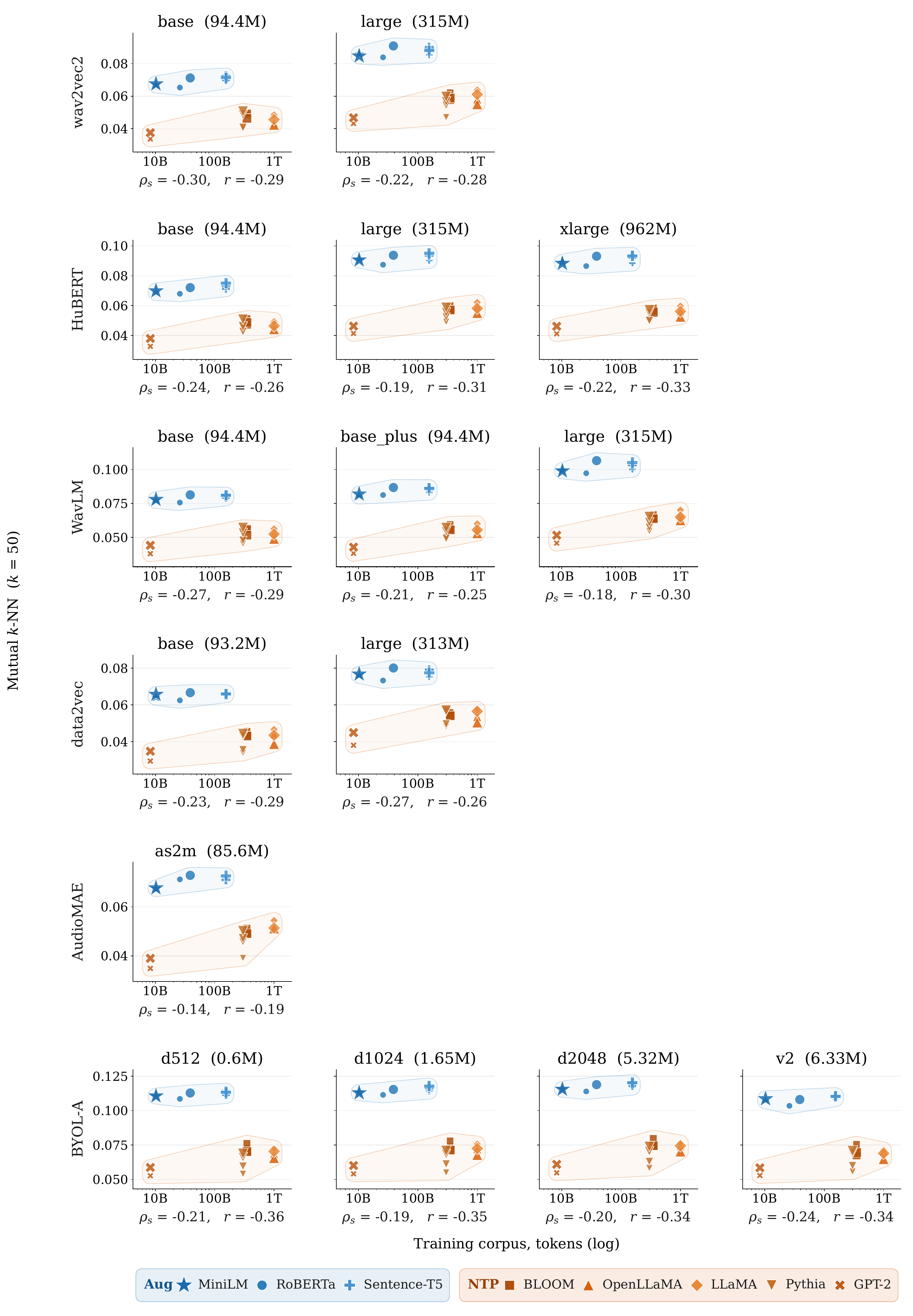}
    \caption{\textbf{CMA by corpus size across audio architectures on the AudioCaps dataset.} CMA by corpus size (tokens trained on) across different audio encoder families (wav2vec2, HuBERT, WavLM, data2vec, AudioMAE, and BYOL-A) at various capacity scales. Notably, Aug models consistently achieve higher CMA, even at substantially smaller model and corpus sizes. Also, considering all models jointly, model/corpus sizes are uncorrelated with CMA, as indicated by both weak Spearman ($\rho_s$) and Pearson ($r$) correlations.}
    \label{fig:data_zoo_audiocaps}
\end{figure}

Fig. \ref{fig:data_zoo_clotho} and \ref{fig:data_zoo_audiocaps} extend Fig. \ref{fig:pretrained_audio}b to these audio models, on Clotho and AudioCaps respectively. Consistent with our primary findings, the results demonstrate that augmentation-based text models (Aug) consistently achieve higher cross-modal alignment across all tested audio encoders compared to next-token prediction (NTP) models of comparable or even substantially smaller model and corpus sizes.

\begin{figure}
    \centering
    \includegraphics[width=\linewidth]{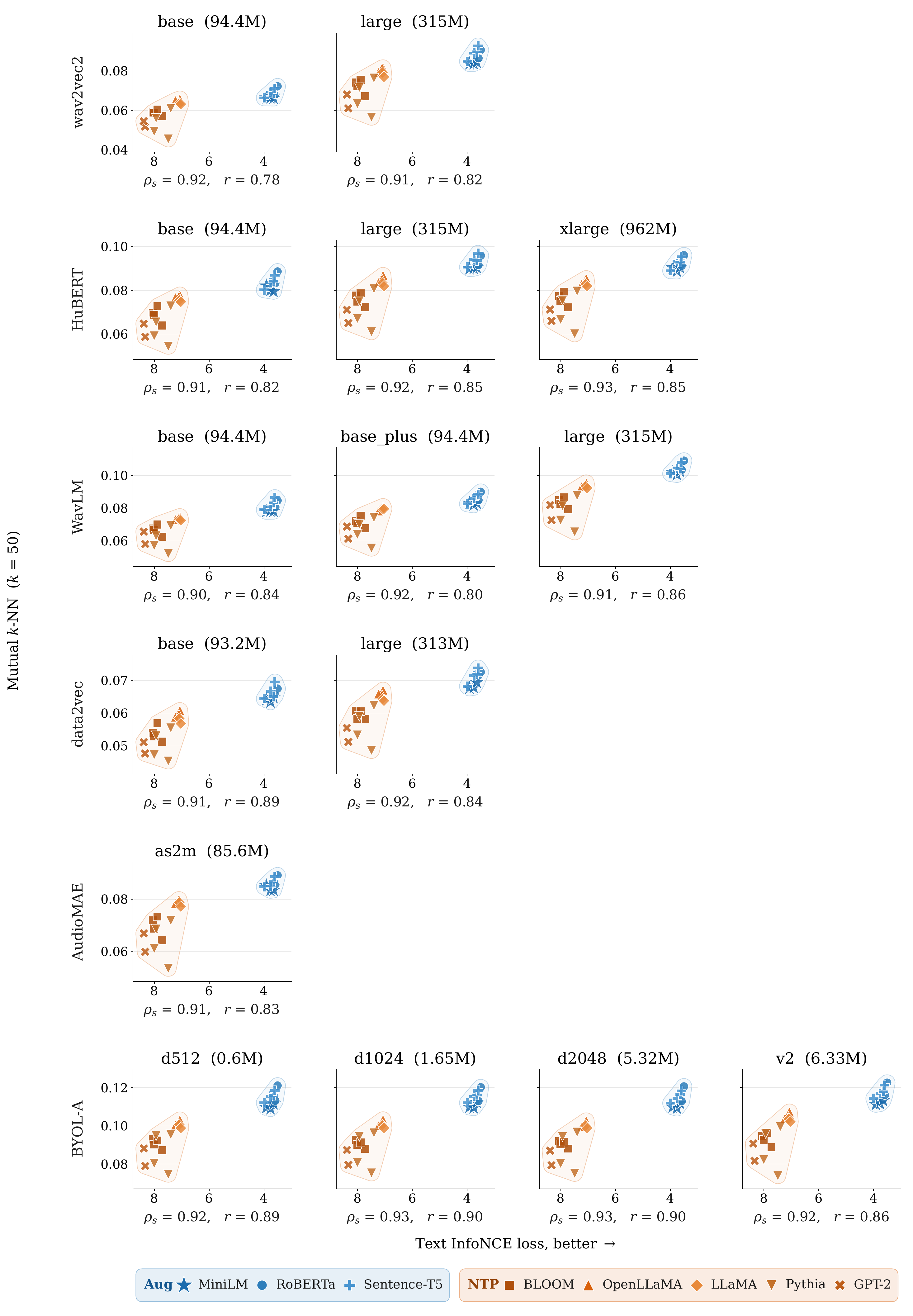}
    \caption{\textbf{CMA by InfoNCE across audio architectures on the CLOTHO dataset.} CMA by unimodal text InfoNCE loss across different audio encoder families (wav2vec2, HuBERT, WavLM, data2vec, AudioMAE, and BYOL-A) at various capacity scales. Notably, the unimodal Aug-SSL loss is highly correlated with CMA, clearly separating low- and high-alignment models, and having large Spearman ($\rho_s$) and Pearson ($r$) correlations.}
    \label{fig:loss_zoo_clotho}
\end{figure}

\begin{figure}
    \centering
    \includegraphics[width=\linewidth]{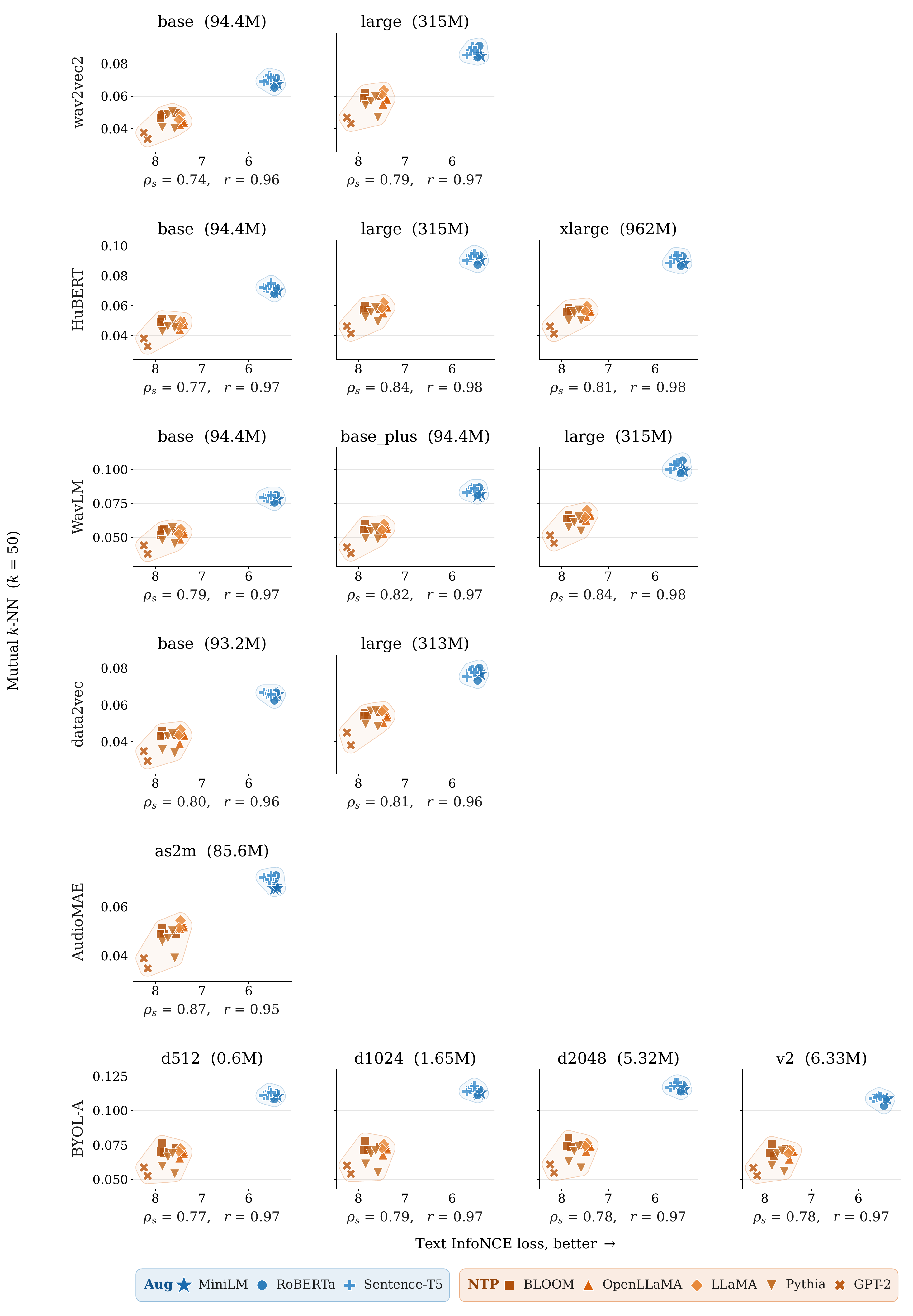}
    \caption{\textbf{CMA by InfoNCE across audio architectures on the AudioCaps dataset.} CMA by unimodal text InfoNCE loss across different audio encoder families (wav2vec2, HuBERT, WavLM, data2vec, AudioMAE, and BYOL-A) at various capacity scales. Notably, the unimodal Aug-SSL loss is highly correlated with CMA, clearly separating low- and high-alignment models, and having large Spearman ($\rho_s$) and Pearson ($r$) correlations.}
    \label{fig:loss_zoo_audiocaps}
\end{figure}

Fig. \ref{fig:loss_zoo_clotho} and \ref{fig:loss_zoo_audiocaps} extend our analysis of the unimodal loss (Fig. \ref{fig:pretrained_audio}c) to these audio architectures, on Clotho and AudioCaps, respectively. The strong correlations between lower unimodal text Aug-SSL InfoNCE loss and higher cross-modal alignment remain remarkably robust across every evaluated audio encoder family and scale.

\section{Additional Cross-Modal Alignment Metrics} \label{app:more_cma_metrics}

In Sec. \ref{sec:results:pretrained}, the evaluation focused on Mutual k-NN with k=50 for the CMA metric. To demonstrate that the connection between unimodal Aug-SSL objectives and CMA holds universally across different local CMA metrics on different $ k$ values, this section extends our analysis to a broader suite of Mutual k-NN and CKNNA on 4 different $k$ values.

\paragraph{CKNNA.} CKNNA is a popular local restriction of CKA:
it evaluates the same kernel alignment, but only on the pairs that are mutual
$k$-nearest neighbors in \emph{both} embedding spaces. Given paired observations
$\mathcal{S} = (x_1,y_1),\dots,(x_n,y_n) \in \mathcal{E}_X \times \mathcal{E}_Y$,
let $K_{ij} = k_X\big(e_X(x_i), e_X(x_j)\big)$ and
$L_{ij} = k_Y\big(e_Y(y_i), e_Y(y_j)\big)$ be the two kernel matrices (usually linear), and define
the mutual-neighbor mask
\begin{equation}
  M_{ij} \;=\; \mathbf{1}\!\left[x_j \in N_X^k(x_i)\right]\cdot
               \mathbf{1}\!\left[y_j \in N_Y^k(y_i)\right],
  \qquad M_{ii} = 0 .
\end{equation}
Replacing each kernel by its masked counterpart in Eq.~\ref{eq:cka} gives
\begin{equation}
  \mathrm{CKNNA}_k(\mathcal{E}_X, \mathcal{E}_Y)
  \;=\;
  \frac{\mathrm{HSIC}\big(M \odot K,\, M \odot L\big)}
       {\sqrt{\mathrm{HSIC}\big(M \odot K,\, M \odot K\big)\,
              \mathrm{HSIC}\big(M \odot L,\, M \odot L\big)}}
  \;\in\; [0,1].
  \label{eq:cknna}
\end{equation}

\begin{figure}
    \centering
    \includegraphics[width=1\linewidth]{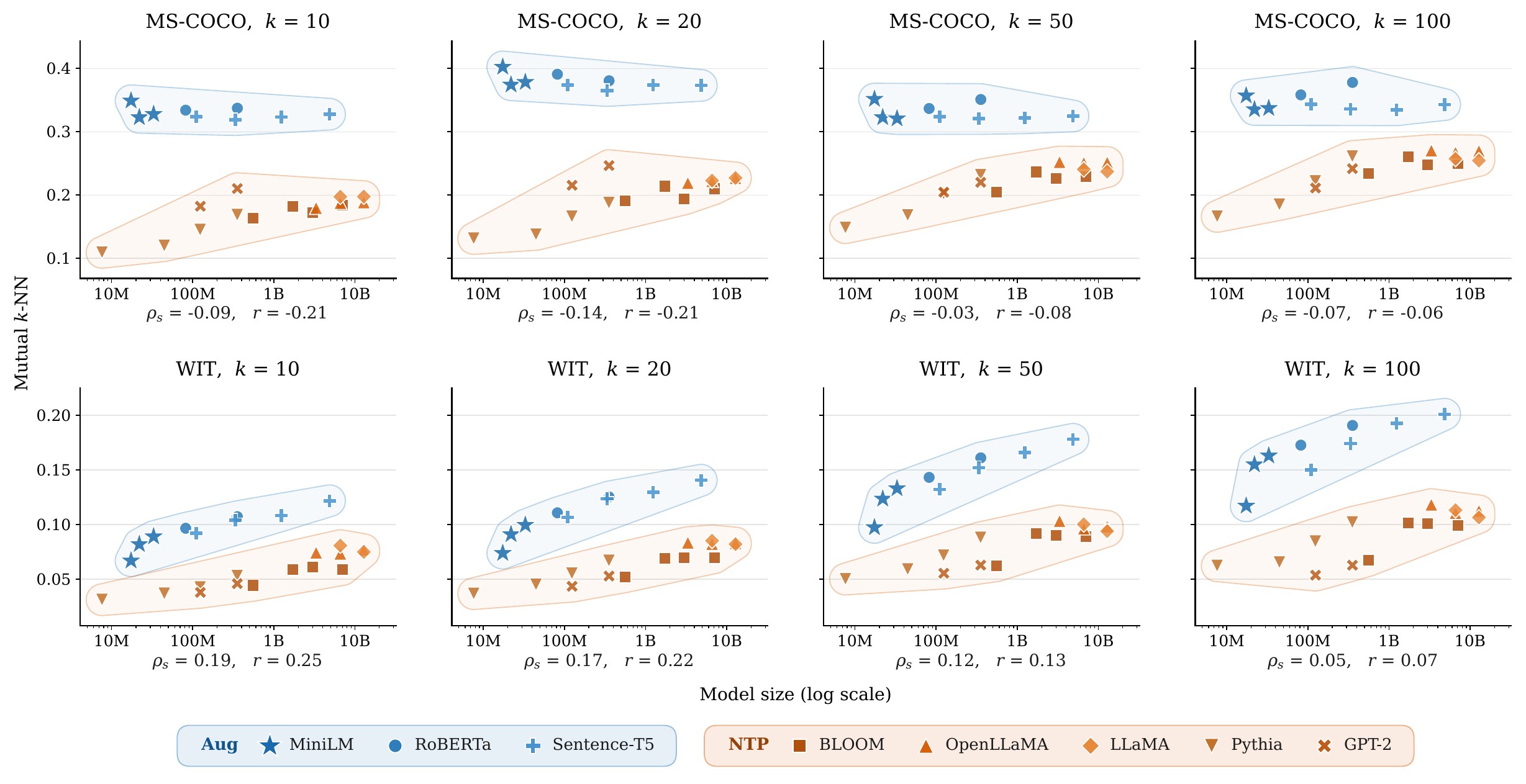}
    \caption{\textbf{Mutual k-NN across k values on the COCO and WIT datasets.} Mutual k-NN by model size (number of parameters) across different k values for the metric. Notably, Aug models consistently achieve higher CMA, even at substantially smaller model and corpus sizes. Also, considering all k values jointly, model/corpus sizes are uncorrelated with CMA, indicated by both weak Spearman ($\rho_s$) and Pearson ($r$) correlations.}
    \label{fig:parameters_mknn_ksweep}
\end{figure}

\begin{figure}
    \centering
    \includegraphics[width=1\linewidth]{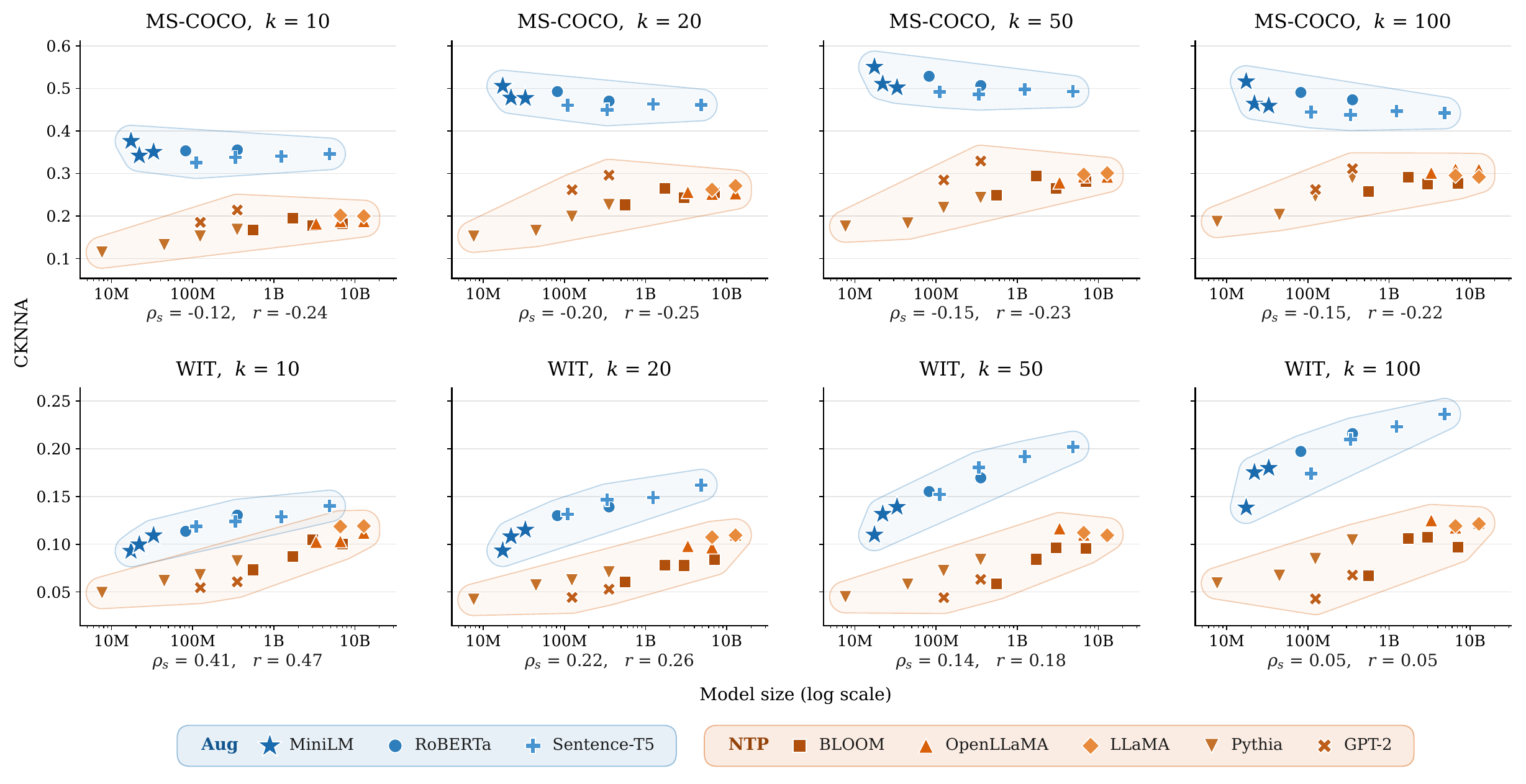}
    \caption{\textbf{CKNNA across k values on the COCO and WIT datasets.} CKNNA by model size (number of parameters) across different k values for the metric. Notably, Aug models consistently achieve higher CMA, even at substantially smaller model and corpus sizes. Also, considering all k values jointly, model/corpus sizes are uncorrelated with CMA, indicated by both weak Spearman ($\rho_s$) and Pearson ($r$) correlations.}
    \label{fig:parameters_cknna_ksweep}
\end{figure}

Fig. \ref{fig:parameters_mknn_ksweep} and \ref{fig:parameters_cknna_ksweep} extend Fig. \ref{fig:pretrained_uncorr}a to Mutual k-NN and CKNNA (respectively) with different $k$ values on COCO and WIT. Consistent with our primary findings, the results demonstrate that augmentation-based text models (Aug) consistently achieve higher cross-modal alignment across the tested CMA metrics and across k values compared to next-token prediction (NTP) models of comparable or even substantially larger sizes.

\begin{figure}
    \centering
    \includegraphics[width=1\linewidth]{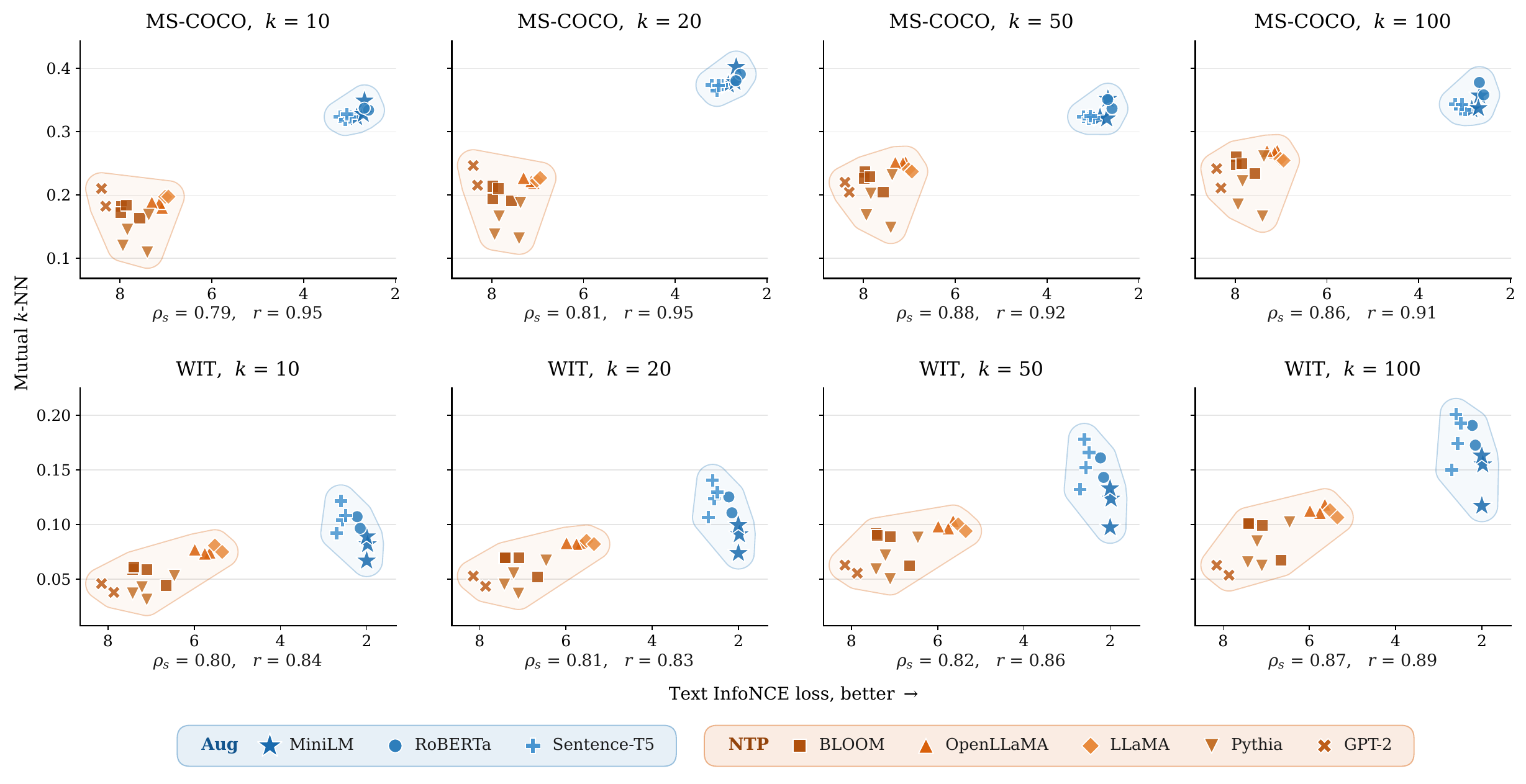}
    \caption{\textbf{Mutual k-NN by InfoNCE across k values on the COCO and WIT datasets.} Mutual k-NN by unimodal text InfoNCE loss across different k values for the metric. Notably, the unimodal Aug-SSL loss is highly correlated with CMA, clearly separating low- and high-alignment models, and having large Spearman ($\rho_s$) and Pearson ($r$) correlations.}
    \label{fig:loss_mknn_ksweep}
\end{figure}

\begin{figure}
    \centering
    \includegraphics[width=1\linewidth]{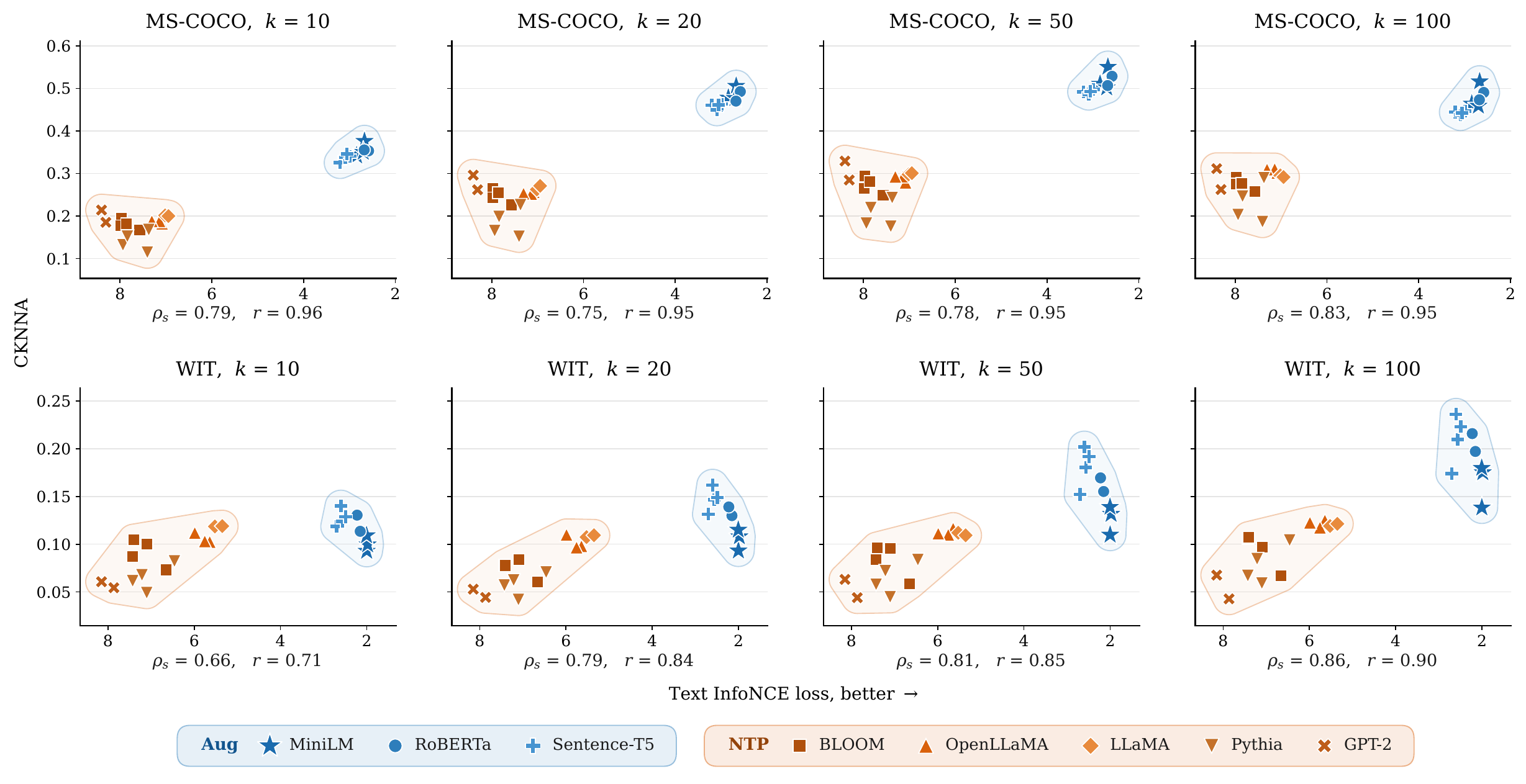}
    \caption{\textbf{CKNNA by InfoNCE across k values on the COCO and WIT datasets.} CKNNA by unimodal text InfoNCE loss across different k values for the metric. Notably, the unimodal Aug-SSL loss is highly correlated with CMA, clearly separating low- and high-alignment models, and having large Spearman ($\rho_s$) and Pearson ($r$) correlations.}
    \label{fig:loss_cknna_ksweep}
\end{figure}

Fig. \ref{fig:loss_mknn_ksweep} and \ref{fig:loss_cknna_ksweep} extend Fig. \ref{fig:pretrained_loss}a to CKNNA with different k values, on COCO and on WIT. The strong correlation between lower unimodal text Aug-SSL InfoNCE loss and higher cross-modal alignment remains remarkably robust regardless of the specific local CMA metric or the k value.

\section{Larger Test Set Sizes} \label{app:test_size}
In Sec. \ref{sec:results:pretrained}, the evaluation used a test set of size 1024. This section extends our analysis to additional larger test set sizes.

\begin{figure}
    \centering
    \includegraphics[width=1\linewidth]{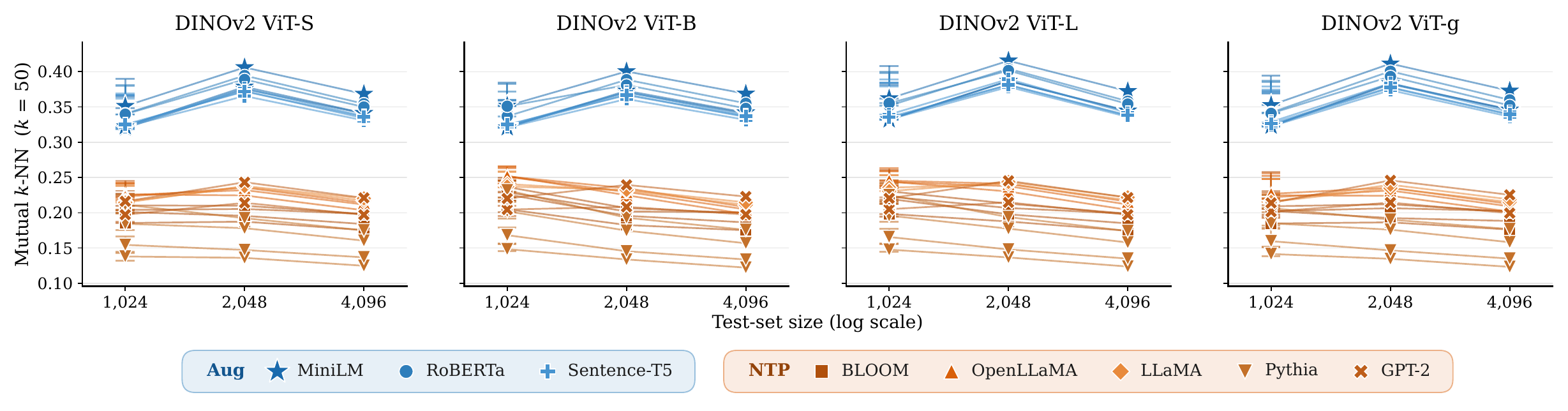}
    \caption{\textbf{CMA by test set size across capacity scales on the COCO dataset.} CMA by test size across different DINOv2 capacity scales. Notably, all text models keep about the same relation across different test set size.}
    \label{fig:valsize_mknn}
\end{figure}

Fig. \ref{fig:valsize_mknn} extends Fig. \ref{fig:pretrained_uncorr} to different test set sizes on COCO. Consistent with our primary findings, the results demonstrate that the text models keep about the same relation between them across different test set sizes and model capacities.

\section{Different Temperatures} \label{app:more_tau}

\begin{figure}[h]
    \centering
    \includegraphics[width=0.6\linewidth]{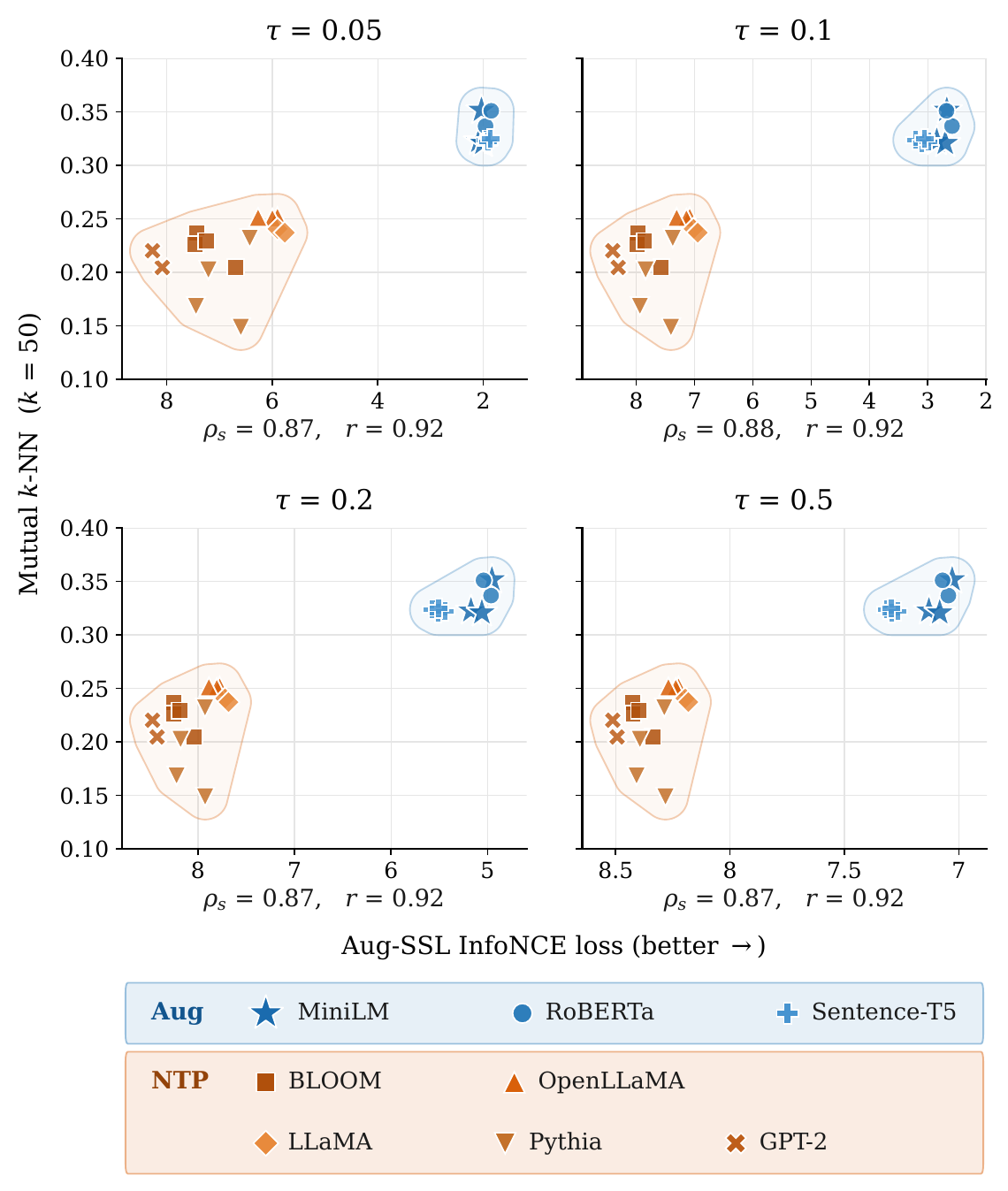}
    \caption{\textbf{CMA by InfoNCE across different temperature on the COCO dataset.} CMA by unimodal text InfoNCE loss across different temperature for the InfoNCE loss. Notably, the unimodal Aug-SSL loss is highly correlated with CMA, very consistent across different temperature,  clearly separating low- and high-alignment models, and having large Spearman ($\rho_s$) and Pearson ($r$) correlations.}
    \label{fig:loss_all_taus}
\end{figure}

In Sec. \ref{sec:results:pretrained}, the evaluation used InfoNCE loss with temperature $\tau=0.1$. Fig. \ref{fig:loss_all_taus} extends Fig. \ref{fig:pretrained_loss}a to different temperatures on COCO. The strong correlation between lower unimodal text Aug-SSL InfoNCE loss and higher cross-modal alignment remains robust regardless of the specific temperature of the InfoNCE loss.

\section{CKA-RBF and \mknn} \label{app:guasscka_mknn}

\begin{figure}[h]
    \centering
    \includegraphics[width=1\linewidth]{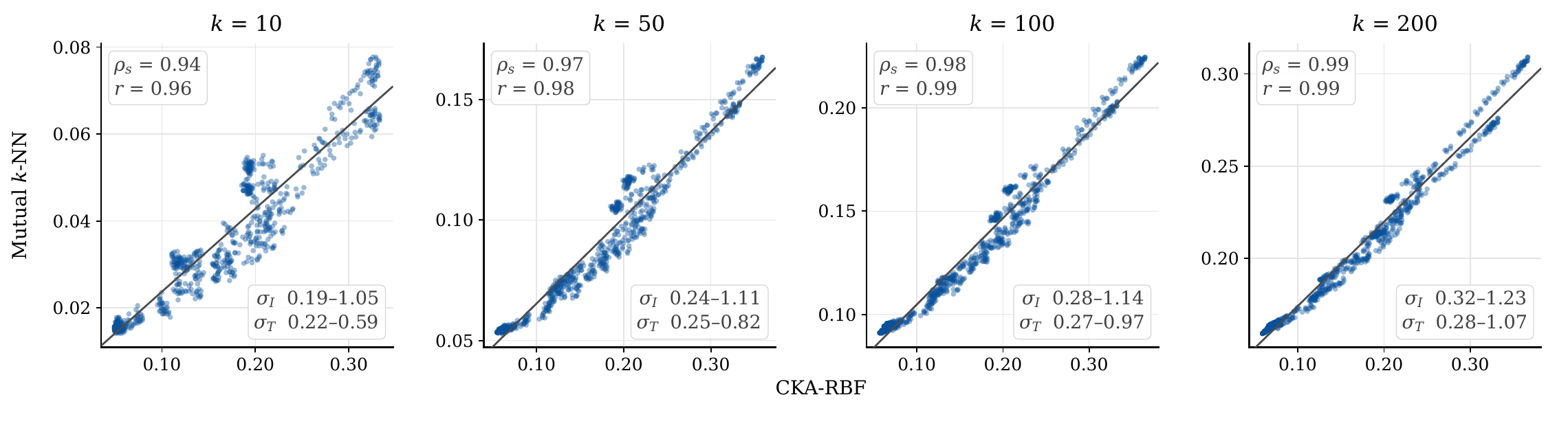}
    \caption{\textbf{Mutual k-NN by CKA-RBF across k values on the COCO dataset.} Mutual k-NN by CKA-RBF across different k values for Mutual k-NN. Notably, Mutual k-NN is highly correlated with CKA-RBF, very consistent across different k, and having large Spearman ($\rho_s$) and Pearson ($r$) correlations.}
    \label{fig:corr_ksweep}
\end{figure}

Fig. \ref{fig:corr_ksweep} extends Fig. \ref{fig:theory}b from Sec. \ref{sec:theory}, evaluating the correlation between Mutual k-NN and CKA-RBF, to different $k$ values. The strong correlation between Mutual k-NN and CKA-RBF remains robust regardless of the specific k value, while adapting the scale $\sigma$ accordingly. This confirms that CKA-RBF can be viewed as a continuous relaxation of the Mutual k-NN.

\section{Visualizing the Bounds} \label{app:theory_bound}

This section empirically visualizes the bounds from Thm. \ref{thm:align_main} and Thm. \ref{thm:misalign_main} across different parameter options.

\begin{figure}[h]
    \centering
    \includegraphics[width=0.7\linewidth]{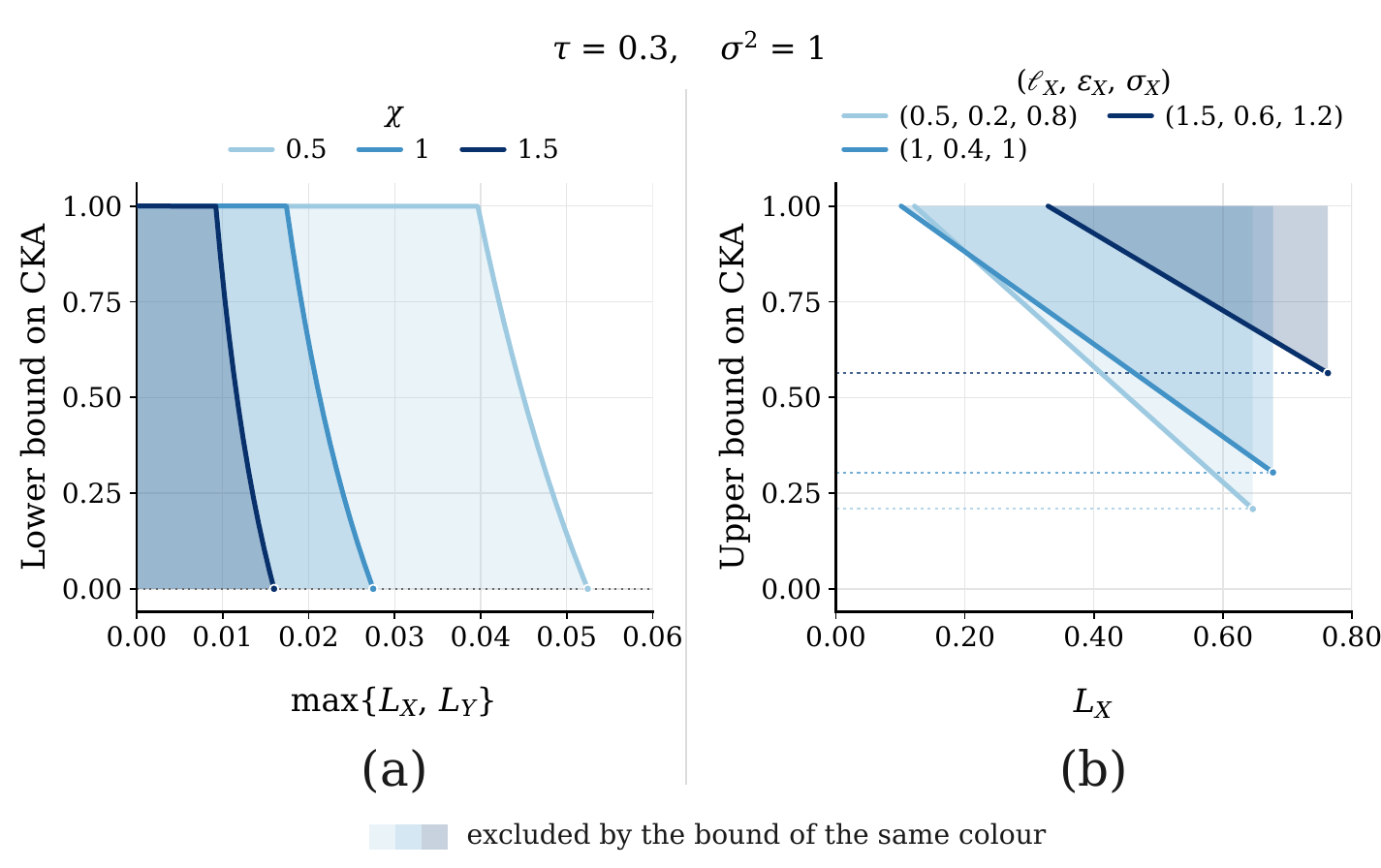}
    \caption{\textbf{Visualization of the (a) lower and (b) upper bounds from Thm. \ref{thm:align_main} and \ref{thm:misalign_main}.}}
    \label{fig:app:both_buonds}
\end{figure}

Fig. \ref{fig:app:both_buonds} visualizes the bounds in $\tau=0.3$ and $\sigma^2=1$. As depicted in Fig. \ref{fig:app:both_buonds}a, the lower bound is effective for sufficiently low losses, and in Fig. \ref{fig:app:both_buonds}b, the upper bound is effective for sufficiently high losses. The latter confirms that low unimodal augmentation-based losses guarantee alignment, whereas only one high loss prevents alignment.

\begin{figure}
    \centering
    \includegraphics[width=0.8\linewidth]{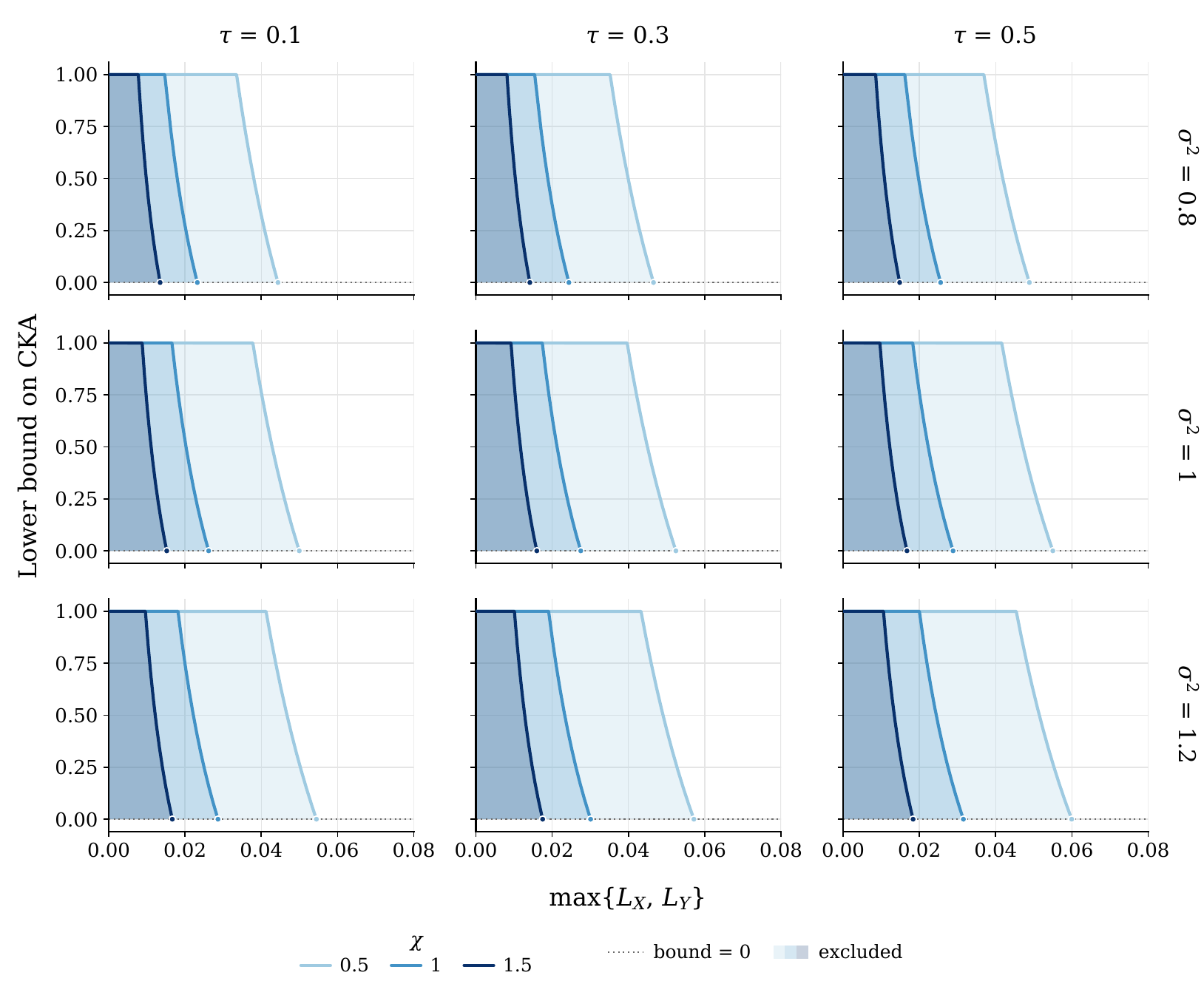}
    \caption{\textbf{Extended visualization of the lower bound from Thm. \ref{thm:align_main},} across different choices of $\tau$ and $\sigma^2$.}
    \label{fig:app:lower_bound}
\end{figure}

\begin{figure}
    \centering
    \includegraphics[width=0.8\linewidth]{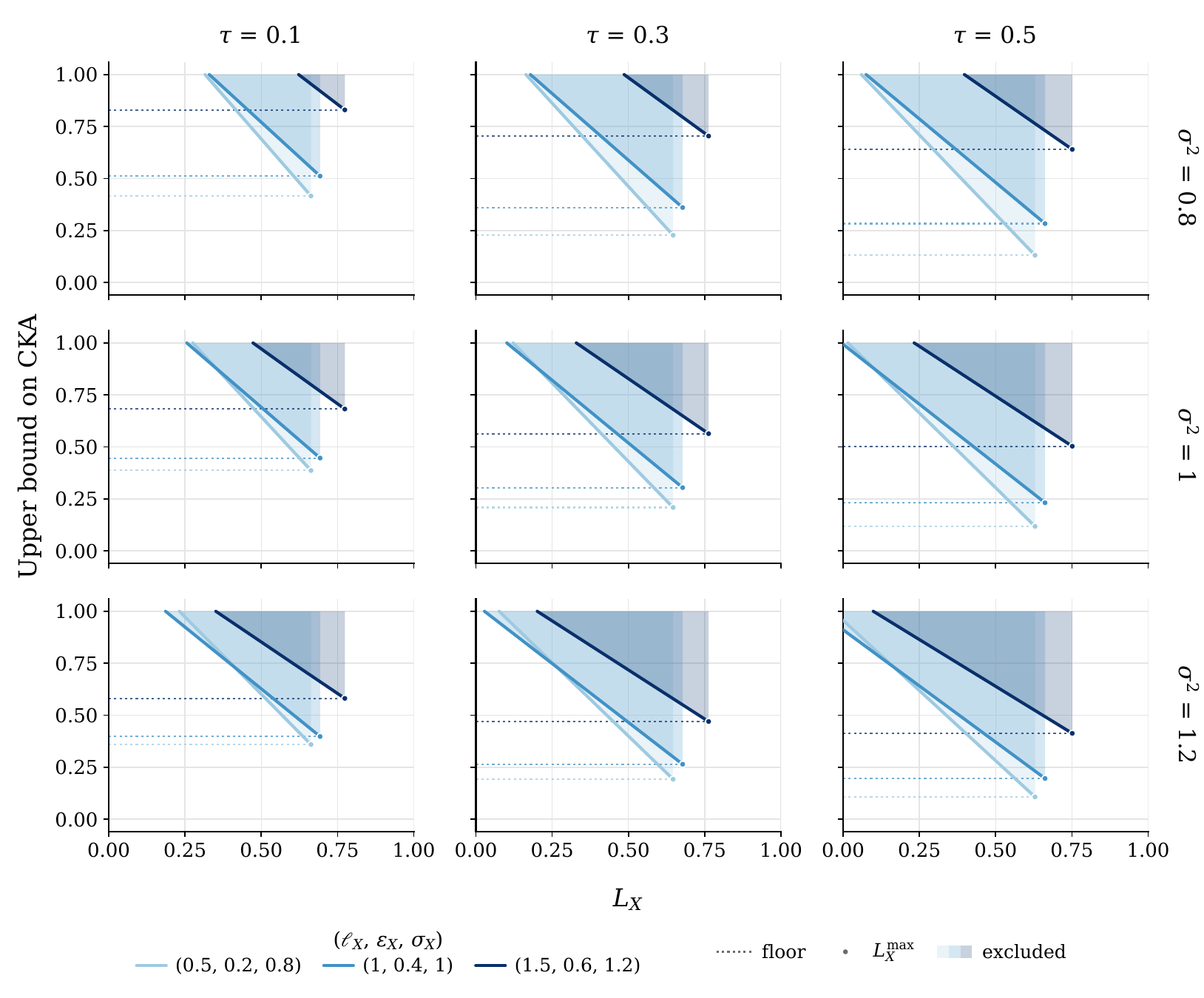}
    \caption{\textbf{Extended visualization of the upper bound from Thm. \ref{thm:misalign_main},} across different choices of $\tau$ and $\sigma^2$.}
    \label{fig:app:upper_bound}
\end{figure}

Fig. \ref{fig:app:lower_bound} and \ref{fig:app:upper_bound} extends \ref{fig:app:both_buonds} to different values of $\tau$ and $\sigma^2$, on Thm. \ref{thm:align_main} and Thm. \ref{thm:misalign_main}, respectively. This show robustness of the bounds across loss and CMA parameters.

\section{Implementation Details} \label{app:implementation_details}
This section specifies the experimental configurations used in Sec.~\ref{sec:results}.

\subsection{Models} \label{app:implementation_details:models}
\paragraph{Pre-trained encoders.}
Tables~\ref{tab:app:text-zoo}, \ref{tab:app:vision-zoo}, and \ref{tab:app:audio-zoo} summarize the 24 text, 14 vision, and 15 audio frozen encoders used in our experiments. $L$ denotes the number of hidden layers, excluding the embedding output, and $d$ denotes the corresponding representation width.
The model \texttt{paraphrase-MiniLM-L3-v2} has no published corpus-size estimate. We therefore exclude it from analyses involving corpus size. 

\begin{table}[t]
\centering
\caption{\textbf{The 24 frozen text encoders.} Aug denotes augmentation-based sentence encoders and NTP denotes next-token-prediction language models. Corpus size is measured as the number of unique tokens across all training stages, as described below.}
\vspace{2pt}
\label{tab:app:text-zoo}
\begin{adjustbox}{max width=\linewidth}
\begin{tabular}{lllrrl}
\toprule
Checkpoint & Family & Type & Params & $L\,/\,d$ & Corpus (tokens) \\
\midrule
\texttt{sentence-transformers/paraphrase-MiniLM-L3-v2} & MiniLM & Aug & 17.4\,M & 3 / 384 & - \\
\texttt{sentence-transformers/all-MiniLM-L6-v2} & MiniLM & Aug & 22.0\,M & 6 / 384 & $1.03\times10^{10}$ \\
\texttt{sentence-transformers/all-MiniLM-L12-v2} & MiniLM & Aug & 33.0\,M & 12 / 384 & $1.03\times10^{10}$ \\
\texttt{sentence-transformers/all-distilroberta-v1} & RoBERTa & Aug & 82.0\,M & 6 / 768 & $2.59\times10^{10}$ \\
\texttt{sentence-transformers/all-roberta-large-v1} & RoBERTa & Aug & 355.4\,M & 24 / 1024 & $3.87\times10^{10}$ \\
\texttt{sentence-transformers/sentence-t5-base} & Sentence-T5 & Aug & 110.2\,M & 12 / 768 & $1.55\times10^{11}$ \\
\texttt{sentence-transformers/sentence-t5-large} & Sentence-T5 & Aug & 335.7\,M & 24 / 1024 & $1.55\times10^{11}$ \\
\texttt{sentence-transformers/sentence-t5-xl} & Sentence-T5 & Aug & 1.24\,B & 24 / 1024 & $1.55\times10^{11}$ \\
\texttt{sentence-transformers/sentence-t5-xxl} & Sentence-T5 & Aug & 4.87\,B & 24 / 1024 & $1.55\times10^{11}$ \\
\addlinespace
\texttt{bigscience/bloom-560m} & BLOOM & NTP & 559.2\,M & 24 / 1024 & $3.50\times10^{11}$ \\
\texttt{bigscience/bloom-1b7} & BLOOM & NTP & 1.72\,B & 24 / 2048 & $3.50\times10^{11}$ \\
\texttt{bigscience/bloom-3b} & BLOOM & NTP & 3.00\,B & 30 / 2560 & $3.50\times10^{11}$ \\
\texttt{bigscience/bloom-7b1} & BLOOM & NTP & 7.07\,B & 30 / 4096 & $3.50\times10^{11}$ \\
\texttt{openlm-research/open\_llama\_3b} & OpenLLaMA & NTP & 3.32\,B & 26 / 3200 & $1.00\times10^{12}$ \\
\texttt{openlm-research/open\_llama\_7b} & OpenLLaMA & NTP & 6.61\,B & 32 / 4096 & $1.00\times10^{12}$ \\
\texttt{openlm-research/open\_llama\_13b} & OpenLLaMA & NTP & 12.85\,B & 40 / 5120 & $1.00\times10^{12}$ \\
LLaMA-1 7B & LLaMA & NTP & 6.61\,B & 32 / 4096 & $1.00\times10^{12}$ \\
LLaMA-1 13B & LLaMA & NTP & 12.85\,B & 40 / 5120 & $1.00\times10^{12}$ \\
\texttt{EleutherAI/pythia-14m} & Pythia & NTP & 7.6\,M & 6 / 128 & $3.00\times10^{11}$ \\
\texttt{EleutherAI/pythia-70m} & Pythia & NTP & 44.7\,M & 6 / 512 & $3.00\times10^{11}$ \\
\texttt{EleutherAI/pythia-160m} & Pythia & NTP & 123.7\,M & 12 / 768 & $3.00\times10^{11}$ \\
\texttt{EleutherAI/pythia-410m} & Pythia & NTP & 353.8\,M & 24 / 1024 & $3.00\times10^{11}$ \\
\texttt{openai-community/gpt2} & GPT-2 & NTP & 124.4\,M & 12 / 768 & $8.25\times10^{9}$ \\
\texttt{openai-community/gpt2-medium} & GPT-2 & NTP & 354.8\,M & 24 / 1024 & $8.25\times10^{9}$ \\
\bottomrule
\end{tabular}
\end{adjustbox}
\end{table}
\begin{table}[t]
\centering
\caption{\textbf{The 14 frozen vision encoders.} All models are loaded through \texttt{timm}. Corpus size denotes the number of training images. Fig.~\ref{fig:pretrained_uncorr} and Fig.~\ref{fig:pretrained_loss}a use \texttt{vit\_base\_patch14\_dinov2}; the full 14-model comparison is reported in App.~\ref{app:more_vision_encoders}.}
\vspace{2pt}
\label{tab:app:vision-zoo}
\begin{adjustbox}{max width=\linewidth}
\begin{tabular}{llrrl}
\toprule
\texttt{timm} checkpoint & Family & Params & $L\,/\,d$ & Corpus (images) \\
\midrule
\texttt{vit\_tiny\_patch16\_224.augreg\_in21k} & ImageNet21K & 9.74\,M & 12 / 192 & ImageNet-21k, $1.42\times10^{7}$ \\
\texttt{vit\_small\_patch16\_224.augreg\_in21k} & ImageNet21K & 30.08\,M & 12 / 384 & ImageNet-21k, $1.42\times10^{7}$ \\
\texttt{vit\_base\_patch16\_224.augreg\_in21k} & ImageNet21K & 102.60\,M & 12 / 768 & ImageNet-21k, $1.42\times10^{7}$ \\
\texttt{vit\_large\_patch16\_224.augreg\_in21k} & ImageNet21K & 325.69\,M & 24 / 1024 & ImageNet-21k, $1.42\times10^{7}$ \\
\texttt{vit\_base\_patch16\_224.mae} & MAE & 85.80\,M & 12 / 768 & ImageNet-1k, $1.28\times10^{6}$ \\
\texttt{vit\_large\_patch16\_224.mae} & MAE & 303.30\,M & 24 / 1024 & ImageNet-1k, $1.28\times10^{6}$ \\
\texttt{vit\_huge\_patch14\_224.mae} & MAE & 630.76\,M & 32 / 1280 & ImageNet-1k, $1.28\times10^{6}$ \\
\texttt{vit\_small\_patch14\_dinov2.lvd142m} & DINOv2 & 22.06\,M & 12 / 384 & LVD-142M, $1.42\times10^{8}$ \\
\texttt{vit\_base\_patch14\_dinov2.lvd142m} & DINOv2 & 86.58\,M & 12 / 768 & LVD-142M, $1.42\times10^{8}$ \\
\texttt{vit\_large\_patch14\_dinov2.lvd142m} & DINOv2 & 304.37\,M & 24 / 1024 & LVD-142M, $1.42\times10^{8}$ \\
\texttt{vit\_giant\_patch14\_dinov2.lvd142m} & DINOv2 & 1.14\,B & 40 / 1536 & LVD-142M, $1.42\times10^{8}$ \\
\texttt{vit\_base\_patch16\_clip\_224.laion2b} & CLIP & 86.19\,M & 12 / 768 & LAION-2B-en, $2.32\times10^{9}$ \\
\texttt{vit\_large\_patch14\_clip\_224.laion2b} & CLIP & 303.97\,M & 24 / 1024 & LAION-2B-en, $2.32\times10^{9}$ \\
\texttt{vit\_huge\_patch14\_clip\_224.laion2b} & CLIP & 632.08\,M & 32 / 1280 & LAION-2B-en, $2.32\times10^{9}$ \\
\bottomrule
\end{tabular}
\end{adjustbox}
\end{table}
\begin{table}[t]
\centering
\caption{\textbf{The 15 frozen audio encoders.} $d$ denotes the widest read-out block. BYOL-A models are convolutional and use unequal block widths: 2048/1024/512/512/512 (\texttt{d512}), 2048/1024/512/1024/1024 (\texttt{d1024}), 2048/1024/512/2048/2048 (\texttt{d2048}), and 2048/1024/2048/2048/3072 (v2). BYOL-A checkpoints are obtained from the authors' repository.}
\vspace{2pt}
\label{tab:app:audio-zoo}
\begin{adjustbox}{max width=\linewidth}
\begin{tabular}{lllrr}
\toprule
Checkpoint & Family & Objective & Params & $L\,/\,d$ \\
\midrule
\texttt{facebook/wav2vec2-base} & wav2vec2 & contrastive & 94.37\,M & 12 / 768 \\
\texttt{facebook/wav2vec2-large-lv60} & wav2vec2 & contrastive & 315.44\,M & 24 / 1024 \\
\texttt{facebook/hubert-base-ls960} & HuBERT & masked & 94.37\,M & 12 / 768 \\
\texttt{facebook/hubert-large-ll60k} & HuBERT & masked & 315.44\,M & 24 / 1024 \\
\texttt{facebook/hubert-xlarge-ll60k} & HuBERT & masked & 962.50\,M & 48 / 1280 \\
\texttt{microsoft/wavlm-base} & WavLM & masked & 94.38\,M & 12 / 768 \\
\texttt{microsoft/wavlm-base-plus} & WavLM & masked & 94.38\,M & 12 / 768 \\
\texttt{microsoft/wavlm-large} & WavLM & masked & 315.45\,M & 24 / 1024 \\
\texttt{facebook/data2vec-audio-base} & data2vec & masked & 93.16\,M & 12 / 768 \\
\texttt{facebook/data2vec-audio-large} & data2vec & masked & 313.28\,M & 24 / 1024 \\
\texttt{AudioNTT2020-BYOLA-64x96d512} & BYOL-A & augmentation & 0.60\,M & 5 / 2048 \\
\texttt{AudioNTT2020-BYOLA-64x96d1024} & BYOL-A & augmentation & 1.65\,M & 5 / 2048 \\
\texttt{AudioNTT2020-BYOLA-64x96d2048} & BYOL-A & augmentation & 5.32\,M & 5 / 2048 \\
\texttt{AudioNTT2022-BYOLA-64x96d2048} & BYOL-A & augmentation & 6.33\,M & 5 / 3072 \\
\texttt{gaunernst/vit\_base\_patch16\_1024\_128.audiomae\_as2m} & AudioMAE & masked & 85.65\,M & 12 / 768 \\
\bottomrule
\end{tabular}
\end{adjustbox}
\end{table}
\paragraph{Corpus sizes.}
We obtained the corpus-sizes in Tables~\ref{tab:app:text-zoo} and \ref{tab:app:vision-zoo} from published sources and classified them as exact when the source reports the corresponding count, or approximate when the value is derived from a training schedule or byte count.
For NTP encoders, corpus sizes reported in bytes are converted using 4.85 bytes per BPE token. For MiniLM students, the reported 3.3B-word corpus is converted using 1.30 WordPiece tokens per word.
For augmentation-based encoders, corpus size includes the backbone pre-training corpus and the distinct examples encountered during the contrastive training stage. We estimate the latter using the average: 2.078 texts per training example and 53.9 tokens per text. This quantifies the number of distinct tokens encountered. Sentence-T5 is the only family for which the number of contrastive training steps is unpublished; we therefore use GTR's 800k steps for its corresponding $2\times10^{9}$-pair training stage.
Vision corpus size is measured as the number of unique training images.

\paragraph{Embedding extraction.}
Table~\ref{tab:app:extraction} summarizes the extraction settings.

\begin{table}[t]
\centering
\caption{\textbf{Embedding extraction.} All hidden layers are retained. Audio is converted to mono and resampled to 16\,kHz, then divided into non-overlapping 10\,s windows. Each window is embedded independently and the resulting vectors are averaged using weights proportional to the number of valid samples in each window. Clotho clips contain an average of 2.647 windows, whereas AudioCaps clips contain exactly one. The waveform-transformer models receive normalized waveforms; BYOL-A receives 64-band log-mel spectrograms normalized using evaluation-set statistics (mean $-5.4703$, standard deviation $4.3999$); AudioMAE receives a $1024\times128$ spectrogram represented using a $64\times8$ patch grid.}
\vspace{2pt}
\label{tab:app:extraction}
\begin{adjustbox}{max width=\linewidth}
\begin{tabular}{lll}
\toprule
Modality & Read-out per layer & Input \\
\midrule
Text (Aug) & mean over real tokens (SBERT pooling) & 128 tokens \\
Text (NTP) & mean over real tokens & 128 tokens \\
Vision & CLS at the second residual add & each model's own preprocessing \\
Audio & mean over time (SUPERB convention) & 10\,s windows at 16\,kHz \\
\bottomrule
\end{tabular}
\end{adjustbox}
\end{table}

\paragraph{Controlled architectures and training.}
Table~\ref{tab:app:controlled} specifies the controlled architectures and training schedules. Both training objectives operate on the 128-dimensional projection-head representations, and alignment is evaluated in the same space. The vision models used a common preprocessing pipeline consisting of resizing the short side to 256 pixels, centre cropping to $224\times224$, and ImageNet normalization.
The curves in Fig.~\ref{fig:controlled_results} and the corresponding appendix figures report the mean $\pm$ one sample standard deviation over three training seeds. Alignment is evaluated on a fixed held-out draw of 4096 of the COCO 2017 validation set pairs.

\begin{table}[t]
\centering
\caption{\textbf{The controlled encoders.} Both models use AdamW with linear warmup followed by cosine decay to zero, 15\,000 training steps, batches of 256 items with five views per item (1280 views per step), $\tau=0.1$ as in Eq.~\ref{eq:infonce-k}, bfloat16 autocasting on compute capability $\geq 8.0$, and float16 with gradient scaling otherwise. Contrastive logits are always computed in float32. No EMA is used. Training uses seeds $5000$, $5001$, and $5002$.}
\vspace{2pt}
\label{tab:app:controlled}
\begin{adjustbox}{max width=\linewidth}
\begin{tabular}{lll}
\toprule
& Vision encoder (ViT) & Text encoder (SBERT) \\
\midrule
Backbone & ViT-S/16, randomly initialised & \texttt{bert-base-uncased}, mean pooling \\
Architecture & 12 layers, 6 heads, width 384, MLP 1536 & 12 layers, width 768 \\
Input & $224^2$, patch 16 & 40 tokens \\
Projection head & $384\to384\to128$ with GELU & $768\to768\to128$ with GELU \\
Learning rate & $5\times10^{-4}$ & $2\times10^{-5}$ \\
Weight decay & $0.05$ & $0.01$ \\
Warmup steps & 1000 & 300 \\
Gradient clip & none & global norm $1.0$ \\
\bottomrule
\end{tabular}
\end{adjustbox}
\end{table}

\paragraph{Loss maximization and chance level.}
The loss-maximizing condition differs from the corresponding minimizing condition only in the sign of the backward pass. That is, initialization, minibatch stream, optimization schedule, and gradient clipping remain unchanged.
For mutual $k$-NN overlap, the chance level is the expected overlap of two independent $k$-neighborhoods among $n$ instances, which is $k/(n-1)$. For $k=50$ and $n=4096$, this gives $50/4095=0.0122$.

\paragraph{Multi-positive InfoNCE objective.}
Equation~\ref{eq:infonce-k} contains one positive per anchor, whereas our controlled experiments use five views per item \citep{tian2023stablerep}. Let a batch contain $B$ items with $V$ views per item, let $z_1,\ldots,z_{BV}$ denote the unit-normalized projection-head outputs, and let $P(i)$ denote the $V-1$ other views of the same item as anchor $i$. All views belonging to other items are treated as negatives. The resulting objective is
\begin{equation} \label{eq:infonce-multi}
\ell^{(V)} \;=\; -\frac{1}{BV}\sum_{i=1}^{BV}\frac{1}{V-1}\sum_{j\in P(i)}
\log\frac{\exp\big(\langle z_i,z_j\rangle/\tau\big)}
{\sum_{l\neq i}\exp\big(\langle z_i,z_l\rangle/\tau\big)}.
\end{equation}
We use $V=5$. When $V=2$, the objective reduces to Eq.~\ref{eq:infonce-k} with $K=2B-2$.
\paragraph{Evaluating the unimodal loss of frozen encoders.}
The loss shown on the $x$-axis of Fig.~\ref{fig:pretrained_loss} is Eq.~\ref{eq:infonce-multi} evaluated on the frozen representations of the five views in the evaluation draw. The complete evaluation draw is processed jointly rather than in mini-batches, so each anchor is compared against all other rows. With $n=1024$ and $V=5$, this gives $K=nV-1=5119$ negatives.
We use $\tau=0.1$. For each encoder, the reported loss is the minimum across layers, mirroring the maximum used for alignment: both read out the model at its best layer. The minimum keeps the loss a property of the encoder alone, identical across all panels. The corresponding chance loss is $\log(nV-1)=8.54$ nats.

\paragraph{Supervised baselines.}
MS-COCO instance annotations are reduced to the largest object per image among the 80 categories. The cross-entropy baseline uses a single $\mathrm{Linear}(384,80)$ classification head on the backbone representation. SupCon uses the same 128-dimensional projection head as the augmentation-based SSL objective, with $\tau=0.1$; all same-label views in a batch are treated as positives, while anchors without another sample of the same label are omitted.
All three objectives are trained from random initialization on the same training pool for 150 epochs (23,250 steps), using batch size 256, learning rate $5\times10^{-4}$, weight decay $0.05$, and 775 warmup steps. Thus, the three baselines differ only in their objectives

To control for label memorization, classes are partitioned according to COCO supercategories. The \emph{animal}, \emph{furniture}, \emph{food}, and \emph{kitchen} categories constitute the seen set of 33 classes (39,867 images); the remaining 46 classes are held out. The \emph{person} category is omitted. Fig.~\ref{fig:controlled_results}b evaluates CMA on the 1,716 validation images belonging to held-out classes.

\paragraph{Alignment metrics.}
The bandwidth of the CKA-RBF, $\sigma$, is determined for each modality as the median distance to the $k$-th nearest neighbor across samples.

\paragraph{Layer aggregation and null calibration.} We follow \cite{groger2026revisiting}. For frozen encoders, the reported alignment score is the maximum across all image--text layer pairs. Since this maximum is affected by the number of layers, we calibrate each encoder pair as presented in \cite{groger2026revisiting}.

\subsection{Datasets} \label{app:implementation_details:datasets}
\paragraph{Evaluation draw.}
Unless otherwise stated, each dataset is evaluated using a fixed subsample of $n=1024$ aligned pairs drawn without replacement using a global seed. The same draw is reused across experimental stages. The larger evaluation sets in App.~\ref{app:test_size} are nested supersets, $1024\subset2048\subset4096$, so that the sample-size analysis varies only $n$.
\paragraph{MS-COCO.}
Evaluation uses \texttt{val2017}. Images with all five captions provide 5,000 image--caption pairs; 4,952 of these images contain an instance annotation and therefore have an associated label. The evaluation draw is sampled from these images. Captions are ordered by annotation ID, and the first caption is used.
Controlled experiments use \texttt{train2017}, containing 118,287 images.

\paragraph{WIT.}
Evaluation uses the \texttt{wit\_1024} benchmark split of \citet{huh2024platonic}, containing 1,024 image--text pairs with one caption per image. For controlled training, we construct a COCO-sized corpus from \texttt{wikimedia/wit\_base} by retaining the first English reference caption for each image.
Benchmark images are removed from the training corpus by caption matching. The remaining data are split using a seeded shuffle into 118,287 training images and 5,000 validation images, matching the sizes of \texttt{train2017} and \texttt{val2017}. 

\paragraph{Clotho.}
We use Clotho v2.1 (Zenodo \texttt{10.5281/zenodo.4783391}), specifically its evaluation split containing 1,045 clips and 5,225 captions. Clips have durations between 15 and 30 seconds (mean 22.5 seconds) and a native sampling rate of 44.1 kHz. We sample 1,024 clips for evaluation. Audio is converted to mono, resampled to 16 kHz, and stored. Each clip has five human captions in a fixed column order, which defines the view order, and we use the first.

\paragraph{AudioCaps.}
Among the 1,470 clips in the official validation and test splits, we use all 856 available valid clips, containing a caption and valid WAV files. Each is exactly 10.00 seconds long at 48 kHz and is downsampled by a factor of three to 16 kHz, yielding a single 10-second window per clip.

\paragraph{Image augmentations.}
For the controlled experiments and frozen-encoder image-loss evaluation, each image is transformed using the following pipeline: random resized crop to $224^2$ with area scale $(0.5,1.0)$; horizontal flip with probability $0.5$; color jitter with brightness, contrast, and saturation $0.2$ and hue $0.05$, applied with probability one; and ImageNet normalization. Five independent draws produce the five views of each image.
The label-free control in Fig.~\ref{fig:controlled_results}b uses the DINO-style multi-crop augmentation associated with its baseline. Each image produces two global crops with scale $(0.4,1.0)$ and four local crops with scale $(0.05,0.4)$, all resized to $224^2$. Horizontal flipping is applied to all crops. Color jitter $(0.4,0.4,0.2,0.1)$ is applied with probability $0.8$, grayscale with probability $0.2$, and Gaussian blur with $\sigma\in(0.1,2.0)$ is applied to the first global crop with probability one, to the second global crop with probability $0.1$, and to local crops with probability $0.5$. Solarization is applied to the second global crop with probability $0.2$.
\paragraph{Text augmentations.}
The main text views are generated using Qwen with a single user message and no system message. The generator receives only the source caption and has no access to the corresponding image, other captions, or labels. The prompt is:
\begin{quote}\ttfamily\footnotesize\raggedright\noindent
You rewrite a caption. Output only JSON.\\[2pt]
Rewrite the caption below as EXACTLY \textless{}N\textgreater{} alternatives with the same meaning. Vary word choice, syntax and clause order; the \textless{}N\textgreater{} must differ from each other and from the original by more than a synonym swap.\\
- Keep the original's register and length (a plain caption, 8-15 words). No commentary.\\
- No preamble, no numbering.\\[2pt]
Return only a JSON array of exactly \textless{}N\textgreater{} strings.\\[2pt]
Caption: "\textless{}CAPTION\textgreater{}"
\end{quote}
Here $\texttt{\textless{}N\textgreater{}}=4$, so the source caption constitutes view 0 and the four generated captions constitute views 1--4. Decoding uses temperature $0.9$, top-$p$ $0.95$, and a fixed generation seed. Outputs are required to contain exactly four English strings. Outputs failing this constraint are regenerated up to two times; after two failed attempts, the source caption is used as fallback. This filtering prevents off-language outputs from artificially increasing within-group variation.
For WIT and Clotho, one prompt constraint is modified to account for the properties of their captions. For WIT, whose captions have a median length of seven words, the length instruction is replaced with: Match the original's register and length, however short it is. Keep proper nouns, titles and technical terms verbatim. No commentary. For Clotho, the instruction is replaced with: Keep the original's register and length (a plain caption, 8-20 words). Keep the same events in the same order; do not add, drop or reorder one. No commentary. All remaining prompt components are unchanged. App.~\ref{app:other_llm} repeats the experiment using \texttt{microsoft/Phi-3.5-mini-instruct} and \texttt{google/gemma-4-12B-it}.

\paragraph{Back-translation views.}
Back-translation provides an alternative text augmentation procedure that does not rely on prompting or stochastic decoding. View 0 is the source caption, while views 1--4 are obtained by translating the source through German, French, Spanish, and Russian, respectively, and translating each result back to English. The translations use eight \texttt{Helsinki-NLP/opus-mt} checkpoints, corresponding to the four pivot languages in both translation directions. Decoding uses beam search with four beams and no sampling.
The resulting corpus is deterministic. Each view uses a different pivot language, producing four translation systems rather than four stochastic samples from a single system.

\subsection{The Bounds} \label{app:implementation_details:bounds}
\paragraph{Evaluation of the bounds.}
The bound figures visualize the analytical bounds derived in Sec.~\ref{sec:theory}. No encoder training or sampling is involved.
The lower bound depends on both modality losses, but a single curve suffices because it is given by $\cos(\theta_X+\theta_Y)$ and decreases monotonically in each loss. Thus, the worst case under $\max{\mathcal{L}_X,\mathcal{L}_Y}\leq\mathcal{L}$ occurs on the diagonal. The upper bound depends only on the loss of one modality and is therefore plotted directly against $\mathcal{L}_X$.
A dotted line marks $\mathrm{CKA}=0$. The excluded corner of each bound is also indicated: the lower bound excludes the lower-left region, whereas the upper bound excludes the upper-right region. Each closed-form implementation is additionally verified against a 50-digit evaluation of the same expression.
\paragraph{Parameter grids and valid domains.}
Both bound figures use $\tau\in\{0.1,0.3,0.5\}$ and $\sigma^2\in\{0.8,1.0,1.2\}$. The lower-bound panels additionally use $\chi\in\{0.5,1,1.5\}$. The upper-bound panels use the parameter triples
\[
(\ell_X,\epsilon_X,\sigma_X)\in
{(0.5,0.2,0.8),(1,0.4,1.0),(1.5,0.6,1.2)}.
\]
The side-by-side comparison uses the slice $\tau=0.3$ and $\sigma^2=1$.
Curves are plotted only over the region in which the corresponding bound is non-vacuous. The lower bound terminates at its zero crossing, below which it is weaker than the trivial constraint $\mathrm{CKA}\geq0$. The upper bound is cut where it crosses one and terminates at its maximum possible value as in Lemma \ref{lem:range}.

\subsection{Hardware and Software} \label{app:implementation_details:hardware}
\paragraph{Hardware and Software.}
All experiments use Python on a single GPU without distributed training or gradient accumulation. GPU selection is performed at runtime. The reported experiments were executed on NVIDIA RTX PRO 6000 Blackwell (98\,GB), B200 (183\,GB), H200 (144\,GB), A100 80\,GB (PCIe and SXM4), L40S (46\,GB), RTX 6000 Ada (49\,GB), and L4 (23\,GB) GPUs, with 8--32 CPU cores and 32--150\,GB host memory per job.

\paragraph{Determinism and seeding.}
Evaluation uses fixed random seeds that are independent of the training seeds. These seeds determine evaluation subsampling, loss-evaluation blocks, kernel-statistic samples, alignment subsamples, and frozen augmentation draws. Consequently, different training runs share the same evaluation data and augmentation realizations and differ only through their training seeds.
Caption generation uses a separate generation seed, while back-translation is deterministic and requires no random seed.

\end{document}